%% file: arxiv.tex
\documentclass[10pt,twocolumn,letterpaper]{article}
\usepackage[pagenumbers]{cvpr}
\input{preamble}
\definecolor{cvprblue}{rgb}{0.21,0.49,0.74}
\usepackage[pagebackref,breaklinks,colorlinks,allcolors=cvprblue]{hyperref}
\input{arxiv_authors}
\hypersetup{
  pdfauthor={\arxivPdfAuthors},
  pdftitle={Alignment under Hard Negative Re-Pairing: Cross-Modal Similarity, Scale and Retrieval}
}
\def\confName{arXiv}
\def\confYear{2026}
\title{Alignment under Hard Negative Re-Pairing:\\Cross-Modal Similarity, Scale and Retrieval}
\author{\arxivAuthorBlock}
\providecommand{\nolinenumbers}{}

\begin{document}
\maketitle
\input{sec/0_abstract}
\input{fig/teaser}
\input{sec/1_intro}
\input{sec/2_related}
\input{sec/3_setup}
\input{sec/4_scale}
\input{sec/5_experiments}

\input{sec/6_discussion}
\FloatBarrier
\label{page:main-end}
\input{sec/8_llm_statement}
{\small
\bibliographystyle{ieeenat_fullname}
\bibliography{main}
}
\input{sec/X_suppl}
\end{document}

%% file: preamble.tex
\usepackage[T1]{fontenc}
\usepackage{microtype}
\usepackage{amsthm,mathtools}
\usepackage{array,tabularx,longtable,multirow}
\usepackage{enumitem,float,placeins}
\graphicspath{{fig/}}
\newcolumntype{Y}{>{\raggedright\arraybackslash}X}

\DeclareMathOperator{\tr}{tr}

\AtEndPreamble{%
  \newtheorem{proposition}{Proposition}%
  \crefname{proposition}{Prop.}{Props.}%
  \Crefname{proposition}{Proposition}{Propositions}%
  \crefname{section}{Sec.}{Secs.}%
  \crefname{figure}{Fig.}{Figs.}%
  \crefname{table}{Tab.}{Tabs.}%
  \renewcommand*{\autoref}{\Cref}%
}
\AddToHook{cmd/appendix/before}{%
  \crefalias{section}{appendix}%
  \crefalias{subsection}{appendix}%
  \crefalias{subsubsection}{appendix}%
}

%% file: arxiv_authors.tex
\newcommand{\arxivAuthorBlock}{%
  Charles Weng\textsuperscript{1}\quad
  Alexander Martin\textsuperscript{2}\quad
  Sandra Arcos-Holzinger\textsuperscript{3}\\[3pt]
  Andrew Yates\textsuperscript{2}\quad
  Dixin Zheng\textsuperscript{1}\quad
  Yue Zhang\textsuperscript{1}\quad
  Landi Zhou\textsuperscript{1}\\[6pt]
  {\small\textsuperscript{1}ByteDance Inc.\quad
  \textsuperscript{2}Johns Hopkins University\quad
  \textsuperscript{3}University of Melbourne}%
}
\newcommand{\arxivPdfAuthors}{Charles Weng, Alexander Martin, Sandra Arcos-Holzinger, Andrew Yates, Dixin Zheng, Yue Zhang, Landi Zhou}

%% file: sec/0_abstract.tex
\begin{abstract}
Do vision and language models represent the world alike? That question hides three: what structure each modality has, whether the two modalities organize images and their captions alike, and whether a model ranks an image's own caption highly. Centered kernel alignment (CKA) and neighborhood agreement answer the second, yet matching and model selection use them as if they identified individual pairs and answered the third. To test this reading, we introduce hard negative re-pairing: every image receives the caption of a similar image, and nothing else changes. Although no image keeps its own caption and annotators judge a majority of the substituted captions wrong, CKA stays higher than under a random re-pairing that keeps 80\% of the original pairs. In all 22 image--text models we test, neighborhood agreement loses far less than top-one retrieval, which collapses. The reason is scale: these diagnostics pool over many neighbors, matching and retrieval are decided among the nearest ones, and an exact decomposition shows that near-neighbor pairs contribute almost nothing to CKA. In the matching problems we test, an exact maximizer of CKA would not return the original pairing. On COCO and Flickr30k, the model with the highest diagnostic value retrieves worse on held-out data than the model chosen by validation recall, even where the two correlate positively. What re-pairing removes is the informative part: that share tracks retrieval on both datasets and ranks encoder pairs before alignment training. A diagnostic should be reported with its question and its scale, and tested against hard negative re-pairing.
\end{abstract}

%% file: fig/teaser.tex
\begin{figure*}[t]
\centering
\includegraphics[width=\textwidth]{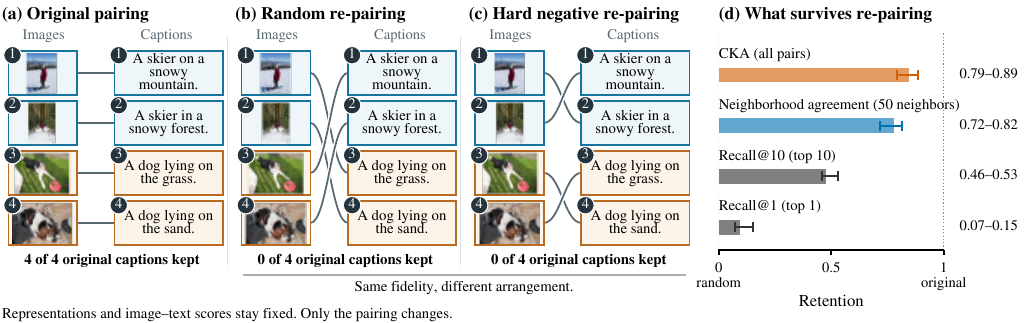}
\caption{\textbf{Hard negative re-pairing, and what survives it.}
(a--c) The same images and captions under three pairings. Both re-pairings keep no original pair. (d) How much of each diagnostic's margin over random re-pairing survives complete hard negative re-pairing, for 22 models on COCO. Diagnostics that pool over many neighbors keep most of it. Top-one retrieval keeps about a tenth.}
\label{fig:teaser}
\end{figure*}

%% file: sec/1_intro.tex
\section{Introduction}
\label{sec:intro}

Do separately trained vision and language models represent the world alike? Prior work reports increasing agreement as models improve~\citep{huh2024platonic,maniparambil2024vision}.
Centered kernel alignment (CKA) compares the similarity structure of paired images and captions~\citep{kornblith2019similarity}. Neighborhood agreement counts neighbors shared across modalities~\citep{huh2024platonic}.
These diagnostics inform unsupervised matching and encoder selection~\citep{maniparambil2024vision,maniparambil2025harnessing,groger2025structure}.
We ask what they establish about the pairing of images with captions and about retrieval.

Interpreting this evidence requires distinguishing three questions.
\begin{enumerate}[label=\textbf{Q\arabic*.},leftmargin=*,itemsep=1pt,topsep=3pt]
\item What structure does each modality have on its own?
\item Do the two modalities organize images and their captions alike?
\item Does a model rank an image's own caption highly?
\end{enumerate}
Statistics of one modality answer Q1. CKA and neighborhood agreement are \emph{relational diagnostics} and answer Q2. Recall@$K$ answers Q3.
Matching requires identifying individual pairs, and model selection requires evidence about Q3. Both uses treat an answer to Q2 as if it identified individual pairs and answered Q3.

We introduce hard negative re-pairing to test these uses (\cref{fig:teaser}). Random re-pairing removes original pairs and scrambles semantic organization at once, leaving both factors free to affect the diagnostics. Hard negative re-pairing separates them by assigning each image the caption of a different, similar image, chosen independently of the evaluated models. Representations and image--text scores stay fixed, and the pairing used to compute a diagnostic changes.
If relational diagnostics identified individual pairs, they would fall toward their random values. CKA stays close to its original value and exceeds that of random re-pairing that keeps 80\% of original pairs, although both annotators judge a majority of the substituted captions wrong. Top-one recall falls to about a tenth of its original value. High relational agreement therefore persists under similar, mostly wrong substitutions.

Why do the questions come apart? Our answer is scale. Two models can both organize skiing images and skiing captions near their respective neighbors, while confusing the details that distinguish one skiing scene from the next. Relational diagnostics pool over many neighbors, so they stay high under the broad agreement that re-pairing preserves. Matching and retrieval resolve the closest alternatives. We call the number of neighbors a diagnostic pools over its \emph{scale} (\cref{tab:diagnostics}). At a coarse scale, a high value shows agreement on the kind of scene and leaves the identity of the individual scene unresolved.

For CKA, a pair decomposition makes the distinction exact (\cref{sec:scale}).
CKA sums how well the two models agree on the ways that any two items differ. Near neighbors form a small share of the pairs, and they carry a smaller share of the sum.
The same decomposition gives the exact change from exchanging two captions. In the matching setting of \citet{maniparambil2024vision}, exchanging captions of near neighbors can raise the objective. The original pairing has improving swaps, so even an exact solver would prefer another pairing in the problems we test.

Scale also changes how we use the diagnostics. A resolution curve shows how dissimilar a substituted caption must be before the change registers. Diagnostics that pool over more neighbors retain agreement across a greater range of substitutions (\cref{sec:study-profile}).
It follows that the \emph{level} of a relational diagnostic, its value under the original pairing, mostly reflects coarse structure. Its \emph{response}, the share that hard negative re-pairing removes, depends on the exact pairs.
Across 22 trained retrievers, the level of CKA or neighborhood agreement correlates negatively with recall on COCO and positively on Flickr30k.
The response of neighborhood agreement correlates positively on every tested split (\cref{sec:study-selection}).
Before alignment, that response also ranks 36 encoder pairs by the retrieval they reach after training (\cref{sec:before-alignment}).

Our contributions build on the three questions.
\begin{itemize}[leftmargin=*,itemsep=1pt,topsep=3pt]
\item We separate marginal, relational and retrieval diagnostics by their inputs and scale. Invariance checks distinguish their inputs, and an exact decomposition explains the role of scale in CKA.
\item We introduce hard negative re-pairing, with human judgments of substituted captions and resolution curves with respect to caption similarity.
\item CKA favors complete re-pairing over a pairing that keeps most original pairs. Relational diagnostics keep most of their margin over random re-pairing while top-one retrieval collapses. Choosing by their level costs held-out recall in every selection comparison, even with positive correlation. The neighborhood agreement response tracks retrieval among trained models and ranks encoder pairs before alignment.
\end{itemize}

%% file: sec/2_related.tex
\section{Related work}
\label{sec:related}

\paragraph{Testing what a diagnostic tracks.}
\citet{kornblith2019similarity} identify transformations a similarity index should ignore.
\citet{ding2021grounding} test whether similarity follows model behavior, and \citet{davari2023reliability} show that CKA can change with little functional change.
Our invariance checks separate the inputs each diagnostic reads (\cref{sec:scale}). CKA stays high when similar captions replace every original caption.

\paragraph{What cross-modal agreement consists of.}
\citet{huh2024platonic} report rising agreement between stronger models.
\citet{groger2026aristotelian} find that neighborhood similarity survives null calibration more clearly than global spectral similarity. \citet{koepke2026back} find shared coarse semantics as datasets grow. Our intervention tests how that structure survives changes to the pairing.
Hard negatives, wrong captions that resemble the right one, serve as distractors in benchmarks such as SugarCrepe and Winoground~\citep{hsieh2023sugarcrepe,thrush2022winoground,yuksekgonul2023when}. We make them the assigned captions and measure Q2 and Q3 together.

\paragraph{Relational diagnostics for choosing models.}
Unimodal similarity has been used to compare encoders before alignment~\citep{groger2025structure,maniparambil2025harnessing}.
\citet{zhang2025assessing} point out that relational proxies do not directly assess individual image--text pairs and evaluate frozen encoders through a learned alignment layer.
We use a fixed-score intervention to measure that gap and evaluate proxies among trained retrievers and before alignment.

\paragraph{Locality and matching.}
The CKNNA of \citet{huh2024platonic} restricts centered kernel entry products to shared neighbors.
\citet{maniparambil2024vision} score a candidate image--caption pair by its local contribution after appending the queries to a shared base.
Our scale restriction selects unordered pairs separately in each modality and compares their full kernel-profile differences.
The objective change in \cref{prop:swap} is a two-exchange gain written for CKA~\citep{taillard1991robust}, the same quadratic-assignment objective that \citet{maniparambil2024vision} optimize approximately with FAQ~\citep{vogelstein2015fast}.
\citet{schnaus2025blind} formulate vision--language correspondence between class representations as quadratic assignment, including objectives based on CKA and mutual nearest neighbors.
Their globally optimal solutions often improve class matching. In our individual image--caption problems, the original pairing already has a lower objective than some local swaps (\cref{sec:study-cka}).

%% file: sec/3_setup.tex
\section{Setup and hard negative re-pairing}
\label{sec:methods}\label{sec:setup}
Our intervention changes the caption paired with each image.

\subsection{Evaluation setup}
We evaluate image representations $I=(v_1,\dots,v_n)$ and text representations $T=(t_1,\dots,t_n)$ under a one-to-one pairing $p$. Image $i$ receives its assigned caption $p(i)$.
A text unit is one caption in the single-caption setting or the mean of an image's caption embeddings in the caption-group setting~\citep{maniparambil2024vision}.
The original pairing is $p_0(i)=i$. Its fidelity to the dataset is
\begin{equation}
 F(p)=\frac{1}{n}\sum_{i=1}^n\mathbf{1}[p(i)=i].
 \label{eq:fidelity}
\end{equation}
Fidelity counts retained dataset pairs. COCO has unlabeled valid pairs~\citep{parekh2021crisscrossed,chun2022eccv}, so human labels assess whether substituted captions describe their images.

A complete re-pairing has fidelity zero. Every representation and image--text score $s(t,v)$ stays fixed.
Recall@$K$ is the fraction of images whose assigned caption is among the $K$ highest-scoring captions, or the reverse for text-to-image recall. We report both directions and their mean.

\subsection{Hard negative re-pairing}
\label{sec:semantic-repairing}\label{sec:semantic-assignment}
At fidelity $F$, we keep the same images fixed for the hard negative and random re-pairings.
The hard negative re-pairing gives each other image the original caption of a similar image, with similarity determined by an external selector.
Let $\mathcal I_F$ be the images to be re-paired. The feasible bijections $\mathcal P_F$ redistribute their captions within that set, excluding each image's own caption and any caption with identical normalized text. A bijection with no fixed points on $\mathcal I_F$ is a derangement. Let $z_i$ be the unit-normalized selector representation for the original item indexed by $i$. We solve
\begin{equation}
 p_F^\star\in\arg\min_{p\in\mathcal P_F}
    \sum_{i\in\mathcal I_F}(1-z_i^\top z_{p(i)}).
 \label{eq:semantic-assignment}
\end{equation}
One joint assignment over $\mathcal I_F$ enforces the one-to-one constraint. The random reference uses the same feasible edges (Supplement~A).

The default selector uses MPNet caption vectors~\citep{song2020mpnet,reimers2019sentencebert} and gives the MPNet re-pairing. An annotation selector uses COCO object categories. For the retrieval models, an image-side selector uses DINOv2 image vectors~\citep{oquab2023dinov2}.
To vary similarity at fixed fidelity, we cap admissible similarities and solve \cref{eq:semantic-assignment} at each cap. We call this series of re-pairings the similarity ladder.

Two annotators independently labeled 400 MPNet substitutions and 100 control pairs, blind to the condition, as valid, wrong detail, wrong content or unsure. Both wrong labels count as wrong, and estimates are weighted to the full re-pairing (Supp.~D). A model judge, Qwen3.8-27B, marks a caption wrong when two paraphrased prompts both return a wrong label. The model-judged re-pairing keeps only substitutions that it marks wrong.

\subsection{Diagnostics}
\label{sec:diagnostics}
Marginal diagnostics read one modality. Examples are local intrinsic dimension, neighbor radii, and effective rank (Supplement~A).
Neighborhood agreement is the mean fraction of an image's $k$ neighbors whose assigned captions are neighbors of its own assigned caption.
\begin{equation}
 A_k(p)=\frac{1}{nk}\sum_{i=1}^n
 \left|N_k^I(i)\cap p^{-1}\left(N_k^T(p(i))\right)\right|.
 \label{eq:agreement}
\end{equation}
We default to $k=50$. The expected overlap under independent uniform neighbor sets is $k/(n-1)$ (Supplement~A).

CKA uses kernels $\mathbf K,\mathbf L$ centered and normalized by their Frobenius norms. Define $\mathbf P_p$ so that $(\mathbf P_p\mathbf L\mathbf P_p^\top)_{ij}=L_{p(i)p(j)}$. Then
\begin{equation}
 \mathrm{CKA}(p)=\langle\mathbf K,\mathbf P_p\mathbf L\mathbf P_p^\top\rangle
 =\sum_{i,j}K_{ij}L_{p(i)p(j)}.
 \label{eq:cka}
\end{equation}
We use an RBF kernel with the modality's median squared distance as its bandwidth parameter. We name the linear variant explicitly.

CKA reads all $n-1$ other items, $A_k$ reads $k$ neighbors, and Recall@$K$ checks the top $K$ candidates.
We call scales of fifty or more coarse and scales of ten or fewer fine.
\Cref{tab:diagnostics} gives the inputs, scales and invariance checks.
\input{tab/diagnostics}

\subsection{Random reference and retention}
\label{sec:random-reference}
For a diagnostic $M$, let $M_{\rm ref}$ be its mean under random re-pairing at the same fidelity. Its margin is $M(p_0)-M_{\rm ref}$.
Retention $\rho_M$ is the fraction of that margin kept, and share lost $H_M$ is the rest.
\begin{equation}
 \rho_M(p)=\frac{M(p)-M_{\rm ref}}{M(p_0)-M_{\rm ref}},
 \qquad H_M(p)=1-\rho_M(p).
 \label{eq:retention}
\end{equation}
The level of a diagnostic is $M(p_0)$. Its response is its share lost under complete hard negative re-pairing.
We average recall directions before computing retention (random draws in Supp.~A).

\subsection{Models, data and protocol}
\label{sec:protocol}
We use COCO 2017 validation~\citep{lin2014microsoftcoco}, its disjoint halves A and B, and Flickr30k validation and test~\citep{young2014flickr30k}. Single-caption evaluation uses one seeded caption per image and one assigned caption per query. Its recall differs from the benchmark that credits any of five captions.
We use 25 seeded subsets per dataset for stability checks.
The caption-group setting reproduces the mean-pooled DINOv2 and averaged All-RoBERTa representations of \citet{maniparambil2024vision}.

The 22 retrievers include CLIP, SigLIP, SigLIP~2, E5-V, Jina-CLIP-v2, LongCLIP-L, LongCLIP-GmP and two Qwen3-VL-Embedding sizes.
Before alignment, we compare 36 encoder pairs from DINOv2 small, base and large, ViT base and large, and ConvNeXt large with E5 or BGE in three sizes each.
We compute proxies on COCO half A and fit linear or two-layer MLP projectors with a symmetric contrastive loss on COCO training data.
Evaluation uses COCO half B and Flickr30k test.

We report Kendall $\tau_b$ with bidirectional recall, the number of seeded subsets with positive $\tau_b$, and order changes between COCO and Flickr30k validation. An order change occurs when a model pair ranks differently across the two datasets. Regret is the held-out recall shortfall from the best evaluated model or pair. Its intervals use paired bootstraps (Supp.~G, H).

%% file: tab/diagnostics.tex
\begin{table*}[t]
\centering\footnotesize
\setlength{\tabcolsep}{3pt}
\caption{\textbf{The three questions differ in inputs, scale and invariance.}
(a) Diagnostics at coarse scales (fifty or more neighbors) and fine scales (ten or fewer). (b) Invariance checks for SigLIP B/16 224 on COCO (Supp.~B). $^{\dagger}$Raw vectors before normalization.}
\label{tab:diagnostics}
\label{tab:controls}
\begin{minipage}[t]{.45\textwidth}
\centering
\textbf{(a) Diagnostics by scale}\par\smallskip
\begin{tabularx}{\linewidth}{@{}p{.98in}YY@{}}
\toprule
Inputs & Coarse scale & Fine scale \\
\midrule
Q1. One modality ($I$ or $T$) & Effective rank & Local intrinsic dimension \\
\addlinespace
Q2. Both modalities and $p$ & CKA, $A_k$ at $k\ge 50$ & $\mathrm{CKA}_k$, $A_k$ at $k\le 10$, matching \\
\addlinespace
Q3. Scores and $p$ & Recall@$K$ at $K\ge 50$ & Recall@1 and Recall@10 \\
\bottomrule
\end{tabularx}
\end{minipage}\hfill
\begin{minipage}[t]{.52\textwidth}
\centering
\textbf{(b) Invariance checks}\par\smallskip
\begin{tabularx}{\linewidth}{@{}Yccc@{}}
\toprule
Check & Q1, marginal & \shortstack{Q2, relational\\($A_{50}$)} & \shortstack{Q3, retrieval\\(R@1)} \\
\midrule
Rotate one tower ($T$) & Fixed & Fixed & $-47.2$ points \\
\addlinespace
Re-pair at random ($p$) & Fixed & $-96.2\%$ & $-47.2$ points \\
\addlinespace
Rescale raw text ($\tilde T$) & Text LID $+45.8\%^{\dagger}$ & $-48.7\%^{\dagger}$ & Fixed \\
\addlinespace
Rotate both towers ($I,T$) & Fixed & Fixed & Fixed \\
\bottomrule
\end{tabularx}
\end{minipage}
\end{table*}

%% file: sec/4_scale.tex
\section{Why the three questions come apart}
\label{sec:scale}
\label{sec:exact-controls}
\label{sec:questions}
The three questions of \cref{sec:intro} differ in what a diagnostic must read.
Marginal diagnostics (Q1) read either $I$ or $T$.
Relational diagnostics (Q2) read $I$, $T$ and the pairing $p$.
Retrieval diagnostics (Q3) read rankings from the score matrix $S_{ij}=s(t_j,v_i)$ and use $p$ to identify the assigned captions.
\Cref{tab:diagnostics}b separates the three questions pairwise. Rotating one tower holds Q1 and Q2 fixed while lowering bidirectional Recall@1 by 47.2 points. Random re-pairing holds Q1 fixed while reducing neighborhood agreement by 96.2\%. Rescaling raw text vectors changes the diagnostics on raw geometry while preserving retrieval after normalization. Rotating both towers together leaves all three diagnostics unchanged. Supplement~B gives the exact invariances and replications. The CKA decomposition then explains why its usual scale gives little weight to near neighbors.

\subsection{CKA as a sum over item pairs}
At the original pairing let $\mathbf w_{ij}=\mathbf e_i-\mathbf e_j$, and let
$c_{ij}=\langle\mathbf K\mathbf w_{ij},\mathbf L\mathbf w_{ij}\rangle$.
The two vectors compare how differently images $i,j$ relate to all images and how differently their captions relate to all captions.
Their inner product measures agreement about the difference.

\begin{proposition}[Pair decomposition]
\label{prop:pairs}
For centered kernels with unit Frobenius norm, $\mathrm{CKA}(p_0)=n^{-1}\sum_{i<j}c_{ij}$.
\end{proposition}
\begin{proof}
$\sum_{i<j}\mathbf w_{ij}\mathbf w_{ij}^\top=n\mathbf I-\mathbf 1\mathbf 1^\top$.
Taking its trace product with $\mathbf K\mathbf L$ gives $n\tr(\mathbf K\mathbf L)$ since both kernels annihilate $\mathbf 1$.
\end{proof}
Up to a common factor, CKA averages contributions over all unordered item pairs. Unrelated items and near neighbors enter the same sum.

\subsection{Exchanging two captions}
\begin{proposition}[Swap identity]
\label{prop:swap}
Let $p_{ij}$ exchange the original captions of images $i$ and $j$.
For $d^I_{ij}=\mathbf w_{ij}^\top\mathbf K\mathbf w_{ij}$ and
$d^T_{ij}=\mathbf w_{ij}^\top\mathbf L\mathbf w_{ij}$,
\begin{equation}
 \mathrm{CKA}(p_{ij})-\mathrm{CKA}(p_0)=d^I_{ij}d^T_{ij}-2c_{ij}.
 \label{eq:swap}
\end{equation}
\end{proposition}
\begin{proof}
The swap matrix is $\mathbf I-\mathbf w_{ij}\mathbf w_{ij}^\top$.
Expanding $\langle\mathbf K,\mathbf P_{p_{ij}}\mathbf L\mathbf P_{p_{ij}}^\top\rangle$ gives the original inner product, two terms equal to $-c_{ij}$ by symmetry, and the product $d^I_{ij}d^T_{ij}$.
\end{proof}
The kernels are positive semidefinite, so the $d$ terms are squared kernel-space distances.
A swap improves the objective when $c_{ij}<d^I_{ij}d^T_{ij}/2$. The condition can hold even when $c_{ij}$ is positive. It also holds with fixed anchors, using the full kernels and restricting swaps to query items.

\subsection{CKA at scale \texorpdfstring{$k$}{k}}
For each modality form the unordered pairs $N_k$ where either item is among the other's $k$ nearest neighbors.
We take the cosine of the stacked kernel-profile differences for these pairs.
\begin{equation}
 \mathrm{CKA}_k(N_k)=
 \frac{\sum_{ij\in N_k}c_{ij}}
 {\sqrt{\sum_{ij\in N_k}\|\mathbf K\mathbf w_{ij}\|^2
             \sum_{ij\in N_k}\|\mathbf L\mathbf w_{ij}\|^2}}.
 \label{eq:cka-scale}
\end{equation}
We average the values from the two neighbor sets.
Under re-pairing the text set moves with the text kernel. The image set stays fixed.
When both sets contain all pairs, each sum becomes $n$ times its corresponding centered inner product. The resulting value is ordinary CKA.

%% file: sec/5_experiments.tex
\section{Experiments and results}
\label{sec:experiments}

\subsection{CKA and fidelity}
\label{sec:study-cka}
\begin{figure*}[t]
\centering\includegraphics[width=\textwidth]{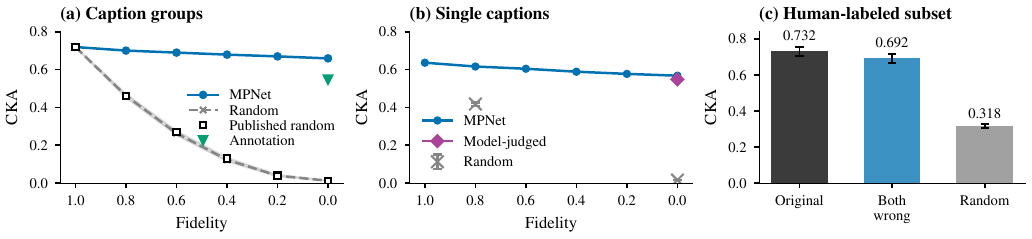}
\caption{\textbf{CKA stays high after complete hard negative re-pairing.}
(a) Caption groups, with the published random curve. (b) Single captions. The diamond marks the re-pairing whose captions the model judge rejects. (c) The 195 pairs both annotators judge wrong, with original captions, substituted captions and random re-pairings of the same caption pool (intervals: Supp.~D).}
\label{fig:cka-curve}
\end{figure*}

If CKA identified individual pairs, it would order pairings by fidelity.
\citet{maniparambil2024vision} match images to captions by maximizing CKA. We reproduce their curve with mean-pooled DINOv2 image features and averaged All-RoBERTa caption embeddings.
Complete MPNet re-pairing scores 0.66, above 0.46 for random re-pairing that keeps 80\% of the pairs.
We call this a \emph{fidelity reversal} (\cref{fig:cka-curve}). It also occurs in the single-caption setting, across 128 alternative re-pairings, with a linear kernel, with row-normalized representations, and with the annotation selector, which uses COCO categories alone in the caption-group setting (Supp.~C).

Two human votes identify the substitutions that clearly contradict their images.
Both annotators call a weighted 61\% of MPNet substitutions wrong (\cref{tab:human}).
Both reject 3 of the 50 original control pairs.
On the pairs with two wrong votes, CKA retains 90\% of its margin over random re-pairing of the same swapped caption pool (\cref{fig:cka-curve}c).
Coarse agreement thus persists on pairs that both annotators reject.

A model judge extends this check to the full re-pairing, and the reversal holds when only substitutions it marks wrong are kept (\cref{fig:cka-curve}b).
The judge reaches 87\% precision against LLM-adjudicated labels, below the 90\% target, and marks 11\% of original captions wrong, above the 5\% ceiling. The claim rests on the human votes (Supp.~D).

\input{tab/human}

\begin{figure*}[t]
\centering\includegraphics[width=\textwidth]{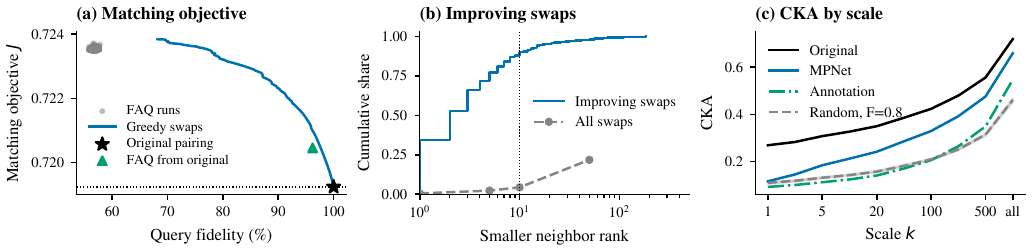}
\caption{\textbf{Near-neighbor swaps improve the matching objective while contributing little to all-pairs CKA.}
(a) Solver outputs and greedy swaps score above the original pairing in one problem. (b) Cumulative share of swaps by neighbor rank. (c) Caption-group CKA by scale. With fewer neighbors, the MPNet re-pairing falls further below the original pairing (gray shows random re-pairings at fidelity 0.8).}
\label{fig:matching}
\end{figure*}

The swap identity tests whether the solver or the objective limits matching.
In each of three matching problems in the published setting, 126 to 135 of the 124,750 possible swaps raise the objective at the original pairing (Supp.~C).
Greedy swaps raise it further as fidelity falls (\cref{fig:matching}a).
All 1,500 stored FAQ outputs score above the original pairing~\citep{vogelstein2015fast}.
An exact maximizer must therefore return another pairing on these problems.

Improving swaps are concentrated among near neighbors.
An unordered pair qualifies if either item is among the other's ten nearest neighbors in at least one modality.
About nine in ten improving swaps qualify, compared with 4.4\% of all candidate swaps (\cref{fig:matching}b).
Pairs within ten caption neighbors make up 1.7\% of all pairs and carry 0.26\% of all-pairs CKA. Image neighbors give a similar share (Supp.~C).
Most improving swaps still have a positive contribution $c_{ij}$. The positive value falls below the threshold in \cref{eq:swap}.

CKA restricted to fewer neighbors responds more strongly.
For complete MPNet re-pairing of caption groups, retention falls from 0.92 over all pairs to 0.40 at the nearest neighbor (\cref{fig:matching}c).
At fine scales the fidelity ordering returns for two re-pairings. The annotation re-pairing falls below the random re-pairings at fidelity 0.8 at fifty neighbors or fewer. The model-judged re-pairing in the single-caption setting falls below them at the nearest neighbor. The MPNet re-pairing stays above them at every scale.

\subsection{Relational diagnostics versus retrieval}
\label{sec:study-profile}
Do relational diagnostics and retrieval respond alike to the same re-pairing? We apply complete MPNet re-pairing to the 22 retrievers. It preserves broad content. Nine in ten substituted captions come from an image sharing a COCO category with the current image, compared with about one in three under random re-pairing.

\begin{figure*}[t]
\centering\includegraphics[width=\textwidth]{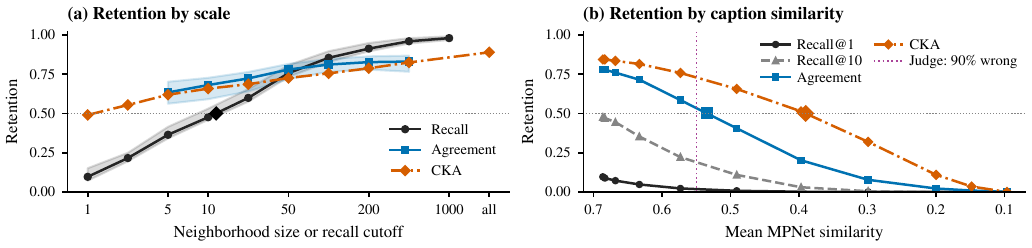}
\caption{\textbf{Retention rises with scale and depends on caption similarity.}
(a) Median and range over 22 COCO retrievers, with CKA for DINOv2 and All-RoBERTa. (b) Medians along the similarity ladder.}
\label{fig:resolution}
\end{figure*}

Retention rises as the diagnostic reads more neighbors (\cref{fig:resolution}a).
After complete re-pairing, neighborhood agreement keeps about three quarters of its margin, Recall@10 about half, and Recall@1 about a tenth.
The ordering holds across the 22 models on COCO. The ordering also holds at a matched scale: with ten neighbors, neighborhood agreement retains more than Recall@10 in all 22 models (Supp.~D).
From about a hundred candidates on, recall retains more than either diagnostic at the same scale.

For SigLIP B/16 224, image-to-text Recall@1 falls from 47\% to 4\%, while neighborhood agreement retains 78\% of its margin.

The similarity ladder measures how far a caption must move before a diagnostic responds.
Define \emph{resolution} as the half-retention similarity.
At the top of the COCO ladder, Recall@1 already retains less than half its margin for each model.
The median halfway points are at similarity 0.53 for neighborhood agreement and 0.39 for CKA (\cref{fig:resolution}b and Supp.~D).
A diagnostic at a coarser scale thus needs a less similar caption before it responds.

The original human votes support this ordering on sampled substitutions.
On pairs both annotators judge wrong, re-pairing lowers median bidirectional Recall@1 over the 22 models from 56.0\% to 2.5\%.
Neighborhood agreement keeps about three quarters of its value (\cref{tab:human}).
Pairs on which both annotators identify wrong content have lower retrieval than pairs on which both identify a wrong detail. Supplement~D.8 shows labeled examples and reports weighted results by label.

Fixed rankings still allow substituted captions to rank first, so the recall collapse is an empirical result. For SigLIP B/16 224, among queries whose original caption falls below rank one, the substituted caption ranks first for only 7.75\% of image queries and 9.46\% of text queries.
The retention ordering also holds for the late-interaction and audio--text configurations in Supplement~F.

\subsection{Model selection by a relational diagnostic or by recall}
\label{sec:study-selection}
A diagnostic used to compare models should rank them as retrieval does. The level mostly reflects coarse structure, which survives re-pairing (\cref{sec:study-profile}). The response is the part that depends on the exact pairs. We compare the two on COCO, its two halves and the two Flickr30k splits.

\begin{figure*}[t]
\centering\includegraphics[width=\textwidth]{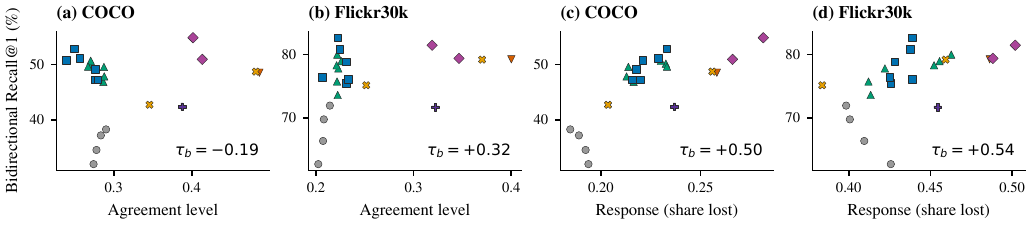}
\caption{\textbf{The response has the same sign on both datasets.}
Each point is one of 22 models. Panels compare Recall@1 with neighborhood agreement level (a, b) and response to complete MPNet re-pairing (c, d). Marker shape indicates model family.}
\label{fig:selection}
\end{figure*}

The level changes sign across datasets.
Neighborhood agreement correlates with bidirectional Recall@1 at $\tau_b=-0.19$ on COCO and $+0.32$ on Flickr30k validation. CKA shows the same sign change.
The level of $A_{50}$ changes more than twice as many cross-dataset orders as recall itself (\cref{fig:selection,tab:correlations}).

\input{tab/correlations}

A finer scale does not repair the level. At each of the ten tested scales, RBF CKA correlates negatively with recall on full COCO and positively on Flickr30k validation (\cref{tab:correlations} gives ten neighbors, and Supp.~E gives the full profile).
The variants spanning kernels, graphs and nearby scales choose the same model.
Its COCO regret is about six points on either half (\cref{tab:picks}).

The response of neighborhood agreement correlates positively with recall on all five splits and in 49 of the 50 subsets.
At fifty neighbors the cross-dataset order changes are comparable to recall's.
The association survives image-side and annotation selectors on their evaluated splits, and it also appears for the response of CKA (\cref{tab:correlations}).
Within SigLIP and SigLIP~2, the response is positive on every split. The $A_{50}$ level is negative on nine of these ten family--split combinations. In the four CLIP models, the level orders the models as recall does on every split, while the response has mixed signs (Supp.~E).
At ten neighbors the response generally correlates more weakly and changes more cross-dataset orders (\cref{tab:correlations}). At fifty neighbors the sixteen-model set is positive in only 18 of 25 Flickr30k subsets, short of the 23 we had required. The full set passes.

The response stays positively associated with recall after controlling for how much each text tower's own neighborhoods change under re-pairing. The sign change of the level traces mostly to the text tower. Measured against fixed reference encoders, the text side accounts for most cross-dataset order changes, and the sign change persists in the person-content and caption-length checks. Its cause remains open (Supp.~B and E).

Model selection puts these differences into units of recall (\cref{tab:picks}).
Among the 16 CLIP, SigLIP and SigLIP~2 models on Flickr30k, the level of neighborhood agreement correlates positively with recall and still picks a model 5.75 points below the pick of validation recall (95\% interval $[3.80,7.80]$).
In every comparison of the table, the level's pick trails the pick of validation recall, by 1.3 to 20.4 points.
The MPNet response picks the best model on both COCO halves in both model sets.
Its pick changes with the selector, and in the 22-model set its Flickr30k gain over the level has an interval spanning zero.
A diagnostic's use for model selection must therefore be judged by the held-out retrieval of its pick. Correlation alone is insufficient.
\input{tab/picks}

\subsection{Encoder selection before alignment}
\label{sec:before-alignment}
Choosing encoders before fitting projectors calls for a proxy.
Following the learned alignment evaluation of \citet{zhang2025assessing}, we compare proxies measured before training with retrieval after projector training.

The MPNet response ranks the encoder grid better than the level of neighborhood agreement (\cref{tab:proxies}).
Its correlations span 0.65--0.69, ahead of unimodal quality rank sums, parameter counts and the short training probe.
Unimodal quality mainly orders text encoders when vision is held fixed (Supplement~H).

Before alignment, CKA at ten neighbors also ranks pairs better than all-pairs CKA in every evaluation (\cref{tab:proxies}).
Ranking the grid and picking its best pair are different tasks. The response's pick trails the best pair by about one to three points, and with one encoder removed the worst regret of CKA at ten neighbors is 6.98 points, against 4.52 for all-pairs CKA and 4.02 for the response (\cref{tab:proxies} and Supp.~H).

\input{tab/proxies}

%% file: tab/human.tex
\begin{table}[t]
\centering\small
\caption{\textbf{Human judgments separate retrieval from neighborhood agreement.}
The wrong rates are reweighted to the full MPNet pairing. The retrieval rows give medians of weighted means over 22 models. Recall is bidirectional (\%).}
\label{tab:human}
\begin{tabular}{@{}lrrrr@{}}
\toprule
Votes and pairing & $n$ & R@1 & R@10 & $A_{50}$ \\
\midrule
\multicolumn{3}{@{}l}{Both vote wrong, weighted} & \multicolumn{2}{r@{}}{61.2\%} \\
\multicolumn{3}{@{}l}{Either votes wrong, weighted} & \multicolumn{2}{r@{}}{65.7\%} \\
\midrule
Both wrong, original & 195 & 56.0 & 86.0 & .256 \\
Both wrong, re-paired & 195 & 2.5 & 29.6 & .197 \\
Other, original & 205 & 39.4 & 79.4 & .315 \\
Other, re-paired & 205 & 9.2 & 56.0 & .260 \\
\bottomrule
\end{tabular}
\end{table}

%% file: tab/correlations.tex
\begin{table*}[t]
\centering\small
\caption{\textbf{The response tracks recall across datasets.} Kendall $\tau_b$ with bidirectional Recall@1 for 22 models. COCO+ and Flickr+ count positive subsets out of 25. Changes count model pairs ordered differently across datasets, out of 231.}
\label{tab:correlations}
\begin{tabular}{@{}lrrrrrrrr@{}}
\toprule
Diagnostic & COCO & Half A & Half B & Flickr val. & Flickr test & COCO $+$ & Flickr $+$ & Changes \\
\midrule
CKA level & $-0.40$ & $-0.33$ & $-0.39$ & $+0.22$ & $+0.27$ & 0 & 25 & 74 \\
$\mathrm{CKA}_{10}$ level & $-0.13$ & $-0.01$ & $-0.13$ & $+0.42$ & $+0.47$ & 4 & 25 & 67 \\
$A_{50}$ level & $-0.19$ & $-0.17$ & $-0.25$ & $+0.32$ & $+0.30$ & 0 & 25 & 60 \\
$A_{10}$ level & $+0.11$ & $+0.07$ & $-0.03$ & $+0.26$ & $+0.23$ & 0 & 25 & 31 \\
\midrule
$A_{50}$ response, MPNet & $+0.50$ & $+0.68$ & $+0.65$ & $+0.54$ & $+0.46$ & 25 & 24 & 25 \\
$A_{50}$ response, image-side & $+0.43$ & $+0.52$ & $+0.60$ & $+0.45$ & $+0.56$ & -- & -- & 29 \\
$A_{50}$ response, annotation & $+0.51$ & $+0.61$ & $+0.52$ & -- & -- & -- & -- & -- \\
CKA response, MPNet & $+0.75$ & $+0.75$ & $+0.74$ & $+0.53$ & $+0.38$ & -- & -- & -- \\
$A_{10}$ response, MPNet & $+0.63$ & $+0.41$ & $+0.40$ & $+0.26$ & $+0.26$ & 25 & 25 & 68 \\
Recall@1 itself & -- & -- & -- & -- & -- & -- & -- & 24 \\
\bottomrule
\end{tabular}
\end{table*}

%% file: tab/picks.tex
\begin{table*}[t]
\centering\footnotesize\setlength{\tabcolsep}{3pt}
\caption{\textbf{Choosing by the level costs held-out recall. The response's pick depends on the selector.} Each entry gives the selected model, held-out bidirectional Recall@1 (\%), and regret (points). A $\to$ B selects on COCO half A and evaluates on half B. Qwen is Qwen3-VL-Embedding.}
\label{tab:picks}
\begin{tabular}{@{}lllll@{}}
\toprule
Models & Selection criterion & COCO A $\to$ B & COCO B $\to$ A & Flickr val. $\to$ test \\
\midrule
22 & Validation recall & Qwen 8B (60.78; 0.00) & Qwen 8B (65.28; 0.00) & SigLIP L/16 384 (80.95; 0.00) \\
22 & $A_{50}$ level; CKA level, all scales & E5-V (54.86; 5.92) & E5-V (59.26; 6.02) & E5-V (78.85; 2.10) \\
22 & $A_{50}$ response, MPNet & Qwen 8B (60.78; 0.00) & Qwen 8B (65.28; 0.00) & Qwen 8B (80.40; 0.55) \\
22 & $A_{50}$ response, image-side & E5-V (54.86; 5.92) & E5-V (59.26; 6.02) & E5-V (78.85; 2.10) \\
22 & $A_{50}$ response, annotations & Qwen 8B (60.78; 0.00) & LongCLIP-GmP (58.56; 6.72) & -- \\
\midrule
16 & Validation recall & SigLIP L/16 384 (59.84; 0.00) & SigLIP L/16 384 (62.22; 0.00) & SigLIP L/16 384 (80.95; 0.00) \\
16 & $A_{50}$ level & CLIP L/14 336 (45.00; 14.84) & CLIP L/14 336 (46.92; 15.30) & SigLIP B/16 256 (75.20; 5.75) \\
16 & CKA level, all pairs & CLIP B/32 224 (39.48; 20.36) & CLIP B/16 224 (43.80; 18.42) & SigLIP L/16 256 (79.65; 1.30) \\
16 & $A_{50}$ response, MPNet & SigLIP L/16 384 (59.84; 0.00) & SigLIP L/16 384 (62.22; 0.00) & SigLIP 2 L/16 384 (78.35; 2.60) \\
\bottomrule
\end{tabular}
\end{table*}

%% file: tab/proxies.tex
\begin{table}[t]
\centering\small
\caption{\textbf{Before alignment, fine-scale CKA and the response rank encoder pairs.} Kendall $\tau_b$ with post-training retrieval (36 pairs, ranges over eight evaluations). Regret in recall points. Worst is the largest regret after removing one encoder.}
\label{tab:proxies}
\begin{tabular}{@{}lrrr@{}}
\toprule
Proxy & $\tau_b$ & Regret & Worst \\
\midrule
MPNet $A_{50}$ response & 0.65--0.69 & 1.2--2.7 & 4.0 \\
$\mathrm{CKA}_{10}$ & 0.68--0.74 & 0.0--1.1 & 7.0 \\
All-pairs CKA & 0.53--0.60 & 0.0--1.1 & 4.5 \\
One-epoch probe & 0.41--0.45 & 0.0--3.3 & 4.8 \\
Unimodal quality & 0.21--0.26 & 0.0--0.9 & 12.7 \\
$A_{50}$ level & 0.07--0.13 & 0.0--0.9 & 12.7 \\
\bottomrule
\end{tabular}
\end{table}

%% file: sec/6_discussion.tex
\section{Discussion and conclusion}
\label{sec:conclusion}

Do vision and language models represent the world alike? CKA and neighborhood agreement answer at a coarse scale. High values show broad agreement between the two models, and most of their margin survives replacing every caption with a similar one. Human annotators judge most substitutions wrong. Identifying the original pairs depends on separating the nearest alternatives. Coarse semantics have been observed under dataset scaling~\citep{koepke2026back}, and neighborhood agreement survives calibration~\citep{groger2026aristotelian}. We show what survives an intervention on the pairing and explain the effect through the CKA decomposition.

\noindent\textbf{Limitations.} The main experiments use COCO and Flickr30k with short captions. The audio, video and dense-caption extensions test retention ordering (Supp.~F). Human labels cover the MPNet re-pairing. The response depends on the selector and neighborhood size (\cref{sec:study-selection}).

Two consequences follow. Maximizing CKA for matching~\citep{maniparambil2024vision} favors near-neighbor swaps over the original pairing in the tested problems. An exact solver therefore prefers another pairing. The level of a relational diagnostic gives inconsistent retrieval rankings across datasets, and choosing by it costs up to twenty points of held-out recall. The response to re-pairing tracks retrieval among trained models and ranks encoder pairs before alignment. We therefore recommend choosing trained models by held-out recall, comparing encoder pairs before alignment by the response to re-pairing, and testing matching and alignment claims against hard negative re-pairing at matched fidelity, with resolution reported. Every diagnostic should be reported with its question and its scale.

%% file: sec/8_llm_statement.tex
\section*{LLM Usage Statement}
Generative AI tools assisted with experimental planning, code, derivations and numerical checks, literature discovery, interpretation, figures, and manuscript drafting and editing. Anthropic Claude assisted with revising the manuscript. OpenAI Codex assisted with the revision and analyses. Qwen3.8-27B produced the model judgments. OpenAI ChatGPT adjudicated the 52 disagreements between the human annotators for judge calibration. The human-only analyses use the original independent votes. The authors are responsible for the submission.

%% file: sec/X_suppl.tex
\clearpage
\onecolumn
\nolinenumbers
\begingroup
\renewcommand{\twocolumn}[1][]{#1}
\maketitlesupplementary
\endgroup
\label{page:supp-start}
\appendix
\counterwithin{figure}{section}
\counterwithin{table}{section}
\counterwithin{equation}{section}
\setcounter{proposition}{0}
\renewcommand{\theproposition}{S\arabic{proposition}}
\renewcommand{\theHproposition}{supp.\arabic{proposition}}
\input{sec/supp/a_details}
\input{sec/supp/a_followup}
\FloatBarrier
\input{sec/supp/b_controls}
\input{sec/supp/moved_invariance}
\input{sec/supp/b_followup}
\FloatBarrier
\input{sec/supp/c_cka}
\input{sec/supp/c_followup}
\FloatBarrier
\input{sec/supp/d_retrieval}
\input{sec/supp/d_followup}
\FloatBarrier
\input{sec/supp/e_selection}
\input{sec/supp/e_followup}
\FloatBarrier
\input{sec/supp/f_extensions}
\FloatBarrier
\input{sec/supp/g_reproduction}
\input{sec/supp/g_followup}
\FloatBarrier
\input{sec/supp/h_grid}
\clearpage

%% file: sec/supp/a_details.tex
\counterwithin{table}{section}
\counterwithin{figure}{section}
\counterwithin{equation}{section}
\makeatletter
\@ifundefined{lemma}{\newtheorem{lemma}{Lemma}[section]}{}
\@ifundefined{corollary}{\newtheorem{corollary}[lemma]{Corollary}}{}
\makeatother
\newcommand{\apprestatedname}{Proposition}
\newtheorem*{apprestatedinner}{\apprestatedname}
\newenvironment{apprestated}[2]
  {\renewcommand{\apprestatedname}{#1}\begin{apprestatedinner}[#2]}
  {\end{apprestatedinner}}
\providecommand{\apref}{\overline{M}_{\mathrm{ref}}}
\providecommand{\apret}[1]{\rho_{#1}}
\providecommand{\apcorr}[1]{A_{#1}^{\mathrm{corr}}}
\providecommand{\apId}{\mathrm{Id}}

\section*{Appendix overview}
\label{app:overview}
\addcontentsline{toc}{section}{Appendix overview}

The appendices follow the order of the main text. \Cref{app:details} fixes
conventions, marginal estimators, the chance correction, and the construction of
assignments and random references. \Cref{app:controls} proves
Propositions~\ref{prop:orthogonal}--\ref{prop:radial} and reports the invariance checks. \Cref{app:cka,app:retrieval,app:selection}
support the three studies of \autoref{sec:experiments}, \Cref{app:extensions} reports
the operator and modality extensions, and \Cref{app:reproduction} records
protocols, environments, and artifacts. The table below links each main-text
statement to its support. Throughout, \emph{one random draw} denotes a single
random re-pairing, and the \emph{random reference} \(\apref\) denotes the mean
over many draws that enters \autoref{eq:retention}.

\begin{center}
\small
\textbf{Where each main-text statement is supported}\par\smallskip
\setlength{\tabcolsep}{3.5pt}
\begin{tabularx}{\linewidth}{@{}p{0.19\linewidth}Yp{0.33\linewidth}@{}}
\toprule
Main text & Statement & Appendix support \\
\midrule
\autoref{sec:semantic-repairing} & Alternative matchings; annotation-based selectors &
\Cref{app:assignment}; \Cref{app:tab:selectors,app:tab:alternatives} \\
\autoref{sec:diagnostics} & LID, neighbor radius, effective rank &
\Cref{app:marginal} \\
\autoref{sec:diagnostics} & Chance level \(k/(n-1)\); corrected reporting &
\Cref{app:lem:chance}; \autoref{app:eq:corrected}; \Cref{app:cor:corrected-retention} \\
\autoref{sec:random-reference} & Random reference and its draws & \Cref{app:reference};
\Cref{app:tab:random-sets} \\
\autoref{sec:questions} & Proofs of Propositions~\ref{prop:orthogonal}--\ref{prop:radial} & \Cref{app:proofs} \\
\autoref{sec:questions}, \autoref{tab:controls} & Control construction; CLIP and SigLIP 2
replications & \Cref{app:control-construction,app:control-results};
\Cref{app:tab:control-responses,app:tab:control-replication} \\
\autoref{sec:experiments} & Checkpoints, preprocessing, caption selection &
\Cref{app:tab:cka-inputs,app:tab:roster}; \Cref{app:retrieval-protocol,app:flickr-protocol} \\
\autoref{sec:study-cka}, \autoref{fig:cka-curve} & Fidelity curve & \Cref{app:cka-curve};
\Cref{app:tab:cka-curve} \\
\autoref{sec:study-cka} & Annotation-only assignment & \Cref{app:cka-annotation} \\
\autoref{sec:study-cka} & 128 alternatives; linear CKA; row normalization &
\Cref{app:tab:cka-alternatives,app:tab:cka-variants} \\
\autoref{sec:study-cka}, \autoref{fig:qap} & QAP setup and outcomes & \Cref{app:qap};
\Cref{app:tab:qap} \\
\autoref{sec:study-profile} & Protocol; one assigned caption per query &
\Cref{app:retrieval-protocol} \\
\autoref{sec:study-profile}, \autoref{fig:22-ckpt} & Per-model results for all 22 &
\Cref{app:tab:panel,app:tab:six-reconciliation} \\
\autoref{sec:study-profile}, \autoref{fig:retention-example} & Representative values; random reference &
\Cref{app:tab:representative,app:tab:figure5} \\
\autoref{sec:study-profile} & Conditional R@1 of 7.75\% and 9.46\% &
\Cref{app:conditional}; \Cref{app:tab:conditional-ranks} \\
\autoref{sec:study-profile} & Retrieval controls and sensitivity analyses &
\Cref{app:retrieval-sensitivity}; \Cref{app:tab:coverage} \\
\autoref{sec:study-profile} & Late interaction; audio &
\Cref{app:tab:late-interaction,app:tab:audio-r1,app:tab:audio-r10} \\
\autoref{sec:study-selection}, \autoref{tab:flickr30k-checkpoint-selection} & Flickr30k protocol, panel, bootstrap interval &
\Cref{app:flickr-protocol}; \Cref{app:tab:flickr-panel,app:tab:flickr-results} \\
\bottomrule
\end{tabularx}
\end{center}

\section{Additional diagnostic and assignment details}
\label{app:details}

This appendix fixes the conventions used by all later appendices. It gives the
marginal estimators and the chance correction for neighborhood agreement
(\autoref{sec:diagnostics}), shows that the correction leaves retention unchanged
(\autoref{sec:random-reference}), and specifies how semantic assignments and random references
are constructed (\autoref{sec:semantic-repairing}).

\subsection{Notation, representation surfaces, and retrieval conventions}
\label{app:notation}

We use the evaluation state \(\mathcal E=(I,T,p,s)\) of \autoref{sec:setup}. The image
and text representations are \(I=(v_1,\ldots,v_n)\) and
\(T=(t_1,\ldots,t_n)\); the bijection \(p:[n]\to[n]\) assigns image \(i\) its
assigned caption \(p(i)\); and \(s\) is the scoring function. The score
matrix \(S_{ij}=s(t_j,v_i)\) has image rows and text columns. The original
pairing is \(p_0(i)=i\), and fidelity \(F(p)\) is defined in
\autoref{eq:fidelity}.

\paragraph{Retrieval directions.}
Image-to-text retrieval ranks row \(i\) of \(S\) for image query \(v_i\); its
assigned caption is \(t_{p(i)}\). Text-to-image retrieval ranks column \(j\)
for text query \(t_j\); its assigned caption is \(v_{p^{-1}(j)}\).
Recall@\(K\) (R@\(K\)) is the fraction of queries whose assigned caption is
among the first \(K\) candidates, and score ties are broken by candidate index.
Bidirectional R@\(K\) is the arithmetic mean of the two directions. Because
re-pairing leaves \(S\) unchanged, every query has the same candidate ranking
under every pairing; only its assigned caption changes. For bidirectional
recall, the two directional recalls are averaged within each condition before
\autoref{eq:retention} is applied, so an aggregate retention is not the mean of the
two directional retentions.

\paragraph{Representation surfaces.}
A surface is the exact array consumed by a diagnostic, including pooling and
normalization. For a global encoder, projected vectors \(\tilde v_i,\tilde t_i\)
precede normalization, and the unit-normalized vectors
\(v_i=\nu(\tilde v_i)\), \(t_i=\nu(\tilde t_i)\), with \(\nu(x)=x/\lVert
x\rVert_2\), are the vectors consumed by the native score. The retrieval
studies (\Cref{app:retrieval}) compute neighborhood agreement and recall on
normalized vectors. The invariance checks (\Cref{app:controls}) evaluate the
marginal and relational diagnostics on both surfaces. The CKA study
(\Cref{app:cka}) uses the saved rows of the published protocol without further
normalization, and \Cref{app:cka-variants} repeats it on row-normalized rows.
Absolute LID is not compared across surfaces or encoders.

\paragraph{Text units.}
A text unit is one selected caption in the retrieval and selection studies and
one caption group, all captions of an image averaged, in the CKA study
(\Cref{app:cka-protocol}).

\subsection{Marginal diagnostics}
\label{app:marginal}

Marginal diagnostics describe one modality at a time on a declared surface.
They use \(I\) or \(T\) alone, never \(p\) or \(S\), and they are not combined
into a scalar alignment score.

\paragraph{Local intrinsic dimensionality.}
At each \(k\in\{10,20,50,100\}\), we use the maximum-likelihood estimator of
\citet{levina2005maximumlikelihood} as implemented in scikit-dimension
\citep{skdim2021}, with its unbiased option and no coordinate
standardization. Our implementation follows LayerScope \citep{arcosholzinger2026layerscopelayerwisecharacterizationvideo},
which builds on GRIDS \citep{arcosholzinger2026grids}. Near-duplicate rows are grouped at absolute tolerance
\(10^{-10}\). For a row requiring direct resolution, all members of its
near-duplicate group are excluded, the first \(k\) positive Euclidean distances
\(0<r_1\le\cdots\le r_k\) are retained, and
\begin{equation}
 \widehat{\mathrm{LID}}_i(k)=\frac{k-2}{\sum_{j=1}^{k}\log(r_k/r_j)}.
 \label{app:eq:lid}
\end{equation}
The reported cohort value is the harmonic mean of the finite positive row
estimates. The raw and unresolved invalid fractions and the number of
recomputed rows are stored, and a condition is rejected if more than 5\% of rows
remain invalid. The estimate describes finite-neighborhood expansion; it is not
a claim about a unique manifold dimension.

\paragraph{Neighbor radius.}
The \(k\)-nearest-neighbor radius of row \(i\) is the Euclidean distance to its
\(k\)th neighbor, excluding the row itself. Neighbors are ordered by distance
and then by row index, which makes boundary ties deterministic. The retained
summary contains every row's radius and its mean, median, standard deviation,
minimum, maximum, and 5th, 25th, 75th, and 95th percentiles.

\paragraph{Effective rank of the covariance spectrum.}
For representations \(Z\in\mathbb R^{n\times d}\) centered across samples,
\(Z_c=Z-\mathbf 1\bar z^\top\) with \(\bar z=n^{-1}\sum_i z_i\), let
\(\sigma_\ell\) be the singular values of \(Z_c\), so that
\(\sigma_\ell^2/n\) are the eigenvalues of the covariance matrix. Following
\citet{roy2007effective},
\begin{equation}
 q_\ell=\frac{\sigma_\ell^2}{\sum_m\sigma_m^2},\qquad
 H(q)=-\sum_\ell q_\ell\log q_\ell,\qquad
 r_{\mathrm{eff}}=\exp H(q).
 \label{app:eq:effective-rank}
\end{equation}
Spectral entropy equals \(\log r_{\mathrm{eff}}\) and is therefore not treated
as a separate diagnostic. The implementation also archives participation rank,
which is not a reported result.

\paragraph{What the invariance checks fix.}
A one-sided rotation leaves every marginal diagnostic fixed while native
retrieval collapses, and positive radial scaling changes pre-normalization
marginal geometry while normalization restores the scored vectors exactly
(\Cref{app:control-results}). LID, neighbor radius, and effective rank therefore
describe the geometry of one modality; none of them establishes relational
agreement or retrieval performance. Pair-dependent intrinsic-dimension
correlation \citep{basile2025idcor} consumes the pairing and is therefore not a
marginal diagnostic in this sense, and neural intrinsic-dimension estimates need
not recover a single underlying manifold dimension
\citep{schulte2026rethinking}.

\subsection{Neighborhood agreement: chance level and corrected reporting}
\label{app:agreement}

Neighborhood agreement \(A_k(p)\) is defined in \autoref{eq:agreement}. The sets
\(N^I_k(i)\) and \(N^T_k(j)\) contain the \(k\) nearest Euclidean neighbors of
image \(i\) and text unit \(j\) within their own modality, exclude the item
itself, and break distance ties by index. Image \(j\) belongs to
\(p^{-1}(N^T_k(p(i)))\) exactly when \(t_{p(j)}\) is a neighbor of
\(t_{p(i)}\), so
\begin{equation}
 A_k(p)=\frac{1}{nk}\sum_{i=1}^{n}\sum_{j\in N^I_k(i)}
 \mathbf 1\!\left[p(j)\in N^T_k(p(i))\right].
 \label{app:eq:agreement-sum}
\end{equation}

\begin{lemma}[Chance level]
\label{app:lem:chance}
Fix neighbor sets \(N^I_k(i)\) and \(N^T_k(j)\) of size \(k\) that exclude
their own index, and let \(p\) be drawn uniformly from all bijections
\([n]\to[n]\). Then \(\mathbb E[A_k(p)]=c_k=k/(n-1)\).
\end{lemma}
\begin{proof}
Fix \(i\) and condition on \(p(i)=a\). Every \(j\in N^I_k(i)\) differs from
\(i\), so \(p(j)\) is uniformly distributed on \([n]\setminus\{a\}\). This set
contains all \(k\) elements of \(N^T_k(a)\) because \(a\notin N^T_k(a)\).
Hence \(\Pr[p(j)\in N^T_k(a)\mid p(i)=a]=k/(n-1)\) for each of the \(k\)
neighbors \(j\), and the conditional expected inner sum in
\autoref{app:eq:agreement-sum} is \(k^2/(n-1)\). This does not depend on \(a\), so
it is also the unconditional expectation. Averaging over \(i\) and dividing by
\(k\) gives \(k/(n-1)\).
\end{proof}

The lemma concerns the unrestricted uniform distribution over all bijections,
including those with fixed points. The random references used in this paper
exclude fixed points in the reassigned subset and exclude replacements with
identical normalized text, and their generator is not uniform
(\Cref{app:reference}). We therefore use \(c_k\) as the analytic reference and
report the empirical reference distributions separately. They agree closely: for
SigLIP B/16 224 at \(k=50\) and \(n=5{,}000\), \(c_{50}=0.010002\), while the
mean raw \(A_{50}\) over the 256 random re-pairings of the model panel is
0.010004.

\paragraph{Chance-corrected agreement.}
Where a table reports corrected agreement, it is
\begin{equation}
 \apcorr{k}(p)=\frac{A_k(p)-c_k}{A_k(p_0)-c_k},
 \label{app:eq:corrected}
\end{equation}
which maps the original pairing to one and analytic chance to zero. Values are
not clipped and may exceed one or fall below zero; the denominator is nonzero
in every reported cohort.

\begin{lemma}[Retention is invariant to affine rescaling]
\label{app:lem:affine}
Let \(M'=aM+b\) with fixed \(a\ne0\) and \(b\), applied to the original, the
semantic, and every random assignment of one cohort. If the denominator of
\autoref{eq:retention} is nonzero, then \(\apret{M'}(q)=\apret{M}(q)\).
\end{lemma}
\begin{proof}
The random-reference mean is linear, so \(\overline{M'}_{\mathrm{ref}}(q)=a\,\apref(q)+b\).
The numerator and the denominator of \autoref{eq:retention} are each multiplied by
\(a\), and \(b\) cancels.
\end{proof}

\begin{corollary}[The correction leaves retention unchanged]
\label{app:cor:corrected-retention}
Within a cohort, \(\apcorr{k}\) is an affine function of \(A_k\) with slope
\(1/(A_k(p_0)-c_k)\ne0\). Raw and chance-corrected agreement therefore give the
same retention, and the correction changes only the reporting scale. The same
holds for recall expressed as a fraction or as a percentage.
\end{corollary}

\subsection{Semantic assignment construction}
\label{app:assignment}

\paragraph{Feasible assignments.}
At replacement rate \(q\), the feasible set \(\mathcal P_q\) of
\autoref{sec:semantic-repairing} permutes assigned captions within a subset \(\mathcal I_q\) of
image identities and keeps all other partners fixed. Within \(\mathcal I_q\),
self-assignments and replacements whose text is identical to the original after
normalization are infeasible. Text normalization case-folds and collapses
whitespace. For caption groups, the comparison uses the ordered concatenation of
the normalized captions (\Cref{app:cka-protocol}). The retrieval studies use
complete re-pairing, \(q=1\) and \(\mathcal I_1=[n]\). The CKA study uses
\(q\in\{0.2,0.4,0.6,0.8,1\}\), and \(\mathcal I_q\) is the prefix of length
\(qn\) of one seeded ordering of image identities. The changed subsets are
therefore nested, and at each \(q\) the semantic and random assignments change
the same images. Every reported assignment satisfies \(F(p)=1-q\).

\paragraph{Selector vectors and solver.}
The external selector is MPNet \citep{song2020mpnet} through Sentence-BERT
\citep{reimers2019sentencebert}: \texttt{sentence-transformers/all-mpnet-base-v2}
at revision \texttt{e8c3b32edf54}, with maximum length 384, attention-mask mean
pooling, and \(\ell_2\) normalization. In the retrieval studies, \(z_i\) encodes
image \(i\)'s selected caption. In the CKA study, \(z_i\) is the normalized
average of the vectors of all captions of image \(i\). Costs
\(C_{ij}=1-z_i^\top z_j\) are float64, and infeasible edges are excluded. SciPy's
dense linear-assignment solver minimizes \autoref{eq:semantic-assignment} jointly over
\(\mathcal I_q\). Every retained solution is a bijection on \(\mathcal I_q\)
with no fixed point and no normalized-identical replacement, so the multiset of
text units is unchanged. No evaluated representation, score, or outcome enters
construction.

\paragraph{Two annotation-based selectors.}
Both annotation selectors use official COCO instance-category sets \(G_i\) and
the cost \(C^{\mathrm A}_{ij}=1-|G_i\cap G_j|/|G_i\cup G_j|\), with zero
similarity for an empty union. They differ in how they resolve the many ties of
this coarse cost (\Cref{app:tab:selectors}).
\begin{itemize}[leftmargin=*,itemsep=2pt,topsep=2pt]
\item \emph{Annotation with MPNet tie-break} (retrieval study,
\Cref{app:retrieval}). Ties are broken deterministically by MPNet distance, so
this selector is not free of the text encoder. It supplies the annotation rows of
\Cref{app:tab:representative,app:tab:conditional-ranks,app:tab:category-overlap}
and the annotation contrasts in \Cref{app:retrieval-cohort}.
\item \emph{Annotation-only} (CKA study, \Cref{app:cka-annotation}). No MPNet
tie-break and no cost perturbation are used. Starting from numerically sorted
image IDs, the source and target traversal orders are permuted independently with
NumPy's \texttt{default\_rng} seed 20260911, and SciPy 1.16.3's dense linear
assignment solves the reordered costs. This yields one deterministic optimum at
the recorded versions, not a uniform sample of optima. It is the annotation-only
assignment of \autoref{sec:study-cka}.
\end{itemize}

\begin{table}[htbp]
\centering
\caption{External selectors. Both annotation selectors minimize category-set
Jaccard distance but resolve its ties differently, so only the CKA study's
selector is free of MPNet.}
\label{app:tab:selectors}
\small
\setlength{\tabcolsep}{4pt}
\begin{tabularx}{\linewidth}{@{}p{0.25\linewidth}p{0.16\linewidth}YY@{}}
\toprule
Selector & Study & Cost and tie rule & Used in \\
\midrule
MPNet (canonical) & Both & \(1-z_i^\top z_j\); exact dense optimum &
\autoref{fig:cka-curve}, \autoref{fig:22-ckpt}, \autoref{fig:retention-example}; \Cref{app:cka,app:retrieval} \\
Annotation with MPNet tie-break & Retrieval & Jaccard; ties by MPNet distance &
\Cref{app:retrieval-representative,app:conditional,app:examples,app:retrieval-cohort} \\
Annotation-only & CKA & Jaccard; ties by seeded traversal order; no MPNet &
\autoref{fig:cka-curve}; \Cref{app:cka-annotation,app:cka-variants} \\
\bottomrule
\end{tabularx}
\end{table}

\paragraph{Alternative matchings.}
Three families of alternatives measure sensitivity to the particular optimum
(\Cref{app:tab:alternatives}). None is an independent population sample.
\emph{Cost-jittered alternatives:} for each objective, the canonical optimum is
accompanied by 15 alternatives. For alternative \(b\), every finite edge
receives independent noise from \(\mathrm{Unif}[-10^{-4}m,10^{-4}m]\), where
\(m\) is the median absolute finite cost, and an alternative is eligible only if
its unperturbed total cost lies within 1\% of the canonical optimum. The
implementation also adds a machine-scale target-index term. Its total is
constant over complete bijections, so it does not break ties; residual ties
follow the deterministic solver and the frozen row order. \emph{Cost-matched
alternatives} change at least half of the canonical assignment while matching
its distribution of assigned semantic costs. They are constructed separately for
the retrieval study (\Cref{app:retrieval-assignment}) and for the CKA study
(\Cref{app:cka-alternatives}).

\begin{table}[htbp]
\centering
\caption{Alternative semantic assignments. Overlap is the fraction of targets
shared with the canonical assignment. Every family was constructed from selector
costs and assignment overlap alone.}
\label{app:tab:alternatives}
\small
\setlength{\tabcolsep}{3.5pt}
\begin{tabularx}{\linewidth}{@{}p{0.2\linewidth}p{0.13\linewidth}rYp{0.17\linewidth}@{}}
\toprule
Family & Study & Count & Construction & Details \\
\midrule
Cost-jittered & Retrieval & 15 + 1 & Edge noise; within 1\% of optimal cost &
\Cref{app:retrieval-assignment} \\
Cost-jittered & CKA & 15 + 1 & As above; 11 of 15 change only 2 of 5,000 targets &
\Cref{app:cka-curve} \\
Cost-matched & Retrieval & 128 & Canonical overlap \(\le0.50\); matched cost profile &
\Cref{app:retrieval-assignment} \\
Cost-matched & CKA & 128 & Canonical overlap \(\le0.50\); matched cost profile &
\Cref{app:cka-alternatives} \\
\bottomrule
\end{tabularx}
\end{table}

\subsection{Random references}
\label{app:reference}
Most full comparisons use 256 seeded draws. The CKA scale curves and subset analyses use 64. The reference describes the implemented generator.

\paragraph{Generator.}
For each seed, text units with identical normalized text are grouped; groups,
and rows within groups, are ordered by seeded hash keys; and the resulting
sequence of targets within \(\mathcal I_q\) is cyclically shifted by the size of
the largest group. Each draw is therefore a valid derangement of
\(\mathcal I_q\) that uses every target once and assigns no identical normalized
text. The generator is not claimed to be uniform over \(\mathcal P_q\), so its
mean \(\apref(q)\) is a reference for this procedure rather than a uniform null
(\autoref{sec:random-reference}).

\paragraph{Which random assignments each result uses.}
\Cref{app:tab:random-sets} lists every set of random assignments. Two
distinctions matter for reading the tables. First, \emph{a single random draw is
not the random reference}. The core study saved one canonical random assignment
(seed 20260809), and it appears as one draw in
\Cref{app:tab:representative,app:tab:conditional-ranks,app:tab:category-overlap}
and in the rank distribution of \Cref{app:tab:rank-cdf}. Retention in
\autoref{eq:retention} instead uses the mean over many draws;
\Cref{app:retrieval-representative} compares the two for the model of
\autoref{fig:retention-example}. Second, the 22-model panel uses one saved set of 256
random assignments (seeds 20260810--20261065) for all 22 models. These are
the first 256 of the 1,000 draws used for the core study's empirical
reference distribution. The extension arms in \Cref{app:extensions} used their
own runner's random derangements and are not pooled with the panel.

\begin{table}[htbp]
\centering
\caption{Random re-pairings. All draws exclude fixed points and identical
normalized text within the changed subset. Only means over many draws enter
retention (\autoref{eq:retention}).}
\label{app:tab:random-sets}
\small
\setlength{\tabcolsep}{3.5pt}
\begin{tabularx}{\linewidth}{@{}p{0.29\linewidth}rp{0.19\linewidth}Y@{}}
\toprule
Set & Draws & Seeds & Used for \\
\midrule
CKA study, each \(q\) & 256 & seeded per \(q\) & \(\apref(q)\) in \autoref{fig:cka-curve};
\Cref{app:cka} \\
Core generator, \(q=1\) & 1,000 & 20260810--20261809 & Empirical reference
distributions for SigLIP, CLIP, and SigLIP 2 (\Cref{app:retrieval-sensitivity}) \\
Model panel, \(q=1\) & 256 & 20260810--20261065 & \(\apref(1)\) for all
22 models: \autoref{fig:22-ckpt}; \Cref{app:retrieval-panel} \\
Extension runner, \(q=1\) & 256 & from 20260810 & Late-interaction and audio
retention (\Cref{app:extensions}) \\
Canonical draw, \(q=1\) & 1 & 20260809 & One-draw rows only; never a
reference mean \\
\bottomrule
\end{tabularx}
\end{table}

\FloatBarrier

%% file: sec/supp/a_followup.tex
\subsection{Image-side selector and similarity ladder}
The image-side selector uses cosine similarities between cached DINOv2 image vectors.
It solves the same one-to-one problem with exclusions for the original caption and duplicate normalized caption text.
The annotation selector in the retrieval study breaks category-cost ties with MPNet similarities. The annotation selector for the caption-group CKA reconstruction uses category overlap alone.
We evaluate 128 cost-matched alternative MPNet re-pairings in the caption-group setting.
They yield CKA in the range 0.658--0.661, above the largest random score at fidelity 0.8.

For each similarity cap we disallow any substitution whose MPNet cosine similarity exceeds that cap and solve the problem again.
The uncapped COCO solution has mean similarity 0.686. When adjacent feasible solutions had mean similarities separated by more than 0.10, we added intermediate caps. The random COCO reference has mean similarity 0.096.
The image-side curves use the same construction in DINOv2 similarity. The main caption-group comparison evaluates DINOv2, so its selector robustness uses the annotation condition.

\Cref{fig:teaser} illustrates the three pairings with the same images and captions.

A label such as CLIP B/16 224 gives the model size, patch size and input resolution. The caption-group comparison uses MPNet and annotations as selectors because DINOv2 is the evaluated image encoder.

%% file: sec/supp/b_controls.tex
\section{Invariance checks and proofs}
\label{app:controls}

This appendix proves Propositions~\ref{prop:orthogonal}--\ref{prop:radial} for the invariance checks in \autoref{sec:questions}, specifies the controls of \autoref{tab:controls} and their strengths, and
reports the measured responses, including the CLIP and SigLIP 2 replications.

\subsection{Proofs of Propositions~\texorpdfstring{\ref{prop:orthogonal}--\ref{prop:radial}}{1--3}}
\label{app:proofs}

Write the representations as matrices whose rows are samples,
\(I\in\mathbb R^{n\times d_I}\) and \(T\in\mathbb R^{n\times d_T}\). An
orthogonal transformation acts on rows, \(I'=IQ_I\) and \(T'=TQ_T\), with
\(Q_I\in O(d_I)\) and \(Q_T\in O(d_T)\). Let \(\apId\) be the identity matrix and
\(\mathbf 1\) the all-ones vector.

\begin{apprestated}{Proposition~\ref{prop:orthogonal}}{Orthogonal invariance}
At fixed \(p\), applying orthogonal transformations \(Q_I\) and \(Q_T\) to \(I\)
and \(T\), respectively, preserves within-modality Euclidean distances,
covariance spectra, and neighbor graphs, and hence the marginal diagnostics used
here and \(A_k(p)\). The distance-based RBF-CKA of \autoref{sec:diagnostics} is also
unchanged. For equal-width unit-normalized representations with
\(s(t,v)=\alpha\langle t,v\rangle+\beta\) and fixed \(\alpha>0\) and \(\beta\),
a shared rotation \(Q_I=Q_T\) additionally preserves \(S\) and all candidate
rankings; a one-sided rotation need not.
\end{apprestated}
\begin{proof}
We treat the image modality; the text modality is identical.

\emph{Distances and neighbor graphs.} For all rows,
\(\lVert(v_i-v_j)Q_I\rVert_2^2=(v_i-v_j)Q_IQ_I^\top(v_i-v_j)^\top
=\lVert v_i-v_j\rVert_2^2\), so the complete matrix of within-modality
distances is unchanged. The neighbor sets \(N^I_k(i)\) are determined by that
matrix and by the index tie rule, so they are unchanged, and so are the
neighbor radii and the Levina--Bickel estimates in \autoref{app:eq:lid}, which
depend on the rows only through these distances.

\emph{Covariance spectra and effective rank.} Centering commutes with right
multiplication: \((IQ_I)_c=IQ_I-\mathbf 1\bar v^\top Q_I=I_cQ_I\). The
covariance matrix becomes \(Q_I^\top\Sigma_IQ_I\), which is similar to
\(\Sigma_I\) and has the same eigenvalues; equivalently, \(I_cQ_I\) has the same
singular values as \(I_c\). The effective rank in
\autoref{app:eq:effective-rank} is a function of these eigenvalues and is unchanged.

\emph{Neighborhood agreement.} By \autoref{app:eq:agreement-sum}, \(A_k(p)\)
depends only on the two neighbor graphs and on \(p\), all of which are fixed.

\emph{RBF-CKA.} Each RBF kernel entry is
\(K_{ij}=\exp(-\lVert v_i-v_j\rVert_2^2/(2h))\) with bandwidth \(h\) equal to
the median nonzero squared within-modality distance. Both the entries and the
bandwidth are functions of the distance matrix, so each uncentered kernel is
unchanged, and so are the centered kernels \(K^I_c\) and \(K^T_c\), their
Frobenius norms, and CKA\((p)\) in \autoref{eq:cka}. Linear CKA, used as a
sensitivity analysis in \Cref{app:cka-variants}, is also unchanged because its
Gram matrix \(IQ_IQ_I^\top I^\top=II^\top\) is invariant.

\emph{Scores.} Orthogonal maps preserve Euclidean norms, so unit-normalized rows
remain unit-normalized. Let \(d_I=d_T\). With \(Q_I=Q_T=Q\),
\(\langle t_jQ,v_iQ\rangle=t_jQQ^\top v_i^\top=\langle t_j,v_i\rangle\), so
every entry of \(S\) and hence every candidate ranking is unchanged. For a
one-sided rotation, \(S'=\alpha IQ_IQ_T^\top T^\top+\beta\mathbf 1\mathbf
1^\top\), and \(S'=S\) holds exactly when \(I(Q_IQ_T^\top-\apId)T^\top=0\). If
\(I\) and \(T\) have full column rank, this forces \(Q_IQ_T^\top=\apId\), that is,
\(Q_I=Q_T\). A one-sided rotation with \(Q_T\ne Q_I\) therefore changes \(S\) for
such representations. Whether it also changes the rankings depends on the data;
the measured controls in \Cref{app:control-results} show that the tested
rotation changes them for all three models.
\end{proof}

\begin{apprestated}{Proposition~\ref{prop:fixed-marginals}}{Fixed marginals under re-pairing}
Changing only \(p\) to another bijection preserves both representation
collections and every marginal diagnostic. The score matrix \(S\) and candidate
rankings remain fixed; only the relevance judgments encoded by \(p\) change.
\end{apprestated}
\begin{proof}
Re-pairing replaces \(p_0\) by another bijection \(p\) and leaves the indexed
collections \(I\) and \(T\) untouched. Every marginal diagnostic is a function
of \(I\) alone or of \(T\) alone (\Cref{app:marginal}), so it is unchanged.
Because \(p\) is a bijection, every text unit remains the assigned caption of
exactly one image; the multiset of designated text units is the multiset of all
text units. The score matrix \(S_{ij}=s(t_j,v_i)\) depends on \(I\), \(T\), and
\(s\) but not on \(p\), so every row and column ranking is unchanged. What
changes is which candidate is designated as relevant: \(t_{p(i)}\) for image
query \(i\) and \(v_{p^{-1}(j)}\) for text query \(j\). Consequently \(A_k(p)\),
CKA\((p)\), and Recall@\(K\) may change. In CKA, re-pairing relabels the rows
and columns of \(K^T_c\) simultaneously, which preserves its Frobenius norm, so
only the numerator of \autoref{eq:cka} can change.
\end{proof}

\begin{apprestated}{Proposition~\ref{prop:radial}}{Positive radial cancellation}
Let \(t_i=\nu(\tilde t_i)\), where \(\tilde t_i\ne0\) and \(\nu(x)=x/\lVert
x\rVert_2\). For any \(a_i>0\), \(\nu(a_i\tilde t_i)=\nu(\tilde t_i)\). Thus,
at fixed \(I\), \(p\), and \(s\), positive sample-wise scaling preserves \(T\),
the score matrix \(S\), candidate rankings, and Recall@\(K\), while raw geometry
can change.
\end{apprestated}
\begin{proof}
For \(a_i>0\), \(\lVert a_i\tilde t_i\rVert_2=a_i\lVert\tilde t_i\rVert_2\),
so \(\nu(a_i\tilde t_i)=a_i\tilde t_i/(a_i\lVert\tilde t_i\rVert_2)=\nu(\tilde
t_i)\). The normalized collection \(T\) is therefore unchanged, and so are
\(S\), every ranking, and Recall@\(K\) at fixed \(I\), \(p\), and \(s\). The
pre-normalization collection \(\{a_i\tilde t_i\}\) generally has different
pairwise distances, so diagnostics computed on it can change. The control with
\(\sigma=1\) raises pre-normalization text LID by 45.76\%
(\Cref{app:tab:control-responses}).
\end{proof}

\paragraph{Scope.}
The propositions concern the distance- and spectrum-based diagnostics and the
normalized dot-product scorer used here. They do not claim invariance for every
statistic or scoring operator. Late-interaction MaxSim (\Cref{app:late-interaction}),
for example, sums token-level maxima and is not covered by the shared-rotation
statement.

\subsection{Control construction and strengths}
\label{app:control-construction}

\Cref{app:tab:control-grid} lists the prespecified strengths. \autoref{tab:controls}
reports the maximum of each: \(\theta=\pi/2\) for both rotations,
\(\sigma=1\) for radial scaling, and a complete derangement (\(q=1\)). The
intervention seed is 20260806. No strength was chosen from outcomes.

\begin{table}[htbp]
\centering
\caption{Prespecified control grid. \autoref{tab:controls} reports the largest
strength of each control.}
\label{app:tab:control-grid}
\small
\begin{tabular}{@{}ll@{}}
\toprule
Quantity & Values \\
\midrule
Neighborhood scale \(k\) & \(\{10,20,50,100\}\); primary \(k=50\) \\
Rotation angle \(\theta\) & \(\{0,\pi/16,\pi/8,\pi/4,\pi/2\}\) \\
Radial log-scale \(\sigma\) & \(\{0,0.1,0.25,0.5,1.0\}\) \\
Re-paired fraction \(q\) & \(\{0,0.1,0.25,0.5,1.0\}\) \\
Retrieval cutoffs & R@\(\{1,5,10\}\) and full ranks \\
\bottomrule
\end{tabular}
\end{table}

\begin{itemize}[leftmargin=*,itemsep=2pt,topsep=2pt]
\item \emph{Rotations.} The generator applies a seeded permutation to the
embedding coordinates, pairs adjacent coordinates, and rotates every pair by
the same Givens angle \(\theta\); an odd final coordinate is left unchanged. The
shared rotation applies the same \(Q\) to images and texts. The one-sided
rotation applies \(Q\) to the texts only.
\item \emph{Radial scaling.} Pre-normalization text vectors are multiplied by
\(a_i=\exp(\sigma\epsilon_i)\) with independent \(\epsilon_i\sim\mathcal
N(0,1)\) and no recentering.
\item \emph{Random derangement.} At fraction \(q\), exactly \(qn\) rows are
selected by the seeded order and rearranged by one Sattolo cycle, while all
other rows keep their original partners. At \(q=1\), this is a single cycle
through all \(n\) rows. This control derangement is distinct from the random
references of \Cref{app:reference}. The implementation requires \(qn\) to be an
integer and rejects one-row subsets.
\end{itemize}

\subsection{Implementation}
\label{app:control-implementation}

The controls use SigLIP B/16 224 (\texttt{google/siglip-base-patch16-224} at
\texttt{7fd15f0689c7}) on the 5,000-pair COCO cohort of
\Cref{app:retrieval-protocol}, with its native scoring rule held fixed. The
projected surface is each tower's final projected \texttt{pooler\_output}; the
normalized surface is the model's native \texttt{image\_embeds} and
\texttt{text\_embeds} (\Cref{app:notation}). Extracted arrays are float32,
while metrics are accumulated in float64. Marginal diagnostics are computed on
both surfaces at all four \(k\); \(A_k\) uses exact Euclidean neighbors with
self excluded and index tie-breaking; and retrieval uses the native normalized
scores with candidate-index tie-breaking. Every control passed its exact gate,
which allows a relative change of at most \(2\times10^{-4}\) in any quantity the
propositions fix. The observed changes were far smaller: across all strengths,
surfaces, and \(k\), fixed marginal diagnostics moved by at most
\(3\times10^{-8}\) in relative terms (floating-point rounding), and fixed
neighborhood agreement and recall values were identical.

\subsection{Measured responses and replications}
\label{app:control-results}

\Cref{app:tab:control-responses} gives the SigLIP responses behind
\autoref{tab:controls} at full precision. The radial row is reported on both
surfaces: scaling changes the pre-normalization geometry but none of the
normalized quantities. Under complete random derangement, both representation
collections are unchanged (\autoref{prop:fixed-marginals}), while \(A_{50}\) falls by 96.24\%
and bidirectional R@1 by 47.23 points.

\begin{table}[htbp]
\centering
\caption{SigLIP B/16 224 responses at the maximum strengths, \(k=50\). Marginal
and \(A_{50}\) entries are relative changes, and R@1 is the change in
bidirectional R@1 in points; 0 denotes a relative change below
\(10^{-7}\). Original normalized-surface values:
image LID 14.995, text LID 12.383, \(A_{50}=0.2758\), R@1 47.08\% (image to
text) and 47.38\% (text to image). Projected-surface originals: text LID 12.365,
text radius 14.52, text effective rank 144.87, \(A_{50}=0.2687\).}
\label{app:tab:control-responses}
\small
\setlength{\tabcolsep}{4pt}
\begin{tabular}{@{}llrrrrr@{}}
\toprule
Control & Surface & Text LID & Text radius & Text eff.\ rank & \(A_{50}\) & R@1 (pp) \\
\midrule
Shared rotation & both & 0 & 0 & 0 & 0 & 0 \\
One-sided rotation & both & 0 & 0 & 0 & 0 & \(-47.22\) \\
Radial scaling & projected & \(+45.76\%\) & \(+62.67\%\) & \(-74.93\%\) & \(-48.69\%\) & 0 \\
Radial scaling & normalized & 0 & 0 & 0 & 0 & 0 \\
Random derangement & both & 0 & 0 & 0 & \(-96.24\%\) & \(-47.23\) \\
\bottomrule
\end{tabular}
\end{table}

Image-side marginal diagnostics are unchanged by every control, since none of
them alters the images. The one-sided rotation reduces image-to-text R@10 from
81.70\% to 0.10\% and text-to-image R@10 from 81.30\% to 0.06\%. Under the
random derangement, \(A_{50}\) falls from 0.275812 to 0.010364 on the normalized
surface, close to \(c_{50}=0.010002\) (\Cref{app:lem:chance}).

\paragraph{CLIP and SigLIP 2 replications.}
\Cref{app:tab:control-replication} repeats the three representation controls
for CLIP ViT-B/16 (\texttt{openai/clip-vit-base-patch16} at
\texttt{57c216476eef}; \citealp{radford2021learning}) and fixed-resolution
SigLIP 2 B/16 224 (\texttt{google/siglip2-base-patch16-224} at
\texttt{75de2d55ec2d}; \citealp{tschannen2025siglip2}), on the same cohort and
at the same maximum strengths. For CLIP, the projected surface applies
\texttt{visual\_projection} and \texttt{text\_projection} to the pooler
outputs. The one-sided rotation leaves corrected
agreement at exactly 1 while bidirectional R@1 falls from 34.59\% to 0.00\% for
CLIP and from 46.88\% to 0.02\% for SigLIP 2. Radial scaling and the CLIP shared
rotation leave every normalized measurement unchanged. For SigLIP 2, the shared
rotation was verified at the score level: both Gram matrices and the
cross-modal score matrix are reproduced to within \(1.3\times10^{-7}\) maximum
absolute error, as \autoref{prop:orthogonal} requires.

\begin{table}[htbp]
\centering
\caption{Control replications on normalized vectors at \(k=50\). Corrected
agreement follows \autoref{app:eq:corrected}; recalls are percentages. The SigLIP 2
shared rotation was checked on Gram and score matrices rather than as a
separately scored condition.}
\label{app:tab:control-replication}
\small
\setlength{\tabcolsep}{3.4pt}
\begin{tabular}{@{}llrrrrr@{}}
\toprule
Model & Condition & \(\apcorr{50}\) & I\(\to\)T R@1 & I\(\to\)T R@10 & T\(\to\)I R@1 & T\(\to\)I R@10 \\
\midrule
\multirow{4}{*}{CLIP B/16 224}
 & Original & 1.0000 & 36.56 & 72.30 & 32.62 & 68.94 \\
 & Shared rotation & 1.0000 & 36.56 & 72.30 & 32.62 & 68.94 \\
 & One-sided rotation & 1.0000 & 0.00 & 0.12 & 0.00 & 0.08 \\
 & Radial scaling & 1.0000 & 36.56 & 72.30 & 32.62 & 68.94 \\
\midrule
\multirow{4}{*}{SigLIP 2 B/16 224}
 & Original & 1.0000 & 45.98 & 79.72 & 47.78 & 81.50 \\
 & Shared rotation & \multicolumn{5}{c}{scores reproduced; max.\ abs.\ error \(1.3\times10^{-7}\)} \\
 & One-sided rotation & 1.0000 & 0.00 & 0.32 & 0.04 & 0.10 \\
 & Radial scaling & 1.0000 & 45.98 & 79.72 & 47.78 & 81.50 \\
\bottomrule
\end{tabular}
\end{table}

\paragraph{Marginal invariance under the semantic assignments.}
For every representative assignment, text LID at all four \(k\), the
neighbor-radius distribution, effective rank, the row-norm multiset, and the
complete text-embedding multiset were unchanged on both surfaces, as
\autoref{prop:fixed-marginals} requires. The canonical MPNet and annotation target captions
were also re-encoded through the pinned SigLIP processor and text tower; both
surfaces matched the cached rows within the declared \(5\times10^{-4}\)
tolerance. Every reported assignment had 5,000 unique targets, no
self-assignment, and no identical normalized caption.

\FloatBarrier

%% file: sec/supp/moved_invariance.tex
\subsection{Invariance statements}
\label{app:invariance-statements}

We apply the controls to SigLIP B/16 224 \citep{zhai2023sigmoid} on 5,000
COCO pairs \citep{lin2014microsoftcoco}, keeping the scoring rule $s$ fixed.
We assess marginal structure using local intrinsic
dimension (LID), neighbor radii, and effective rank; relational diagnostics
using neighborhood agreement $A_{50}$; and retrieval using bidirectional
Recall@1. \autoref{tab:controls} reports changes at the maximum prespecified
control strengths. The one-sided rotation separation also holds for
CLIP ViT-B/16 \citep{radford2021learning} and SigLIP 2 B/16-224
\citep{tschannen2025siglip2}; shared rotation and radial scaling also
preserve normalized retrieval in both models (\autoref{app:control-results}, \autoref{app:tab:control-replication}).

\begin{proposition}[Orthogonal invariance]
\label{prop:orthogonal}
At fixed $p$, applying orthogonal transformations $Q_I$ and $Q_T$ to $I$ and
$T$, respectively, preserves within-modality Euclidean distances, covariance
spectra, and neighbor graphs, and hence the marginal diagnostics used here
and $A_k(p)$. The distance-based RBF-CKA in \autoref{sec:diagnostics} is also
unchanged. For equal-width unit-normalized representations with
$s(t,v)=\alpha\langle t,v\rangle+\beta$ and fixed $\alpha>0,\beta$,
a shared rotation $Q_I=Q_T$ additionally preserves $S$ and all candidate
rankings; a one-sided rotation need not.
\end{proposition}

\begin{proposition}[Fixed marginals under re-pairing]
\label{prop:fixed-marginals}
Changing only $p$ to another bijection preserves both representation
collections and every marginal diagnostic. The score matrix $S$ and
candidate rankings remain fixed; only the relevance judgments encoded
by $p$ change.
\end{proposition}

\begin{proposition}[Positive radial cancellation]
\label{prop:radial}
Let $t_i=\nu(\tilde t_i)$, where $\tilde t_i\ne 0$ and
$\nu(x)=x/\lVert x\rVert_2$. For any $a_i>0$,
$\nu(a_i\tilde t_i)=\nu(\tilde t_i)$.
Thus, at fixed $I,p,s$, positive sample-wise scaling preserves $T$, the
score matrix $S$, candidate rankings, and Recall@$K$, while raw geometry
can change.
\end{proposition}
Proofs are in \autoref{app:proofs}; the control construction and measured responses are in \autoref{app:control-construction,app:control-implementation,app:control-results}.
Under one-sided rotation, \autoref{prop:orthogonal} guarantees that the
marginal diagnostics (Q1) and relational diagnostics (Q2) remain fixed,
while bidirectional Recall@1 (Q3) falls by 47.2 points. This
separates Q3 from both Q1 and Q2. Under random re-pairing,
\autoref{prop:fixed-marginals} guarantees that Q1 remains fixed, while
$A_{50}$ falls by 96.2\%, separating Q1 from Q2. These two controls
therefore cover all three pairs.

\autoref{prop:radial} explains why raw geometry can change while
normalized retrieval remains fixed. Shared rotation provides a
consistency check in which all three diagnostic families remain unchanged.

%% file: sec/supp/b_followup.tex
\subsection{Shared and surviving neighbor relations}
\label{sec:selectivity}
For a pairing $p$, classify each directed image-neighbor edge by whether the corresponding text-neighbor edge exists at the original pairing and after re-pairing.
Let $A_{\mathrm{kept}}$ be the fraction present in both, $A_{\mathrm{lost}}$ the fraction present only originally, and $A_{\mathrm{new}}$ the fraction present only after re-pairing. Then
\[
 A_k(p_0)=A_{\mathrm{kept}}+A_{\mathrm{lost}},\qquad
 A_k(p)=A_{\mathrm{kept}}+A_{\mathrm{new}}.
\]
Let $S_T$ be the fraction of directed text neighbors retained after reindexing by $p$. By changing the membership of only this fraction of the text graph, the fraction of common edges changes by at most $1-S_T$. The same argument gives the counterpart $1-S_I$ for the image graph under inverse reindexing.
Thus $|A_k(p)-A_k(p_0)|\le\min(1-S_I,1-S_T)$.
The change in agreement is $A_k(p_0)-A_k(p)=A_{\mathrm{lost}}-A_{\mathrm{new}}$.
We define \emph{selectivity} as the loss rate for originally shared relations divided by the overall text loss rate,
\begin{equation}
\mathrm{sel}(p)=\frac{A_{\mathrm{lost}}/A_k(p_0)}{1-S_T(p)}.
\label{eq:selectivity}
\end{equation}
A value below one means shared edges survive more often. Neighbor relations that the two towers share survive MPNet re-pairing more often than text-neighbor relations overall (selectivity below one). On COCO, the median share of text relations that survive MPNet re-pairing is 0.285.

\paragraph{Numerical checks.}
The scale implementation compared the all-pairs expression, direct sums of kernel-profile differences, and 200 brute-force swaps on synthetic kernels. The maximum error was $2.22\times10^{-16}$.
The matching replay checked 200 swaps per query problem against direct objective differences.
The 5,000-item caption-group replay matched the all-pairs CKA computed from pair sums and direct CKA to $1.91\times10^{-13}$.

%% file: sec/supp/c_cka.tex
\section{CKA and fidelity}
\label{app:cka}

This appendix supports \autoref{sec:study-cka}. It gives the reconstruction protocol, the
fidelity curve behind \autoref{fig:cka-curve}, the annotation-only assignment, the 128
alternative matchings, the linear-CKA and row-normalized variants, and the QAP
audit behind \autoref{fig:qap}.

\subsection{Reconstruction protocol}
\label{app:cka-protocol}

The study reconstructs the DINOv2/All-RoBERTa COCO experiment of
\citet{maniparambil2024vision} with the pinned public inputs of
\Cref{app:tab:cka-inputs}. The accompanying public implementation is resolved at
revision \texttt{3de1963b533c}. The original release did not pin model
revisions, so these revisions are our reconstruction inputs. Reconstruction used
Python 3.11.2, NumPy 1.26.4, SciPy 1.16.3, PyTorch 2.6.0, Transformers 4.57.6,
and Sentence Transformers 3.4.1.

\begin{table}[htbp]
\centering
\caption{Public inputs for the CKA study. Revision prefixes identify the full
revisions in the retained manifests. Both evaluated representations have width
1,024; MPNet's is 768.}
\label{app:tab:cka-inputs}
\small
\begin{tabularx}{\linewidth}{@{}lXl@{}}
\toprule
Role & Public model & Revision \\
\midrule
Image representation & \texttt{facebook/dinov2-large} \citep{oquab2023dinov2} & \texttt{47b73eefe95e} \\
Text representation & \texttt{sentence-transformers/all-roberta-large-v1} & \texttt{ace3f8f363e3} \\
External selector & \texttt{sentence-transformers/all-mpnet-base-v2} & \texttt{e8c3b32edf54} \\
\bottomrule
\end{tabularx}
\end{table}

\paragraph{Cohort and pooling.}
The cohort contains the 5,000 official COCO 2017 validation images
\citep{lin2014microsoftcoco} in the order of the official JSON file, with all
captions attached to each image: 25,014 captions in 4,987 groups of five, 12 of
six, and one of seven. DINOv2 image rows average the final hidden state over
the CLS and patch tokens. All-RoBERTa applies its native attention-mask mean
pooling and per-caption \(\ell_2\) normalization, and the released protocol then
averages the caption vectors of each image. The MPNet selector vectors \(z_i\)
are averaged in the same way and then row-normalized. All-RoBERTa and MPNet use
token limits 256 and 384. The result is one image row and one text row for each
of the 5,000 images.

\paragraph{Caption-group exclusions.}
Each image's captions follow annotation order. Two caption groups count as
identical when the ordered concatenations of their normalized captions agree.
Groups are not sorted, and sharing one caption does not make two groups
identical. This is the implemented comparison, not a test of semantic
equivalence.

\paragraph{RBF-CKA.}
For the saved rows \(z_i\) of one modality, let
\(d_{ij}=\max(\lVert z_i\rVert^2+\lVert z_j\rVert^2-2z_i^\top z_j,0)\) and let
\(h\) be the median of the nonzero \(d_{ij}\). Then
\begin{equation}
 K_{ij}=\exp\!\left(-\frac{d_{ij}}{2h}\right),\qquad
 H=\apId-\tfrac1n\mathbf 1\mathbf 1^\top,\qquad K_c=HKH,
 \label{app:eq:rbf}
\end{equation}
and \autoref{eq:cka} evaluates \(\mathrm{CKA}(p)\) from \(K^I_c\) and \(K^T_c\). The
median is PyTorch's lower middle element for an even number of nonzero entries.
Saved features are float32, while kernels and reductions use float64 and the
equivalent double-mean centering. The bandwidths are \(h=789.516856\) for images
and \(h=1.263683\) for text. A float32 literal-source check agrees within
\(5\times10^{-5}\). No per-feature scaling is applied here; the QAP procedure of
\Cref{app:qap} does apply it.

\paragraph{Partial re-pairing.}
At replacement rate \(q\), the changed subset \(\mathcal I_q\) is the prefix of
one seeded ordering of image identities (\Cref{app:assignment}). The semantic
assignment is re-solved inside \(\mathcal I_q\), and 256 random re-pairings are
drawn inside the same subset, so both conditions change the same images and
have \(F(p)=1-q\).

\paragraph{Ordinary-shuffle replication.}
To compare with the published curve, one ordinary seeded partial shuffle per
level reproduces the source procedure. It may retain fixed points, and the
original shuffle code and seed were not released, so this series is a protocol
replication rather than an exact reconstruction of the published permutations.
JSON order is our reconstruction choice; the released extraction script's
TorchVision loader sorts image IDs. Consistent relabeling preserves CKA, but
seeded subsets depend on row order, so we do not claim the authors' exact
sampled assignments.

\subsection{The fidelity curve}
\label{app:cka-curve}

\Cref{app:tab:cka-curve} gives the values plotted in \autoref{fig:cka-curve}. At
\(F(p)=0.8\), hard negative re-pairing gives 0.700670, against 0.462279 for the
random mean. With no original pairs left, hard negative re-pairing still gives
0.659390, which exceeds the random mean at \(F(p)=0.8\) and also its largest
draw, 0.471440. The original pairing scores highest (0.719270).

\begin{table}[htbp]
\centering
\caption{RBF-CKA against fidelity (\autoref{fig:cka-curve}). ``Ordinary
shuffle'' is our replication of the published procedure, which may retain
fixed points. The hard negative and random conditions change the same images at
each fidelity level. MC SE is the Monte Carlo standard error of the 256-draw
random mean.}
\label{app:tab:cka-curve}
\small
\setlength{\tabcolsep}{4.4pt}
\begin{tabular}{@{}rrrrrr@{}}
\toprule
\(F(p)\) & Published & Ordinary shuffle & Hard negative & Random mean & MC SE \\
\midrule
1.0 & 0.720 & 0.719270 & --- & 0.719270 & --- \\
0.8 & 0.460 & 0.468954 & 0.700670 & 0.462279 & 0.000209 \\
0.6 & 0.270 & 0.262519 & 0.689973 & 0.263955 & 0.000232 \\
0.4 & 0.130 & 0.131983 & 0.679331 & 0.123607 & 0.000178 \\
0.2 & 0.040 & 0.041178 & 0.669915 & 0.039682 & 0.000112 \\
0.0 & 0.010 & 0.012628 & 0.659390 & 0.012945 & 0.000018 \\
\bottomrule
\end{tabular}
\end{table}

At complete re-pairing, retention (\autoref{eq:retention}) is
\((0.659390-0.012945)/(0.719270-0.012945)=0.915223\). The canonical assignment
and its 15 cost-jittered alternatives span 0.913111--0.915223. Four of these
alternatives differ from the canonical assignment at 6.02--6.70\% of targets,
and the other 11 change only two of 5,000 targets, so they are not treated as
allocation-diverse; \Cref{app:cka-alternatives} constructs substantially
different alternatives. Five cohorts that each omit a different 20\% of images,
with both kernels rebuilt and the assignment re-solved, give retention
0.909106--0.916212.

\subsection{Annotation-only assignment and kernel response}
\label{app:cka-annotation}

The annotation-only assignment (\Cref{app:assignment}) replaces every caption
group and was saved before its CKA was evaluated. Its CKA is 0.546883, above the
random mean (0.462279) and the largest random draw (0.471440) at \(F(p)=0.8\);
its retention is 0.755939. It uses COCO object annotations and no text encoder,
so the ordering reversal of \autoref{sec:study-cka} does not depend on MPNet. The check is
motivated by the documented relationship between the two text encoders:
\texttt{all-roberta-large-v1} \citep{allrobertaModelCard} and
\texttt{all-mpnet-base-v2} are Sentence-Transformers models whose model cards
describe the same training project and overlapping training data. This follow-up was specified after the
original results.

To describe how much each assignment changes the evaluated text geometry, let
\(L=K^T_c\) be the original centered text kernel and \(L_p\) its simultaneous
row and column relabeling under \(p\). We report
\begin{equation}
 \delta_L(p)=\frac{\lVert L-L_p\rVert_F}{\lVert L\rVert_F},\qquad
 \delta_L(p)^2=2\left(1-\frac{\langle L,L_p\rangle_F}{\lVert L\rVert_F^2}\right),
 \label{app:eq:kernel-change}
\end{equation}
where the identity uses that permutation preserves the Frobenius norm. By
Cauchy--Schwarz, \(\lvert\mathrm{CKA}(p)-\mathrm{CKA}(p_0)\rvert\le\delta_L(p)\).
This bound describes sensitivity; it does not decompose the observed CKA
difference.

\begin{table}[htbp]
\centering
\caption{Selector sensitivity and text-kernel response. Random rows average
the 256 random re-pairings at that fidelity. Substitute cosine is the mean
cosine similarity between each image's original and assigned caption-group
vectors (normalized All-RoBERTa rows), averaged over all 5,000 rows, including
unchanged rows at \(F(p)=0.8\).}
\label{app:tab:cka-kernel}
\small
\setlength{\tabcolsep}{5pt}
\begin{tabular}{@{}lrrrr@{}}
\toprule
Assignment & \(F(p)\) & CKA & Substitute cosine & \(\delta_L\) \\
\midrule
Original & 1.0 & 0.719270 & 1.000000 & 0.000000 \\
Hard negative (MPNet) & 0.8 & 0.700670 & 0.956324 & 0.282460 \\
Hard negative (MPNet) & 0.0 & 0.659390 & 0.830577 & 0.508814 \\
Annotation-only & 0.0 & 0.546883 & 0.616015 & 0.864871 \\
Random mean & 0.8 & 0.462279 & 0.835758 & 0.846225 \\
Random mean & 0.0 & 0.012945 & 0.175648 & 1.406548 \\
\bottomrule
\end{tabular}
\end{table}

At \(F(p)=0\), the MPNet assignment preserves more of the text geometry than the
annotation-only assignment, which preserves more than random re-pairing. Kernel
change alone does not order CKA across conditions: the annotation-only
assignment changes the text kernel slightly more than random re-pairing at
\(F(p)=0.8\), yet it gives the higher CKA. These measurements are consistent with
CKA responding to how similar the replacement partners are, but they do not
establish a causal mechanism.

\subsection{128 alternative matchings}
\label{app:cka-alternatives}

This check asks whether the reversal depends on the particular MPNet optimum.
Before any new assignment was evaluated, we specified a construction that uses
only the MPNet caption-group vectors, the same row order, and the same forbidden
edges. No DINOv2 or All-RoBERTa value enters construction. The check was
specified after the original results.

Exact linear assignments with increasing penalties on canonical edges bound the
cost of leaving the canonical assignment. At maximum canonical overlap 0.50, the
Lagrangian lower and feasible upper bounds on the minimum semantic-cost ratio
are 1.006812 and 1.006839, so a complete re-pairing that changes at least half
of the canonical matches exists within the primary 1\% cost tolerance. The
specified in advance 2\% and 5\% fallback tolerances were not needed.

Each accepted assignment must be a complete derangement with the
caption-group exclusions, share at most 50\% of targets with the canonical
assignment and at most 80\% with every other accepted assignment, have mean
semantic cost at most 1.01 times the canonical mean, and match the canonical
distribution of assigned costs: Wasserstein distance at most 0.02, and largest
decile difference at most 0.05, times the canonical mean cost. The search
screens 16 proposals per penalty/noise setting and extends a setting to at most
256 proposals once eight are accepted; the overall cap is 1,024. The first three
settings accepted 1, 1, and 3 of 16 proposals and stopped. The fourth accepted
128 of 173, for 221 proposals in total. It uses the smallest frontier penalty
that attains 0.50 overlap, plus independent uniform edge noise with half-width
0.04 times the canonical mean cost. All assignments and gate diagnostics were
saved before any new CKA was computed.

\begin{table}[htbp]
\centering
\caption{The 128 alternative matchings on the CKA surface. Ranges are over the
constructed family and describe sensitivity to the allocation; they are not
population intervals. Cost ratios use the canonical mean semantic cost.}
\label{app:tab:cka-alternatives}
\small
\begin{tabular}{@{}lr@{}}
\toprule
Quantity & Value or range \\
\midrule
Original pairs retained & 0 in every assignment \\
Canonical matches changed & 50.00--52.00\% \\
Largest overlap between two alternatives & 74.12\% \\
Mean semantic-cost ratio & 1.008986--1.009997 \\
Wasserstein distance / canonical mean cost & 0.010527--0.013764 \\
Largest decile difference / canonical mean cost & 0.015052--0.037391 \\
RBF-CKA minimum / median / maximum & 0.658094 / 0.659450 / 0.661051 \\
Retention minimum / median / maximum & 0.913389 / 0.915309 / 0.917575 \\
Alternatives above the random mean at \(F(p)=0.8\) & 128 / 128 \\
Alternatives above every random draw at \(F(p)=0.8\) & 128 / 128 \\
\bottomrule
\end{tabular}
\end{table}

The canonical CKA of 0.659390 lies near the middle of the family: 60 of the 128
alternatives score no higher. Every alternative exceeds 0.471440, the largest of
the 256 random re-pairings that keep 80\% of original pairs. The reversal
therefore holds across substantially different allocations with the same cost
profile.

\subsection{Linear CKA and row-normalized representations}
\label{app:cka-variants}

\paragraph{The fidelity-matched comparison.}
The comparison in \autoref{sec:study-cka} pits re-pairing with no original pairs against the
random reference that keeps 80\% of them. \Cref{app:tab:cka-variants} repeats
it under linear CKA, with biased centered Gram matrices, and on row-normalized
representations, in which every saved row is \(\ell_2\)-normalized before either
kernel is built. All variants evaluate the same saved assignments: the semantic
and random assignments at every replacement rate, the annotation-only
assignment, and the 128 alternatives. This analysis was run on 2026-09-25, after
the original results; it constructs and selects no assignment. Before writing
any output, the runner reproduced the original anchors, the 128 alternative
values, the annotation-only value, and the six complete-reassignment cells of
\Cref{app:tab:cka-diagnostics} to within \(8\times10^{-16}\).

\begin{table}[htbp]
\centering
\caption{The fidelity reversal under linear CKA and row normalization. Every
column evaluates the same saved assignments. Under every variant, hard negative
re-pairing with no original pairs (bold), the annotation-only assignment, and
all 128 cost-matched alternatives exceed both the mean and the largest of the 256
random re-pairings that keep 80\% of original pairs. The first column
reproduces \autoref{fig:cka-curve}.}
\label{app:tab:cka-variants}
\small
\setlength{\tabcolsep}{4pt}
\begin{tabular}{@{}lrrrr@{}}
\toprule
 & \multicolumn{2}{c}{Saved rows} & \multicolumn{2}{c}{Row-normalized} \\
\cmidrule(lr){2-3}\cmidrule(l){4-5}
Condition & RBF-CKA & Linear CKA & RBF-CKA & Linear CKA \\
\midrule
Original pairing, \(F(p)=1\) & 0.7193 & 0.7030 & 0.7184 & 0.7038 \\
\addlinespace[2pt]
Hard negative, \(F(p)=0.8\) & 0.7007 & 0.6847 & 0.6995 & 0.6848 \\
Random mean, \(F(p)=0.8\) & 0.4623 & 0.4503 & 0.4623 & 0.4519 \\
Largest random draw, \(F(p)=0.8\) & 0.4714 & 0.4601 & 0.4716 & 0.4617 \\
\addlinespace[2pt]
Hard negative, \(F(p)=0\) & \textbf{0.6594} & \textbf{0.6439} & \textbf{0.6569} & \textbf{0.6415} \\
Annotation-only, \(F(p)=0\) & 0.5469 & 0.5351 & 0.5440 & 0.5317 \\
128 alternatives, \(F(p)=0\) & 0.658--0.661 & 0.642--0.645 & 0.656--0.659 & 0.640--0.643 \\
Random mean, \(F(p)=0\) & 0.0129 & 0.0089 & 0.0128 & 0.0089 \\
\bottomrule
\end{tabular}
\end{table}

Under every variant, hard negative re-pairing at \(F(p)=0\) exceeds the random
mean at \(F(p)=0.8\) by 0.190--0.197, and it also exceeds the largest of those 256
random draws. The annotation-only assignment exceeds that largest draw by
0.070--0.075, and all 128 alternatives exceed it. The curves keep the same shape
at the intermediate rates: under each variant, hard negative re-pairing at every
fidelity level scores above the random mean at \(F(p)=0.8\), and the random mean
decreases monotonically as fidelity decreases. The artifact retains every
variant's complete curve and all random draws.

\paragraph{Complete re-pairing with preprocessing matched across diagnostics.}
\Cref{app:tab:cka-diagnostics} reports retention at complete re-pairing when
all three relational diagnostics consume identical arrays on each surface, with
the same semantic assignment and the same 256 random re-pairings. Neighborhood
agreement \(A_{50}\) uses exact Euclidean neighbors on the declared surface,
excludes self, and breaks distance ties by index.

\begin{table}[htbp]
\centering
\caption{Complete re-pairing, \(F(p)=0\), with preprocessing matched within each
surface. Retention uses the mean of the 256 random re-pairings. The two surfaces
are reported separately.}
\label{app:tab:cka-diagnostics}
\small
\setlength{\tabcolsep}{4.8pt}
\begin{tabular}{@{}llrrrr@{}}
\toprule
Surface & Diagnostic & Original & Hard negative & Random mean & Retention \\
\midrule
\multirow{3}{*}{Saved rows}
 & RBF-CKA & 0.719270 & 0.659390 & 0.012945 & 0.915223 \\
 & Linear CKA & 0.702959 & 0.643851 & 0.008894 & 0.914838 \\
 & \(A_{50}\) & 0.370848 & 0.315332 & 0.009989 & 0.846156 \\
\midrule
\multirow{3}{*}{Row-normalized}
 & RBF-CKA & 0.718416 & 0.656919 & 0.012772 & 0.912849 \\
 & Linear CKA & 0.703807 & 0.641547 & 0.008896 & 0.910406 \\
 & \(A_{50}\) & 0.400760 & 0.338592 & 0.009999 & 0.840905 \\
\bottomrule
\end{tabular}
\end{table}

High CKA retention persists on both surfaces and under both kernels.
Neighborhood agreement retains somewhat less of its contrast, 84.6\% on saved
rows and 84.1\% after normalization.

\subsection{QAP audit}
\label{app:qap}

\paragraph{Problem setup.}
\citet{maniparambil2024vision} match captions to images by maximizing CKA over
one-to-one assignments, a quadratic assignment problem (QAP). We retain their
caption-to-image setup. The 320 anchors are the images closest in cosine
similarity to the centers of a 320-cluster K-means partition of the DINOv2 image
rows (the source defaults, \texttt{random\_state}=0), and their original pairs
are held fixed in all three problems. Each problem must then recover the
partners of 500 non-anchor query captions among 500 query images, 820 nodes in
total. Query sets are drawn with seeds 0, 1, and 2, and the image order within
the query block is shuffled with seed 0. As in the released
code, every feature is divided by its standard deviation before the kernels are
built. Each modality's kernel \(A\) or \(B\) is the median-bandwidth RBF kernel
of \autoref{app:eq:rbf}, double-centered and divided by its Frobenius norm. For a
permutation matrix \(P\) the objective is
\begin{equation}
 J(P)=\langle A,PBP^\top\rangle_F,
 \label{app:eq:qap}
\end{equation}
which is the CKA of the matching \(P\) over all 820 nodes. Because of the
feature scaling, \(J\) is not numerically identical to the saved-row CKA of
\autoref{fig:cka-curve}. The problems use our frozen JSON-ordered rows
(\Cref{app:cka-protocol}); they do not reconstruct the authors' sampled anchors
and queries.

\paragraph{Solver.}
We use the Fast Approximate QAP (FAQ) algorithm \citep{vogelstein2015fast} as
implemented by SciPy's \texttt{quadratic\_assignment} with \texttt{method="faq"},
maximization, the 320 anchors fixed through \texttt{partial\_match},
randomized doubly stochastic initialization, and tolerance \(10^{-5}\). Every
start receives an explicit seed, and all 500 starts of each problem enter the
summaries. An accuracy-blind 32-start feasibility check passed the frozen
diversity criteria before the 500-start study.

\paragraph{Outcomes.}
Fidelity counts exact recoveries among the 500 query partners; the 320 anchors
are excluded. \Cref{app:tab:qap} gives every problem's summary, and \autoref{fig:qap}
plots all 1,500 outputs. Means and rank correlations weight every start equally,
including repeated output permutations. Selecting the highest-CKA output gives
fidelity 0.25 and 0.73 points below, and 0.41 points above, the mean
over the 500 runs. That mean equals the expected fidelity of choosing a run
uniformly at random.

\begin{table}[htbp]
\centering
\caption{All 500 randomized FAQ runs per query set. Fidelity is exact recovery
over 500 queries. ``Top-CKA'' is the fidelity of the highest-objective output,
``Best'' the highest fidelity among outputs, and ``Gap'' their difference in
points. Correlations are between \(J\) and fidelity over all 500
runs.}
\label{app:tab:qap}
\small
\setlength{\tabcolsep}{3.4pt}
\begin{tabular}{@{}rrrrrrrrr@{}}
\toprule
Seed & Unique & \(J\) range & Mean fid. & Top-CKA & Best & Gap & Spearman & Kendall \\
\midrule
0 & 70  & \(2.279\times10^{-4}\) & 56.85\% & 56.6\% & 57.8\% & 1.2 & \(-0.128\) & \(-0.120\) \\
1 & 26  & \(2.617\times10^{-4}\) & 65.13\% & 64.4\% & 65.6\% & 1.2 &  0.138 &  0.102 \\
2 & 167 & \(6.246\times10^{-4}\) & 60.59\% & 61.0\% & 62.4\% & 1.4 &  0.446 &  0.321 \\
\bottomrule
\end{tabular}
\end{table}

These results do not show that the objective is worse than random run
selection, nor that it carries no information about exact recovery. They show
that the highest-CKA output and the highest-fidelity output differ in every
problem: another output from the same runs recovers six, six, and seven more
partners. The highest-fidelity output is identified with ground-truth labels for
this audit only and is not proposed as a matching rule.

\subsection{Validation}
\label{app:cka-validation}

The pipeline reconstructs all assignments, both kernels, the random means, and
the reported ratios. A feature-space identity recovers the complete-re-pairing
linear CKA to \(4.44\times10^{-16}\), and direct neighbor-set intersection
recovers \(A_{50}\) exactly. A small-array literal-centering implementation
checks RBF-CKA on both surfaces. The construction records reproduce all ensemble
costs and the exclusion and overlap gates. All 1,500 QAP outputs satisfy the
anchor constraints, their objectives reconstruct directly from \autoref{app:eq:qap},
and an independent inverse-permutation calculation reproduces every reported
fidelity. Artifacts A12, A13, A16, A17, A19, and A22 (\Cref{app:tab:artifacts})
hold the machine-readable records.

\FloatBarrier

%% file: sec/supp/c_followup.tex
\subsection{Improving swaps and contributions at different scales}
\Cref{tab:swap-detail} shows the objectives at the original pairings and after greedy ascent.
The exact gains use the archived kernels converted to float64. All 1,500 stored objectives replay on their archived kernels. An independent CPU reconstruction differs by up to $2.99\times10^{-7}$ and misses the $10^{-9}$ replay tolerance.
The comparison therefore uses the archived kernels in float64.
FAQ started at the original pairing also leaves it, with final fidelities of 96.2\%, 94.8\%, and 96.0\%.
In the same query problems, complete MPNet re-pairing scores 0.659--0.669 against a largest random score of 0.593 at fidelity 0.8.

\input{tab/swap_detail}

At rank ten, the improving-swap shares are 89.6\%, 92.9\%, and 88.2\%, against about 4.4\% of all swaps. The share of improving swaps with a negative pair contribution is 5.2\%, 16.7\%, and 7.1\%.
The top-ten mass-to-pair-share ratio is 0.154--0.196 across query sets and modality graphs. In the caption-group setting the text and image ratios are 0.064 and 0.074. Across the two graphs of each of the 22 COCO models, the median ratio is 0.14 and the range is 0.10--0.19.
The paired results are reproduced from the archived matching kernels with image rows relabeled to the original pairing.

\subsection{Scale-specific fidelity comparisons}
\Cref{fig:matching}b uses the smaller symmetric image or text neighbor rank. Its panel (c) shows the range of 64 random draws at fidelity 0.8.
The scale analysis builds one symmetric set per modality from the union of each item's nearest-neighbor lists. Text neighbors move with the text kernel after re-pairing.
An all-pairs scale uses every unordered pair and reproduces ordinary CKA.
In each condition we use the first 64 random draws at the corresponding fidelity. The mean for the first 64 at fidelity 0.8 was 0.461699, within 0.000581 of the 256-draw mean.

\begin{table}[htbp]
\centering
\caption{\textbf{Fine-scale crossover depends on the selector.}
The largest $k$ is the coarsest tested scale below all 64 random draws at fidelity 0.8. Retention uses random draws at fidelity zero.}
\label{tab:crossover}
\begin{tabular}{llrrrr}
\toprule
Setting & Re-pairing & Kernel & Retention at $k=1$ & All pairs & Largest $k$ \\
\midrule
Caption group & MPNet & RBF & 0.402 & 0.915 & None \\
Caption group & Annotations & RBF & 0.308 & 0.756 & 50 \\
Caption group & MPNet & Linear & 0.387 & 0.915 & None \\
Caption group & Annotations & Linear & 0.299 & 0.758 & 100 \\
Single caption & MPNet & RBF & 0.490 & 0.889 & None \\
Single caption & Model-judged & RBF & 0.428 & 0.858 & 1 \\
Single caption & MPNet & Linear & 0.476 & 0.889 & None \\
Single caption & Model-judged & Linear & 0.419 & 0.859 & 1 \\
\bottomrule
\end{tabular}
\end{table}
The two-scale contrast depends on the substituted captions. The complete caption-group MPNet condition remains above every 80\%-fidelity random draw at each scale. The annotation and model-judged conditions cross as shown in \cref{tab:crossover}.

Taking the smaller of the image- and text-kernel perturbations does not restore the CKA ordering. The annotation and fidelity-0.8 random means have perturbation minima 0.864871 and 0.845916, with CKA 0.546883 and 0.462279, respectively.

\begin{figure}[htbp]
\centering\includegraphics[width=.82\linewidth]{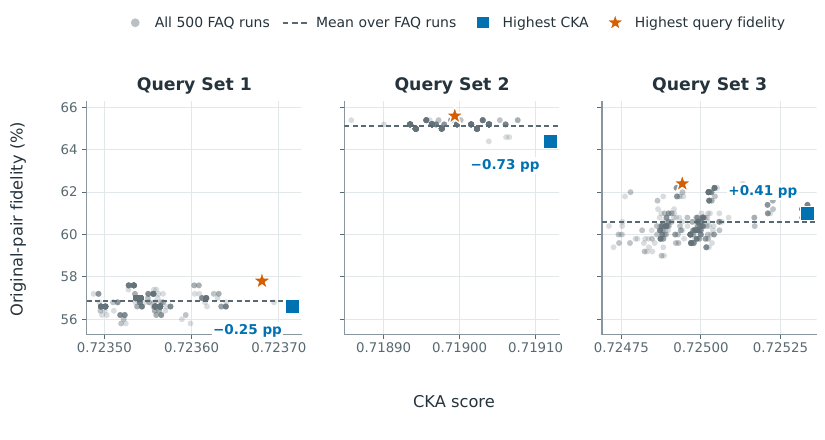}
\caption{\textbf{Solver objectives and recovered fidelity.} The archived plot shows the candidate sets across the three matching problems.}
\label{fig:qap}
\end{figure}

%% file: tab/swap_detail.tex
\begin{table}[t]
\centering\footnotesize\setlength{\tabcolsep}{3pt}
\caption{\textbf{Every original pairing has improving swaps.} Each problem has 320 fixed anchors, 500 queries, 124,750 query swaps and 500 FAQ random starts. Fidelity is measured over queries.}
\label{tab:swap-detail}
\begin{tabular}{@{}lrrr@{}}
\toprule
Quantity & Problem 1 & Problem 2 & Problem 3 \\
\midrule
Original objective & 0.7192 & 0.7157 & 0.7199 \\
Improving swaps & 135 & 126 & 127 \\
\quad Within ten neighbors & 89.6\% & 92.9\% & 88.2\% \\
\quad Nearest neighbor & 33.3\% & 33.3\% & 37.0\% \\
All swaps within ten neighbors & 4.4\% & 4.4\% & 4.4\% \\
Greedy swaps to a local maximum & 112 & 87 & 99 \\
\quad Final objective & 0.7238 & 0.7203 & 0.7249 \\
\quad Final fidelity & 68.2\% & 73.8\% & 71.2\% \\
FAQ from original, final fidelity & 96.2\% & 94.8\% & 96.0\% \\
FAQ random starts above original & 500 & 500 & 500 \\
\quad Best objective & 0.7237 & 0.7191 & 0.7253 \\
\quad Mean fidelity & 56.8\% & 65.1\% & 60.6\% \\
Top-ten caption-neighbor pairs: &  &  &  \\
\quad Share of pairs & 1.70\% & 1.69\% & 1.70\% \\
\quad Share of CKA & 0.26\% & 0.26\% & 0.26\% \\
\bottomrule
\end{tabular}
\end{table}

%% file: sec/supp/d_retrieval.tex
\section{Relational diagnostics and retrieval of the assigned caption}
\label{app:retrieval}

This appendix supports \autoref{sec:study-profile}. It gives the protocol for the 22
models, their complete results, the representative models and the
random reference behind \autoref{fig:retention-example}, assigned ranks conditional on original
top-1 success, qualitative examples, and the sensitivity analyses, with the
models each analysis covers.

\subsection{Protocol}
\label{app:retrieval-protocol}

\paragraph{Cohort and caption selection.}
The cohort contains 5,000 unique COCO validation images
\citep{lin2014microsoftcoco} from \texttt{lmms-lab/COCO-Caption2017} at
revision \texttt{3bdd5827e243}. Image identities are ranked by a seeded stable
hash (seed 20260804), and one caption per selected image is then chosen by a
separate seeded hash (seed 20260805). Neither step observes a representation,
an annotation, or an outcome. The study therefore uses one outcome-independent
caption realization, and its magnitudes are conditional on it.

\paragraph{One assigned caption per query.}
Each image query has exactly one relevant caption and each caption query exactly
one relevant image, in galleries of 5,000 images and 5,000 captions. Standard
COCO evaluation instead counts any of an image's five captions as relevant, so
absolute recalls here are not comparable with standard benchmark numbers.

\paragraph{Checkpoints and preprocessing.}
\Cref{app:tab:roster} lists the 22 models. The 16 CLIP, SigLIP, and
SigLIP 2 models \citep{radford2021learning,zhai2023sigmoid,tschannen2025siglip2}
are loaded with their \texttt{AutoProcessor} at the pinned revision. Images are
converted to RGB and passed through the processor's resize, crop, and
normalization configuration; text uses max-length padding and truncation under
the versioned tokenizer. The scored vectors are the native normalized
\texttt{image\_embeds} and \texttt{text\_embeds}. SigLIP So400m/14 224 has a
published 16-token text limit and truncates 762 of the 5,000 captions
(15.24\%); no other model in the panel truncates a caption.

The six additional encoders use their native processors, with no common manual
resize: 512-pixel inputs and a 1,024-token processing length for Jina-CLIP-v2
\citep{koukounas2024jinaclipv2}, under which no caption is truncated; 224-pixel
center crops and a 248-token window for both LongCLIP variants
\citep{zhang2024longclip}; any-resolution 336-pixel tiling for E5-V
\citep{jiang2024e5v}; and a dynamic budget of 4,096--1,310,720 pixels with the
library's query path for Qwen3-VL-Embedding-2B and -8B
\citep{li2026qwen3vlembedding}. E5-V follows its published templates: the
caption, or \texttt{<image>}, is followed by a newline and
\texttt{Summary above sentence in one word:} or
\texttt{Summary above image in one word:}, respectively. Every output is
explicitly \(\ell_2\)-normalized; the Jina and Qwen outputs are renormalized in
float64 because their loader outputs did not meet a strict unit-norm tolerance.

\begin{table}[tbp]
\centering
\caption{The 22 global image--text models of \autoref{sec:study-profile} at their
evaluated revisions. \(d\) is the width of the scored vectors. For the 16 CLIP,
SigLIP, and SigLIP 2 models, the image column gives the native input
resolution and the text column the native token limit; the six additional
encoders use the native processors described in \Cref{app:retrieval-protocol}.
Full revisions are retained with each record.}
\label{app:tab:roster}
\scriptsize
\setlength{\tabcolsep}{3pt}
\begin{tabular}{@{}llllrr@{}}
\toprule
Model & Identifier & Revision & \(d\) & Image & Text \\
\midrule
CLIP B/32 224 & \texttt{openai/clip-vit-base-patch32} & \texttt{3d74acf9a28c} & 512 & 224 & 77 \\
CLIP B/16 224 & \texttt{openai/clip-vit-base-patch16} & \texttt{57c216476eef} & 512 & 224 & 77 \\
CLIP L/14 224 & \texttt{openai/clip-vit-large-patch14} & \texttt{32bd64288804} & 768 & 224 & 77 \\
CLIP L/14 336 & \texttt{openai/clip-vit-large-patch14-336} & \texttt{ce19dc912ca5} & 768 & 336 & 77 \\
\midrule
SigLIP B/16 224 & \texttt{google/siglip-base-patch16-224} & \texttt{7fd15f0689c7} & 768 & 224 & 64 \\
SigLIP B/16 256 & \texttt{google/siglip-base-patch16-256} & \texttt{b078df89e446} & 768 & 256 & 64 \\
SigLIP B/16 384 & \texttt{google/siglip-base-patch16-384} & \texttt{41aec1c83b32} & 768 & 384 & 64 \\
SigLIP L/16 256 & \texttt{google/siglip-large-patch16-256} & \texttt{d0da9f876e7d} & 1024 & 256 & 64 \\
SigLIP L/16 384 & \texttt{google/siglip-large-patch16-384} & \texttt{ce005573a409} & 1024 & 384 & 64 \\
SigLIP So400m/14 224 & \texttt{google/siglip-so400m-patch14-224} & \texttt{d04cf29fca7b} & 1152 & 224 & 16 \\
\midrule
SigLIP 2 B/16 224 & \texttt{google/siglip2-base-patch16-224} & \texttt{75de2d55ec2d} & 768 & 224 & 64 \\
SigLIP 2 B/16 256 & \texttt{google/siglip2-base-patch16-256} & \texttt{3f9f96cb90da} & 768 & 256 & 64 \\
SigLIP 2 B/16 384 & \texttt{google/siglip2-base-patch16-384} & \texttt{f775b65a7976} & 768 & 384 & 64 \\
SigLIP 2 L/16 256 & \texttt{google/siglip2-large-patch16-256} & \texttt{787800c8990e} & 1024 & 256 & 64 \\
SigLIP 2 L/16 384 & \texttt{google/siglip2-large-patch16-384} & \texttt{1b426889ea62} & 1024 & 384 & 64 \\
SigLIP 2 So400m/14 224 & \texttt{google/siglip2-so400m-patch14-224} & \texttt{78e403963a4f} & 1152 & 224 & 64 \\
\midrule
E5-V & \texttt{royokong/e5-v} & \texttt{684c4c91ebab} & 4096 & native & native \\
Jina-CLIP-v2 & \texttt{jinaai/jina-clip-v2} & \texttt{e10d47f5691d} & 1024 & native & native \\
LongCLIP-GmP ViT-L/14 & \texttt{zer0int/LongCLIP-GmP-ViT-L-14} & \texttt{b476fa98520c} & 768 & native & native \\
LongCLIP-L (Diffusers) & \texttt{zer0int/LongCLIP-L-Diffusers} & \texttt{89090388cbab} & 768 & native & native \\
Qwen3-VL-Embedding-2B & \texttt{Qwen/Qwen3-VL-Embedding-2B} & \texttt{9f2f7e710d6d} & 2048 & native & native \\
Qwen3-VL-Embedding-8B & \texttt{Qwen/Qwen3-VL-Embedding-8B} & \texttt{2c4565515e0f} & 4096 & native & native \\
\bottomrule
\end{tabular}
\end{table}

\paragraph{Extraction of the six additional encoders.}
All six were extracted afresh on 2026-09-24 at the revisions of
\Cref{app:tab:roster}, with PyTorch 2.6.0, Sentence Transformers 5.4.0, and
Transformers 4.57.6 (Jina, both LongCLIPs) or 5.5.0 (E5-V, both Qwens). Before
the full cohort, each model was checked on the first 16 rows: outputs had to be
finite and normalized, reach bidirectional R@1 of at least 0.25 on that small
gallery, and reproduce under a repeated extraction with the same batches to
within \(5\times10^{-4}\); Qwen's cosine scorer also had to agree with the
normalized dot products to within \(5\times10^{-5}\). These are setup checks,
not results. The fresh original-pair recalls differ from the earlier extension
records by at most 0.16 points.

\paragraph{Scores, neighbors, and retention.}
Scores are float32 dot products of the normalized vectors, and ties are broken
by candidate index. Neighbor sets use exact Euclidean distances on the
normalized vectors, exclude the item itself, and break ties by index. The
primary scale is \(k=50\); \(k\in\{10,20,100\}\) are secondary. All 22
models use one saved array of assignments: the identity, the canonical
MPNet assignment, its 15 alternatives, and 256 random re-pairings
(\Cref{app:reference}). Retention follows \autoref{eq:retention} with
\(\apref(1)\) the mean over those 256 re-pairings; for bidirectional recall, the
two directions are averaged within each condition first.

\paragraph{Protocol status.}
The 16-model panel's roster and analysis rules were fixed before its
outcomes were inspected; three of its rows were already known from the
representative studies. The six-encoder analysis of neighborhood agreement was
specified after their retrieval outcomes were known, and its cohort, assignments,
graph scales, and reporting rule were fixed before their agreement outcomes were
inspected (\Cref{app:status}).

\subsection{Per-model results for all 22 models}
\label{app:retrieval-panel}

\Cref{app:tab:panel} gives every model's original values and its three
retentions, the values plotted in \autoref{fig:22-ckpt}. All 22 satisfy
\(\rho_{A_{50}}>\rho_{\mathrm{R@10}}>\rho_{\mathrm{R@1}}\). Neighborhood
agreement keeps 71.9--81.6\% of its original-minus-random gap, bidirectional
R@10 keeps 46.0--53.1\%, and bidirectional R@1 keeps 7.3--14.9\%. Every
retention uses the same 256 random re-pairings, so the rows are directly
comparable.

\begin{table}[tbp]
\centering
\caption{Complete results for the 22 models under complete hard negative
re-pairing (\(q=1\)), the values plotted in \autoref{fig:22-ckpt}. Original
\(A_{50}\) and bidirectional recalls (\%) are original-pairing values; each
\(\rho\) is \autoref{eq:retention} with the mean of the same 256 random
re-pairings for every model. All 22 rows satisfy
\(\rho_{A_{50}}>\rho_{\mathrm{R@10}}>\rho_{\mathrm{R@1}}\). Ranges:
\(\rho_{A_{50}}\) 0.7193--0.8160,
\(\rho_{\mathrm{R@10}}\) 0.4601--0.5313,
\(\rho_{\mathrm{R@1}}\) 0.0732--0.1485.}
\label{app:tab:panel}
\small
\setlength{\tabcolsep}{4pt}
\begin{tabular}{@{}lrrrrrr@{}}
\toprule
 & \multicolumn{2}{c}{Neighborhood agreement} & \multicolumn{2}{c}{Bidirectional R@10} & \multicolumn{2}{c}{Bidirectional R@1} \\
\cmidrule(lr){2-3}\cmidrule(lr){4-5}\cmidrule(l){6-7}
Model & \(A_{50}(p_0)\) & \(\rho_{A_{50}}\) & \(p_0\) (\%) & \(\rho_{\mathrm{R@10}}\) & \(p_0\) (\%) & \(\rho_{\mathrm{R@1}}\) \\
\midrule
CLIP B/32 224 & 0.2742 & 0.8081 & 69.12 & 0.4704 & 32.04 & 0.1485 \\
CLIP B/16 224 & 0.2779 & 0.8091 & 70.62 & 0.4876 & 34.59 & 0.1306 \\
CLIP L/14 224 & 0.2835 & 0.8116 & 72.19 & 0.4838 & 37.22 & 0.1262 \\
CLIP L/14 336 & 0.2902 & 0.8160 & 73.11 & 0.4834 & 38.30 & 0.1232 \\
\midrule
SigLIP B/16 224 & 0.2758 & 0.7787 & 81.50 & 0.4709 & 47.23 & 0.0959 \\
SigLIP B/16 256 & 0.2794 & 0.7825 & 81.70 & 0.4710 & 47.30 & 0.0877 \\
SigLIP B/16 384 & 0.2764 & 0.7804 & 82.63 & 0.4765 & 49.17 & 0.0938 \\
SigLIP L/16 256 & 0.2564 & 0.7699 & 83.75 & 0.4744 & 51.10 & 0.0855 \\
SigLIP L/16 384 & 0.2495 & 0.7668 & 84.19 & 0.4802 & 52.87 & 0.0842 \\
SigLIP So400m/14 224 & 0.2392 & 0.7795 & 83.05 & 0.4754 & 50.81 & 0.0852 \\
\midrule
SigLIP 2 B/16 224 & 0.2868 & 0.7833 & 80.61 & 0.4762 & 46.88 & 0.0962 \\
SigLIP 2 B/16 256 & 0.2878 & 0.7864 & 80.90 & 0.4770 & 47.85 & 0.0980 \\
SigLIP 2 B/16 384 & 0.2874 & 0.7859 & 82.59 & 0.4747 & 49.57 & 0.0950 \\
SigLIP 2 L/16 256 & 0.2719 & 0.7666 & 82.47 & 0.4662 & 50.16 & 0.0869 \\
SigLIP 2 L/16 384 & 0.2704 & 0.7692 & 82.95 & 0.4601 & 50.68 & 0.0856 \\
SigLIP 2 So400m/14 224 & 0.2671 & 0.7656 & 82.35 & 0.4619 & 49.63 & 0.0848 \\
\midrule
E5-V & 0.4868 & 0.7422 & 82.99 & 0.5311 & 48.52 & 0.1099 \\
Jina-CLIP-v2 & 0.3876 & 0.7629 & 78.44 & 0.5207 & 42.40 & 0.1231 \\
LongCLIP-GmP ViT-L/14 & 0.4822 & 0.7449 & 83.02 & 0.5313 & 48.72 & 0.1047 \\
LongCLIP-L (Diffusers) & 0.3458 & 0.7971 & 76.83 & 0.5062 & 42.74 & 0.1144 \\
Qwen3-VL-Embedding-2B & 0.4135 & 0.7344 & 84.23 & 0.5017 & 50.93 & 0.0831 \\
Qwen3-VL-Embedding-8B & 0.4017 & 0.7193 & 86.74 & 0.4905 & 54.83 & 0.0732 \\
\bottomrule
\end{tabular}
\end{table}

\paragraph{The six additional encoders before and after the follow-up.}
The six additional encoders were first evaluated for retrieval only, in an
extension run that drew its random derangements with its own generator
(rejection sampling, seeded from 20260810) and resolved score ties
optimistically. The follow-up recomputed all three diagnostics under the
panel's conventions, including the stable tie rule
(\Cref{app:tab:six-reconciliation}). Retentions change by at most 0.004 for
R@10 and 0.003 for R@1. In particular, the largest R@10 retention among the six
is 0.5324 (E5-V) in the earlier run and 0.5313 (LongCLIP-GmP) in the follow-up;
\autoref{fig:22-ckpt} uses the follow-up values, so its range ends at 0.531. Only
the follow-up values share the panel's random reference and are reported
alongside the 16 other models.

\begin{table}[htbp]
\centering
\caption{The six additional encoders before and after the follow-up. The
earlier extension run measured retrieval only, with its own random derangements
and optimistic handling of score ties. The follow-up re-extracted all six from
the recorded revisions and recomputed all three diagnostics with the panel's
saved assignments and stable tie rule. Recalls are bidirectional percentages;
the follow-up values are those of \Cref{app:tab:panel} and
\autoref{fig:22-ckpt}.}
\label{app:tab:six-reconciliation}
\small
\setlength{\tabcolsep}{3.6pt}
\begin{tabular}{@{}lrrrrrrr@{}}
\toprule
 & \multicolumn{3}{c}{Earlier extension run} & \multicolumn{4}{c}{Follow-up (panel reference)} \\
\cmidrule(lr){2-4}\cmidrule(l){5-8}
Encoder & R@10 & \(\rho_{\mathrm{R@10}}\) & \(\rho_{\mathrm{R@1}}\) & R@10 & \(\rho_{\mathrm{R@10}}\) & \(\rho_{\mathrm{R@1}}\) & \(\rho_{A_{50}}\) \\
\midrule
E5-V & 82.97 & 0.5324 & 0.1126 & 82.99 & 0.5311 & 0.1099 & 0.7422 \\
Jina-CLIP-v2 & 78.60 & 0.5167 & 0.1230 & 78.44 & 0.5207 & 0.1231 & 0.7629 \\
LongCLIP-GmP ViT-L/14 & 83.08 & 0.5287 & 0.1034 & 83.02 & 0.5313 & 0.1047 & 0.7449 \\
LongCLIP-L (Diffusers) & 76.86 & 0.5050 & 0.1157 & 76.83 & 0.5062 & 0.1144 & 0.7971 \\
Qwen3-VL-Embedding-2B & 84.17 & 0.5008 & 0.0837 & 84.23 & 0.5017 & 0.0831 & 0.7344 \\
Qwen3-VL-Embedding-8B & 86.73 & 0.4917 & 0.0739 & 86.74 & 0.4905 & 0.0732 & 0.7193 \\
\bottomrule
\end{tabular}
\end{table}

\subsection{Representative models and the random reference}
\label{app:retrieval-representative}

\Cref{app:tab:representative} reports image-to-text values for the three
representative models. It separates two quantities that must not be
conflated: one random re-pairing (the canonical draw, seed 20260809) and the
random reference \(\apref\), the mean over the 256 panel draws that enters
\autoref{eq:retention}. For SigLIP B/16 224, image-to-text mean reciprocal rank is
0.1536 under the MPNet assignment, 0.1013 under the annotation assignment, and
0.0015 under the one-draw random assignment. For CLIP, it is 0.4855, 0.1454, and
0.0015 for the original, MPNet, and one-draw conditions, and text-to-image R@10
is 68.94\%, 33.84\%, and 0.08\%. For SigLIP 2, text-to-image R@1 is 47.78\%,
5.00\%, and 0.00\%, and text-to-image R@10 is 81.50\%, 39.70\%, and 0.10\%.
Recall and reciprocal rank always evaluate the assigned caption, not the
original one. The CLIP one-draw image-to-text R@10 of 0.14\% lies below the
prespecified screening reference of 0.52\%, analytic chance \(10/5000\) plus five
binomial standard errors; because globally constrained assignments do not give
independent query outcomes, this is a heuristic screen rather than a calibrated
test.

\begin{table}[htbp]
\centering
\caption{Representative models, image-to-text, \(k=50\), normalized
vectors. The one-draw rows show the single canonical random re-pairing (seed
20260809). The reference rows are the mean over the 256 panel draws, the
\(\apref\) of \autoref{eq:retention}. The annotation row uses the MPNet tie-break
(\Cref{app:tab:selectors}). Recalls are percentages.}
\label{app:tab:representative}
\small
\setlength{\tabcolsep}{3.6pt}
\begin{tabular}{@{}llrrrr@{}}
\toprule
Model & Condition & \(A_{50}\) & \(\apcorr{50}\) & R@10 & R@1 \\
\midrule
\multirow{5}{*}{SigLIP B/16 224} & Original pairing & 0.275812 & 1.0000 & 81.70 & 47.08 \\
 & \textbf{Hard negative (MPNet)} & \textbf{0.216984} & \textbf{0.7787} & \textbf{38.44} & \textbf{4.10} \\
 & Hard negative (annotation) & 0.182965 & 0.6507 & 24.32 & 2.74 \\
 & Random, one draw & 0.010508 & 0.0019 & 0.06 & 0.00 \\
 & Random reference \(\apref\) & 0.010004 & 0.0000 & 0.19 & 0.01 \\
\midrule
\multirow{4}{*}{SigLIP 2 B/16 224} & Original pairing & 0.286752 & 1.0000 & 79.72 & 45.98 \\
 & \textbf{Hard negative (MPNet)} & \textbf{0.226776} & \textbf{0.7833} & \textbf{37.26} & \textbf{4.04} \\
 & Random, one draw & 0.010220 & 0.0008 & 0.20 & 0.00 \\
 & Random reference \(\apref\) & 0.010012 & 0.0000 & 0.18 & 0.01 \\
\midrule
\multirow{4}{*}{CLIP B/16 224} & Original pairing & 0.277868 & 1.0000 & 72.30 & 36.56 \\
 & \textbf{Hard negative (MPNet)} & \textbf{0.226720} & \textbf{0.8091} & \textbf{35.22} & \textbf{4.26} \\
 & Random, one draw & 0.010268 & 0.0010 & 0.14 & 0.00 \\
 & Random reference \(\apref\) & 0.009995 & 0.0000 & 0.19 & 0.01 \\
\bottomrule
\end{tabular}
\end{table}

\paragraph{Single draw and random reference for \autoref{fig:retention-example}.}
\autoref{fig:retention-example} shows SigLIP B/16 224 image-to-text values, and its brackets
are the retentions of \autoref{eq:retention}. \Cref{app:tab:figure5} places the
canonical single draw next to \(\apref\). For R@10, the single draw (0.06\%)
lies at the 1st percentile of the core generator's 1,000 draws, whose mean is
0.18\%, while the 256-draw panel mean is 0.19\%. The random reference is 0.0100
for \(A_{50}\), 0.19\% for R@10, and 0.01\% for R@1. Retention computed against
either the single draw or \(\apref\) rounds to the displayed 0.78, 0.47, and
0.09, because every random value is small relative to its original-minus-random
gap.

\begin{table}[htbp]
\centering
\caption{Single random draw and random reference for SigLIP B/16 224,
image-to-text, the model and direction of \autoref{fig:retention-example}. The random
reference of \autoref{eq:retention} is the mean over the 256 panel draws.
Retention against either rounds to the values displayed in \autoref{fig:retention-example}:
0.78, 0.47, and 0.09.}
\label{app:tab:figure5}
\small
\setlength{\tabcolsep}{4pt}
\begin{tabular}{@{}lrrrrrr@{}}
\toprule
 & & & \multicolumn{2}{c}{Random} & \multicolumn{2}{c}{Retention against} \\
\cmidrule(lr){4-5}\cmidrule(l){6-7}
Diagnostic & Original & Hard negative & One draw & \(\apref\) (256) & One draw & \(\apref\) \\
\midrule
\(A_{50}\) & 0.2758 & 0.2170 & 0.0105 & 0.0100 & 0.778 & 0.779 \\
R@10 (\%) & 81.70 & 38.44 & 0.06 & 0.19 & 0.470 & 0.469 \\
R@1 (\%) & 47.08 & 4.10 & 0.00 & 0.01 & 0.087 & 0.087 \\
\bottomrule
\end{tabular}
\end{table}

\begin{table}[htbp]
\centering
\caption{Assigned-pair Recall@\(K\) (\%) for SigLIP B/16 224, image-to-text,
over the full 5,000-caption gallery. The score matrix and every candidate
ranking are identical in all four rows; only the assigned caption changes.
Both hard negative assignments lie between the original pairing and one random
re-pairing at every cutoff.}
\label{app:tab:rank-cdf}
\small
\setlength{\tabcolsep}{3.6pt}
\begin{tabular}{@{}lrrrrrrrrr@{}}
\toprule
Condition & \(K=1\) & 5 & 10 & 20 & 50 & 100 & 250 & 500 & 1,000 \\
\midrule
Original pairing & 47.08 & 72.14 & 81.70 & 88.74 & 96.26 & 99.00 & 99.76 & 99.92 & 99.96 \\
Hard negative (MPNet) & 4.10 & 25.82 & 38.44 & 52.18 & 71.82 & 83.32 & 91.60 & 95.52 & 97.94 \\
Hard negative (annotation) & 2.74 & 15.68 & 24.32 & 36.04 & 54.70 & 67.78 & 79.56 & 86.76 & 91.86 \\
Random, one draw & 0.00 & 0.04 & 0.06 & 0.42 & 0.96 & 2.06 & 4.98 & 9.88 & 20.10 \\
\bottomrule
\end{tabular}
\end{table}

\subsection{Assigned ranks conditional on original top-1 success}
\label{app:conditional}

With scores fixed, the original partner stays in the gallery. Under a complete
derangement with deterministic tie handling, a query's top-ranked candidate can
be its original partner or its assigned caption, but not both, so
\(\mathrm{R@1}(p_0)+\mathrm{R@1}(p)\le1\). For SigLIP B/16 224 image-to-text
retrieval, original R@1 is 47.08\%, so this bound allows assigned R@1 up to
52.92\%; it does not force the observed 4.10\%. To remove the constraint, we
condition on whether the original partner ranks first, using the saved
single-caption representations and assignments
(\Cref{app:tab:conditional-ranks}). The analysis was specified after the
original results and before these summaries were computed.

\begin{table}[htbp]
\centering
\caption{SigLIP B/16 224 assigned ranks conditional on original top-1 success.
Recalls are percentages; median ranks are out of 5,000. ``Random'' is the
canonical single draw, not the reference mean. Each direction conditions on its
own original ranking.}
\label{app:tab:conditional-ranks}
\small
\setlength{\tabcolsep}{4.2pt}
\begin{tabular}{@{}llrrrrr@{}}
\toprule
Direction & Assignment & Original top-1 & Queries & R@1 & R@10 & Median rank \\
\midrule
\multirow{6}{*}{Image \(\to\) text}
 & Hard negative (MPNet) & yes & 2,354 & 0.00 & 43.76 & 14 \\
 & Hard negative (MPNet) & no & 2,646 & \textbf{7.75} & 33.71 & 23 \\
 & Hard negative (annotation) & yes & 2,354 & 0.00 & 26.30 & 40 \\
 & Hard negative (annotation) & no & 2,646 & 5.18 & 22.56 & 41 \\
 & Random, one draw & yes & 2,354 & 0.00 & 0.00 & 2,547.5 \\
 & Random, one draw & no & 2,646 & 0.00 & 0.11 & 2,501.5 \\
\midrule
\multirow{6}{*}{Text \(\to\) image}
 & Hard negative (MPNet) & yes & 2,369 & 0.00 & 38.75 & 17 \\
 & Hard negative (MPNet) & no & 2,631 & \textbf{9.46} & 38.31 & 18 \\
 & Hard negative (annotation) & yes & 2,369 & 0.00 & 25.75 & 39 \\
 & Hard negative (annotation) & no & 2,631 & 5.47 & 24.17 & 40 \\
 & Random, one draw & yes & 2,369 & 0.00 & 0.17 & 2,521 \\
 & Random, one draw & no & 2,631 & 0.04 & 0.08 & 2,590 \\
\bottomrule
\end{tabular}
\end{table}

Among queries whose original partner is not ranked first, assigned R@1 under
the MPNet assignment is 7.75\% image-to-text and 9.46\% text-to-image, with
R@10 of 33.71\% and 38.31\%. Low assigned-pair R@1 therefore persists outside
the stratum that the constraint forces to fail. This is not a test of whether
the replacement captions are acceptable descriptions. Every saved rank is
recovered exactly from the cached score matrix, and the weighted conditional
summaries recover the pooled values. For text queries, the saved rank at source
index \(i\) belongs to query \(p(i)\) and is inverse-permuted before
conditioning.

\subsection{Qualitative examples and object overlap}
\label{app:examples}

\begin{figure}[htbp]
\centering
\includegraphics[width=\linewidth]{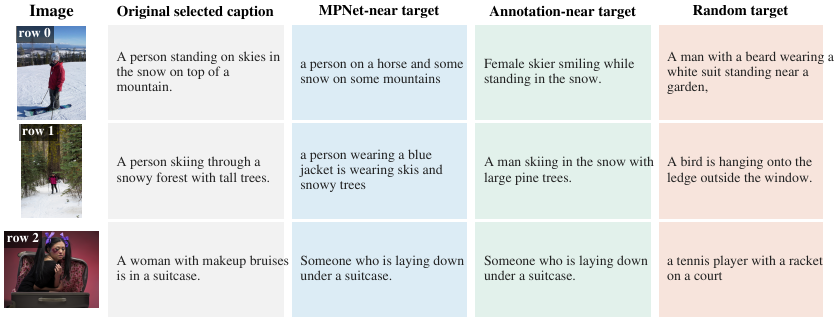}
\caption{The first three rows of the frozen cohort order. From left to right:
the image, its original caption, the captions assigned to it by the MPNet and
the annotation hard negative assignments, and the caption assigned by one
random re-pairing. Hard negative captions can preserve a scene, object, or
action while describing a different image; the displayed random captions are
unrelated. Every
assignment is a complete bijection, and every displayed caption belongs to a
different image.}
\label{app:fig:examples}
\end{figure}

\Cref{app:fig:examples} shows the first three rows of the frozen cohort order.
The display rule uses no MPNet distance, rank, agreement, or other outcome, so
the examples are not selected for visual success. They illustrate how a hard
negative caption can preserve an object or scene while belonging to a different
image.

We also compare official COCO instance-category sets between each source image
and the image whose caption it is assigned (\Cref{app:tab:category-overlap}).
This check uses the three single-caption assignments of the retrieval study,
including the annotation assignment with its MPNet tie-break; it does not use
the CKA study's annotation-only assignment. An empty union has Jaccard zero, and
48 of the 5,000 source images have an empty category set.

\begin{table}[htbp]
\centering
\caption{Object-category overlap under complete re-pairing. The random row is
the canonical single draw. Sharing object categories, or even an identical
category set, does not make a caption an acceptable description.}
\label{app:tab:category-overlap}
\small
\begin{tabular}{@{}lrrrr@{}}
\toprule
Assignment & Mean Jaccard & Any shared & Identical nonempty & Empty unions \\
\midrule
Hard negative (MPNet) & 0.496039 & 91.84\% & 20.86\% & 20 \\
Hard negative (annotation) & 0.859223 & 99.04\% & 63.10\% & 48 \\
Random, one draw & 0.075186 & 34.00\% & 0.26\% & 0 \\
\bottomrule
\end{tabular}
\end{table}

The MPNet assignment favors images that share annotated objects, relative to
random re-pairing. The check measures neither the attributes nor the relations
in the captions. The human annotation study in this supplement tests caption validity directly.

\subsection{Sensitivity analyses and their coverage}
\label{app:retrieval-sensitivity}

\Cref{app:tab:coverage} lists each analysis and the models it covers. The
neighborhood scales and the 16 MPNet assignments cover the full 22-model
panel. The allocation, cohort, query, and reference-precision analyses were run
on the representative models, and the ranking analysis on the 16-model
panel for which it was specified.

\begin{table}[htbp]
\centering
\caption{Coverage of the retrieval-study sensitivity analyses. ``Representative''
denotes SigLIP B/16 224, CLIP B/16 224, and SigLIP 2 B/16 224.}
\label{app:tab:coverage}
\small
\setlength{\tabcolsep}{3.5pt}
\begin{tabularx}{\linewidth}{@{}p{0.31\linewidth}p{0.26\linewidth}Yl@{}}
\toprule
Analysis & Checkpoints & What varies & Location \\
\midrule
Scales \(k=10,20,100\) & All 22 & Graph scale & \Cref{app:tab:scales} \\
16 MPNet assignments & All 22 & Semantic assignment & \Cref{app:tab:scales} \\
128 cost-matched alternatives & Representative & Allocation at matched cost &
\Cref{app:retrieval-assignment} \\
Annotation selector (MPNet tie-break) & SigLIP & Selector &
\Cref{app:tab:representative,app:tab:robust-primary} \\
25 rebuilt cohorts & Representative & Cohort, graphs, assignments, references &
\Cref{app:retrieval-cohort} \\
Source-row resampling & SigLIP & Queries & \Cref{app:retrieval-precision} \\
1,000-draw reference distributions & Representative & Random draws &
\Cref{app:retrieval-precision} \\
Retention-ranking analysis & 16-model panel & Query groups, assignments &
\Cref{app:retrieval-ranking} \\
\bottomrule
\end{tabularx}
\end{table}

\begin{table}[tbp]
\centering
\caption{Neighborhood-scale and assignment sensitivity for all 22 models.
Left: \(\rho_{A_k}\) at each graph scale under the canonical assignment; at
every scale it exceeds \(\rho_{\mathrm{R@10}}\) in \Cref{app:tab:panel}
(22/22 models). Right: ranges over the canonical MPNet assignment
and its 15 cost-jittered alternatives
(\Cref{app:tab:alternatives}). The ordering
\(\rho_{A_{50}}>\rho_{\mathrm{R@10}}>\rho_{\mathrm{R@1}}\) holds in
352 of 352 model--assignment combinations.}
\label{app:tab:scales}
\scriptsize
\setlength{\tabcolsep}{3.2pt}
\begin{tabular}{@{}lrrrrccc@{}}
\toprule
 & \multicolumn{4}{c}{\(\rho_{A_k}\), canonical assignment} & \multicolumn{3}{c}{Range over 16 MPNet assignments} \\
\cmidrule(lr){2-5}\cmidrule(l){6-8}
Model & \(k=10\) & \(k=20\) & \(k=50\) & \(k=100\) & \(\rho_{A_{50}}\) & \(\rho_{\mathrm{R@10}}\) & \(\rho_{\mathrm{R@1}}\) \\
\midrule
CLIP B/32 224 & 0.7277 & 0.7474 & 0.8081 & 0.8442 & 0.806--0.808 & 0.469--0.473 & 0.148--0.155 \\
CLIP B/16 224 & 0.7196 & 0.7608 & 0.8091 & 0.8476 & 0.806--0.809 & 0.486--0.491 & 0.130--0.138 \\
CLIP L/14 224 & 0.7110 & 0.7567 & 0.8116 & 0.8446 & 0.808--0.812 & 0.483--0.487 & 0.126--0.129 \\
CLIP L/14 336 & 0.7149 & 0.7607 & 0.8160 & 0.8474 & 0.812--0.816 & 0.482--0.487 & 0.123--0.126 \\
\midrule
SigLIP B/16 224 & 0.6938 & 0.7281 & 0.7787 & 0.8145 & 0.777--0.781 & 0.470--0.474 & 0.096--0.099 \\
SigLIP B/16 256 & 0.6861 & 0.7286 & 0.7825 & 0.8174 & 0.781--0.785 & 0.471--0.474 & 0.088--0.094 \\
SigLIP B/16 384 & 0.6874 & 0.7224 & 0.7804 & 0.8158 & 0.779--0.782 & 0.476--0.478 & 0.094--0.098 \\
SigLIP L/16 256 & 0.6610 & 0.7170 & 0.7699 & 0.8024 & 0.767--0.771 & 0.472--0.475 & 0.085--0.088 \\
SigLIP L/16 384 & 0.6706 & 0.7144 & 0.7668 & 0.8015 & 0.764--0.767 & 0.478--0.481 & 0.084--0.085 \\
SigLIP So400m/14 224 & 0.6770 & 0.7259 & 0.7795 & 0.8108 & 0.776--0.779 & 0.473--0.476 & 0.085--0.088 \\
\midrule
SigLIP 2 B/16 224 & 0.6849 & 0.7265 & 0.7833 & 0.8126 & 0.781--0.784 & 0.475--0.477 & 0.096--0.097 \\
SigLIP 2 B/16 256 & 0.6887 & 0.7295 & 0.7864 & 0.8137 & 0.784--0.787 & 0.476--0.477 & 0.098--0.099 \\
SigLIP 2 B/16 384 & 0.6857 & 0.7209 & 0.7859 & 0.8151 & 0.784--0.786 & 0.473--0.475 & 0.095--0.097 \\
SigLIP 2 L/16 256 & 0.6764 & 0.7133 & 0.7666 & 0.8060 & 0.763--0.767 & 0.465--0.469 & 0.086--0.088 \\
SigLIP 2 L/16 384 & 0.6756 & 0.7141 & 0.7692 & 0.8062 & 0.767--0.770 & 0.460--0.463 & 0.085--0.087 \\
SigLIP 2 So400m/14 224 & 0.6752 & 0.7163 & 0.7656 & 0.8022 & 0.764--0.767 & 0.461--0.464 & 0.085--0.089 \\
\midrule
E5-V & 0.6119 & 0.6642 & 0.7422 & 0.7899 & 0.741--0.742 & 0.529--0.532 & 0.110--0.113 \\
Jina-CLIP-v2 & 0.6420 & 0.6904 & 0.7629 & 0.8073 & 0.762--0.764 & 0.519--0.521 & 0.123--0.127 \\
LongCLIP-GmP ViT-L/14 & 0.6046 & 0.6608 & 0.7449 & 0.7886 & 0.744--0.745 & 0.528--0.531 & 0.105--0.108 \\
LongCLIP-L (Diffusers) & 0.6907 & 0.7396 & 0.7971 & 0.8318 & 0.794--0.797 & 0.506--0.507 & 0.113--0.117 \\
Qwen3-VL-Embedding-2B & 0.6135 & 0.6624 & 0.7344 & 0.7781 & 0.733--0.735 & 0.500--0.503 & 0.083--0.084 \\
Qwen3-VL-Embedding-8B & 0.5958 & 0.6427 & 0.7193 & 0.7650 & 0.718--0.720 & 0.489--0.491 & 0.073--0.074 \\
\bottomrule
\end{tabular}
\end{table}

\subsubsection{Assignment sensitivity}
\label{app:retrieval-assignment}

\paragraph{16 MPNet assignments.}
Across the canonical assignment and its 15 cost-jittered alternatives, the
ordering \(\rho_{A_{50}}>\rho_{\mathrm{R@10}}>\rho_{\mathrm{R@1}}\) holds for
every model and every assignment (\Cref{app:tab:scales}). For SigLIP B/16
224, all 16 assignments exceed the random-reference mean for corrected agreement,
R@10, and mean reciprocal rank; the prespecified gate required at least 14 for
agreement and R@10. The annotation assignment has the same direction for all
three outcomes.

\paragraph{128 cost-matched alternatives.}
Random re-pairing changes both the semantic cost of each assigned pair and the
global allocation. This control asks whether the canonical allocation remains
exceptional among allocations with approximately the same cost profile. It was
frozen separately and constructed from selector costs alone. Define canonical
overlap \(o(p,p^\star)=n^{-1}\sum_i\mathbf 1[p(i)=p^\star(i)]\) and the
constrained frontier
\(\Phi(\gamma)=\min_{p\in\mathcal P_1:\,o(p,p^\star)\le\gamma}\sum_iC_{i,p(i)}\).
Penalized assignments
\(L(\lambda)=\min_{p\in\mathcal P_1}\sum_i[C_{i,p(i)}+\lambda\mathbf
1[p(i)=p^\star(i)]]\) give the dual lower bound
\(L(\lambda)-\lambda\gamma n\le\Phi(\gamma)\). At the primary target
\(\gamma=0.50\), the feasible cost ratio is 1.006704 and the dual lower bound is
1.006633; the target \(\gamma=0.25\) requires ratio 1.026166, outside the 1\%
budget. The generator solves
\(p_b=\arg\min_{p\in\mathcal P_1}\sum_i[C_{i,p(i)}+0.014\,\mathbf
1[p(i)=p^\star(i)]+\epsilon^{(b)}_{i,p(i)}]\) with independent
\(\epsilon^{(b)}_{ij}\sim\mathrm{Unif}[-h,h]\). The specified in advance schedule was
\(h\in\{0.005,0.008,0.010,0.012\}\); 0.008 was the first passing scale, and the
first 128 qualifying assignments were retained. Eligibility required mean cost
ratio at most 1.01, normalized sorted-cost Wasserstein error at most 0.02,
largest normalized interior-decile error at most 0.05, canonical overlap at
most 0.50, and pairwise overlap at most 0.80. The retained family has mean cost
ratios 1.00750--1.00854, Wasserstein errors 0.00810--0.00944, decile errors
0.01193--0.02127, and canonical overlap 0.4810--0.4998, and every two
alternatives differ on at least 1,006 targets. All are complete derangements
with the original exclusions. The family matches the distribution of assigned
costs, not each image's cost.

\begin{table}[htbp]
\centering
\caption{Canonical MPNet assignment versus the 128 cost-matched alternatives.
Intervals are the 2.5--97.5\% quantiles of the family, and ``Percentile'' is the
fraction of the 128 values no greater than canonical. The generator is not a
uniform conditional sampler, so these are not randomization \(p\)-values.}
\label{app:tab:matched}
\small
\begin{tabular}{@{}llrrrr@{}}
\toprule
Model & Metric & Canonical & Mean & Central 95\% & Percentile \\
\midrule
SigLIP B/16 224 & \(A_{50}\) & 0.216984 & 0.217240 & [0.215705, 0.218649] & 0.398 \\
SigLIP B/16 224 & \(\apcorr{50}\) & 0.778684 & 0.779647 & [0.773874, 0.784948] & 0.398 \\
SigLIP B/16 224 & I\(\to\)T R@10 & 0.3844 & 0.3809 & [0.3758, 0.3854] & 0.891 \\
SigLIP B/16 224 & T\(\to\)I R@10 & 0.3852 & 0.3868 & [0.3820, 0.3918] & 0.258 \\
CLIP B/16 224 & \(A_{50}\) & 0.226720 & 0.226658 & [0.225074, 0.228201] & 0.508 \\
CLIP B/16 224 & \(\apcorr{50}\) & 0.809054 & 0.808822 & [0.802910, 0.814583] & 0.508 \\
CLIP B/16 224 & I\(\to\)T R@10 & 0.3522 & 0.3565 & [0.3514, 0.3608] & 0.055 \\
CLIP B/16 224 & T\(\to\)I R@10 & 0.3384 & 0.3377 & [0.3327, 0.3420] & 0.617 \\
\bottomrule
\end{tabular}
\end{table}

Every canonical value lies inside its central 95\% range
(\Cref{app:tab:matched}). Monte Carlo standard errors are at most 0.000281 for
corrected agreement and 0.000233 for R@10. The same neighborhood-agreement and
R@10 regime therefore appears without the exact canonical allocation. For
SigLIP 2 B/16 224, the same 128 alternatives give corrected agreement
0.7771--0.7984, image-to-text R@10 35.94--37.22\%, and R@1 3.70--4.24\%, against
random-reference maxima of 0.0028, 0.42\%, and 0.08\%. Its canonical value lies
inside the central 95\% range for agreement, image-to-text R@1 and mean
reciprocal rank, and text-to-image R@10.

\subsubsection{Cohort and graph sensitivity}
\label{app:retrieval-cohort}

The prespecified reconstruction partitions the 5,000 identities
deterministically into 25 groups of 200. Each reconstruction removes one group,
keeps the same 4,800 image and caption identities in every condition, rebuilds
the neighbor graphs, re-solves the MPNet and annotation assignments inside the
retained cohort, and draws 256 new random re-pairings. Full-cohort values use
the 1,000-draw references. The contrast is the hard negative value minus the
within-cohort random-reference mean, in chance-corrected agreement at \(k=50\).
For the \(G=25\) reconstruction estimates \(\hat\theta_{(-g)}\), we report the
jackknife-scaled interval, the full-cohort estimate plus or minus
\(1.96\,\mathrm{SE}_{\mathrm{JK}}\), with
\begin{equation}
 \mathrm{SE}_{\mathrm{JK}}=\sqrt{\frac{G-1}{G}\sum_{g=1}^{G}
 (\hat\theta_{(-g)}-\bar\theta)^2},\qquad
 \bar\theta=\frac1G\sum_g\hat\theta_{(-g)}.
 \label{app:eq:jackknife}
\end{equation}
It summarizes sensitivity to this deterministic pipeline perturbation, not
population sampling uncertainty.

\begin{table}[htbp]
\centering
\caption{Contrasts at \(k=50\) across the 25 rebuilt cohorts. Every contrast is
positive in all 25 reconstructions, and every jackknife-scaled interval
excludes zero; this is not a population significance test. The SigLIP
annotation contrast uses the MPNet tie-break.}
\label{app:tab:robust-primary}
\scriptsize
\setlength{\tabcolsep}{3pt}
\begin{tabular}{@{}lrrrrrr@{}}
\toprule
Contrast & Full & Mean & Median & Range & Jackknife SE & JK-scaled interval \\
\midrule
SigLIP MPNet & 0.778680 & 0.779902 & 0.779307 & [0.772023, 0.787279] & 0.018629 & [0.742167, 0.815192] \\
SigLIP annotation & 0.650649 & 0.652019 & 0.652536 & [0.642217, 0.658561] & 0.021803 & [0.607916, 0.693381] \\
CLIP MPNet & 0.809074 & 0.810852 & 0.810292 & [0.801860, 0.821293] & 0.022201 & [0.765561, 0.852586] \\
\bottomrule
\end{tabular}
\end{table}

\begin{table}[htbp]
\centering
\caption{Every reconstruction estimate at \(k=50\). The group index identifies
the removed group of 200 identities; M and A denote the MPNet and annotation
selectors.}
\label{app:tab:all-25}
\footnotesize
\setlength{\tabcolsep}{2.8pt}
\begin{tabular}{@{}rrrr@{\hspace{1.0em}}rrrr@{}}
\toprule
Group & SigLIP M & SigLIP A & CLIP M & Group & SigLIP M & SigLIP A & CLIP M \\
\midrule
0  & 0.780692 & 0.653811 & 0.801860 & 13 & 0.778899 & 0.642391 & 0.812611 \\
1  & 0.778360 & 0.651058 & 0.806480 & 14 & 0.776389 & 0.648981 & 0.807505 \\
2  & 0.785368 & 0.652319 & 0.810292 & 15 & 0.784254 & 0.649576 & 0.814846 \\
3  & 0.772023 & 0.655846 & 0.807835 & 16 & 0.780897 & 0.656111 & 0.817357 \\
4  & 0.783694 & 0.653232 & 0.812439 & 17 & 0.780897 & 0.655069 & 0.817539 \\
5  & 0.778497 & 0.653076 & 0.811210 & 18 & 0.783314 & 0.656556 & 0.813766 \\
6  & 0.782534 & 0.658561 & 0.807740 & 19 & 0.783191 & 0.657612 & 0.817824 \\
7  & 0.778063 & 0.646461 & 0.806445 & 20 & 0.781546 & 0.658101 & 0.807825 \\
8  & 0.773015 & 0.652536 & 0.805187 & 21 & 0.787279 & 0.649994 & 0.814045 \\
9  & 0.779307 & 0.657843 & 0.811582 & 22 & 0.784294 & 0.652639 & 0.821293 \\
10 & 0.778351 & 0.646542 & 0.808393 & 23 & 0.773585 & 0.650013 & 0.809536 \\
11 & 0.776658 & 0.650107 & 0.812713 & 24 & 0.777737 & 0.642217 & 0.806259 \\
12 & 0.778702 & 0.649833 & 0.808719 & -- & -- & -- & -- \\
\bottomrule
\end{tabular}
\end{table}

\begin{table}[htbp]
\centering
\caption{Secondary neighborhood scales in the rebuilt cohorts. Every estimate is
positive in all 25 reconstructions. These scales are descriptive sensitivity
analyses.}
\label{app:tab:secondary-k}
\small
\begin{tabular}{@{}lrrrr@{}}
\toprule
Contrast & \(k\) & Full cohort & Reconstruction mean & Reconstruction range \\
\midrule
SigLIP MPNet & 10 & 0.693821 & 0.695438 & [0.684960, 0.710334] \\
 & 20 & 0.728141 & 0.727194 & [0.718173, 0.738984] \\
 & 100 & 0.814513 & 0.814160 & [0.808764, 0.820831] \\
SigLIP annotation & 10 & 0.499492 & 0.503693 & [0.483831, 0.513656] \\
 & 20 & 0.558116 & 0.560517 & [0.542462, 0.570559] \\
 & 100 & 0.705217 & 0.705341 & [0.694120, 0.710384] \\
CLIP MPNet & 10 & 0.719648 & 0.720228 & [0.709879, 0.736996] \\
 & 20 & 0.760792 & 0.762811 & [0.751077, 0.775702] \\
 & 100 & 0.847623 & 0.846807 & [0.839961, 0.853999] \\
\bottomrule
\end{tabular}
\end{table}

Re-solving inside a retained cohort changes 14--18\% of eligible targets in
every reconstruction (MPNet: 14.35--17.54\%, overlap with the full-cohort
assignment 0.8246--0.8565; annotation: 14.36--17.96\%, overlap 0.8204--0.8564),
so the reconstructions also vary the assignment. The full-cohort reconstruction
audit reproduced all 12 stored agreement values across the four \(k\) with
maximum absolute error \(1.11\times10^{-16}\).

For SigLIP 2 B/16 224, the same 25 cohorts use the frozen within-cohort MPNet
assignment, rebuilt graphs, and 256 new random re-pairings each
(\Cref{app:tab:siglip2-cohort}). Every reconstruction is positive at every
scale.

\begin{table}[htbp]
\centering
\caption{SigLIP 2 B/16 224 contrasts in chance-corrected agreement. ``Full''
subtracts the 1,000-draw full-cohort reference mean; each reconstruction
subtracts its own 256-draw mean.}
\label{app:tab:siglip2-cohort}
\small
\begin{tabular}{@{}rrrrr@{}}
\toprule
\(k\) & Full & Reconstruction range & Jackknife-scaled interval & Positive \\
\midrule
10 & 0.68487 & [0.67268, 0.70587] & [0.60276, 0.76698] & 25/25 \\
20 & 0.72654 & [0.72358, 0.74396] & [0.67524, 0.77783] & 25/25 \\
50 & 0.78330 & [0.77983, 0.79354] & [0.74690, 0.81971] & 25/25 \\
100 & 0.81264 & [0.80716, 0.81929] & [0.78662, 0.83867] & 25/25 \\
\bottomrule
\end{tabular}
\end{table}

\subsubsection{Query resampling and reference precision}
\label{app:retrieval-precision}

\paragraph{Source-row resampling.}
Holding galleries and assignments fixed, each of 2,000 draws resamples the
5,000 saved source rows with replacement, using the same indices for the MPNet
assignment and the one-draw random assignment
(\Cref{app:tab:row-resampling}). Image-to-text rows pair the same image query
across conditions. Text-to-image rows are indexed by source image \(i\), with
text query \(p(i)\), so paired rows can represent different text queries. The
two directions use separate seeded draws, and the endpoints are the 2.5th and
97.5th percentiles. These intervals describe conditional source-row sensitivity,
not uncertainty over assignments or populations.

\begin{table}[htbp]
\centering
\caption{SigLIP B/16 224: 2,000 paired fixed-gallery source-row resamples per
direction. The contrast is the MPNet assignment minus the one-draw random
assignment.}
\label{app:tab:row-resampling}
\small
\begin{tabular}{@{}llrr@{}}
\toprule
Direction & Contrast & Estimate & 95\% interval \\
\midrule
Image \(\to\) text & R@10 (points) & \(+38.38\) & [37.04, 39.76] \\
Text \(\to\) image & R@10 (points) & \(+38.40\) & [37.14, 39.80] \\
Image \(\to\) text & Mean reciprocal rank & \(+0.1520\) & [0.1458, 0.1585] \\
Text \(\to\) image & Mean reciprocal rank & \(+0.1594\) & [0.1529, 0.1667] \\
\bottomrule
\end{tabular}
\end{table}

\paragraph{Reference distributions.}
For SigLIP B/16 224, none of the 1,000 core draws reached the hard negative
value for corrected agreement, image-to-text R@10, or mean reciprocal rank; the
add-one upper-tail frequency is \(1/1001\). This is a descriptive comparison with
the generator, not a randomization \(p\)-value. At \(k=50\), the SigLIP reference
mean of corrected agreement is \(4.09\times10^{-6}\) (SD \(8.64\times10^{-4}\),
Monte Carlo SE \(2.73\times10^{-5}\)), and the CLIP mean is
\(-1.98\times10^{-5}\) (SD \(8.96\times10^{-4}\), Monte Carlo SE
\(2.83\times10^{-5}\)); across rebuilt cohorts, the largest Monte Carlo SE is
\(6.23\times10^{-5}\). \Cref{app:tab:siglip2-null} gives the SigLIP 2 reference;
every SigLIP 2 hard negative outcome lies beyond the range of all 1,000 draws.

\begin{table}[htbp]
\centering
\caption{SigLIP 2 B/16 224 reference distribution over 1,000 random re-pairings.}
\label{app:tab:siglip2-null}
\small
\begin{tabular}{@{}lrrrr@{}}
\toprule
Outcome & Mean & Central 95\% & Maximum & MC SE of mean \\
\midrule
\(\apcorr{50}\) & \(-0.000017\) & [\(-0.001714\), 0.001699] & 0.002811 & 0.000028 \\
Image-to-text R@1 & 0.0108\% & [0.0000, 0.0400]\% & 0.0800\% & 0.00046 pp \\
Image-to-text R@10 & 0.1828\% & [0.0800, 0.3000]\% & 0.4200\% & 0.00187 pp \\
Image-to-text MRR & 0.001696 & [0.001349, 0.002146] & 0.002633 & 0.0000068 \\
Text-to-image R@10 & 0.1809\% & [0.0800, 0.3200]\% & 0.3600\% & 0.00187 pp \\
\bottomrule
\end{tabular}
\end{table}

\subsubsection{Retention rankings across the 16-model panel}
\label{app:retrieval-ranking}

The prespecified ranking analysis asks whether the three retentions order the
16 CLIP, SigLIP, and SigLIP 2 models differently. It was not extended to the
six additional encoders. We use \emph{broad lineage} for CLIP versus the SigLIP
lineage and keep the three prespecified family labels for the inversion
analysis; SigLIP 2 is a newer generation within the SigLIP lineage. A robust
inversion requires both paired model-difference intervals, from the
fixed-gallery delete-group analysis over 25 query groups of 200, to exclude zero.

\begin{table}[htbp]
\centering
\caption{Prespecified retention-ranking analysis over the 16-model panel.
``Cross-family'' uses the three labels CLIP, SigLIP, and SigLIP 2.}
\label{app:tab:rankings}
\small
\setlength{\tabcolsep}{4.5pt}
\begin{tabular}{@{}lrrrrrr@{}}
\toprule
Retention pair & \(\tau_b\) & \(\rho_s\) & Inversions & Cross-family & Robust & Robust cross-family \\
\midrule
\(A_{50}\) vs.\ R@10 & 0.4833 & 0.6529 & 31 & 16 & 2 & 0 \\
\(A_{50}\) vs.\ R@1 & 0.6667 & 0.8765 & 20 & 7 & 1 & 0 \\
R@10 vs.\ R@1 & 0.3500 & 0.4147 & 39 & 23 & 4 & 0 \\
\bottomrule
\end{tabular}
\end{table}

No cross-family inversion among the three retentions meets the robustness
criterion (\Cref{app:tab:rankings}; \(\rho_s\) is Spearman's coefficient). The
rank associations are nevertheless imperfect, and seven robust within-family
inversions remain: for \(A_{50}\) versus R@10, SigLIP B/16 224 and B/16 256 each
reverse order with SigLIP L/16 384; for \(A_{50}\) versus R@1, CLIP B/32 224
reverses with CLIP L/14 336; and for R@10 versus R@1, CLIP B/32 224 reverses with
CLIP B/16 224, L/14 224, and L/14 336, and SigLIP B/16 224 with SigLIP L/16 384.
These are seven metric-pair occurrences rather than seven distinct model
pairs. The four CLIP models form the Pareto front over the three
retentions, which describes the profile rather than defining a leaderboard.

Across the 25 query groups, the Kendall \(\tau_b\) replicate ranges are
[0.3333, 0.5167], [0.6000, 0.7333], and [0.2333, 0.4500] for the three pairs,
with jackknife standard errors 0.2105, 0.2049, and 0.2820. These quantify
sensitivity to the fixed query cohort; the 16 correlated models are not
treated as an independent sample of models. Repeating the analysis across the 16
MPNet assignments gives \(\tau_b\) ranges [0.3833, 0.4833], [0.6000, 0.7000],
and [0.3000, 0.4000], with 31--37, 18--24, and 36--42 inversions, so the pattern
is not specific to the canonical assignment. Original agreement and retention
are distinct model descriptors: original \(A_{50}\) has Kendall \(\tau_b\)
0.5833, 0.3333, and 0.5500 with the three retentions, with 11, 6, and 9 robust
cross-family inversions.

\FloatBarrier

%% file: sec/supp/d_followup.tex
\subsection{Resolution on Flickr30k}
With MPNet, the median half-retention recall cutoff is 12.83 on validation and 13.79 on test.
For $A_{50}$, median half-retention similarities are 0.462 and 0.443. For CKA between the two towers they are 0.429 and 0.416.
Complete MPNet re-pairing retains 0.498--0.616 of the agreement margin on validation and 0.519--0.633 on test. Recall@10 ranges are 0.422--0.595 and 0.429--0.568. Recall@1 ranges are 0.035--0.066 and 0.035--0.069.
The image-side re-pairings repeat the similarity ladder with DINOv2 vectors. The judge mark at 0.55 in \cref{fig:resolution}b is based on individual-pair similarity (663 wrong verdicts among 736 below the threshold in the canonical COCO MPNet pairing). The horizontal axis gives assignment means. The terminal point is the random reference.
On COCO, at matched cutoff ten, neighborhood agreement retains more than bidirectional Recall@10 for each of the 22 models (\cref{app:tab:scales}).

\subsection{Human annotation and model judge}
\label{supp:annotation}
\paragraph{Annotation statement.}
Two authors independently labeled 500 public COCO image--caption pairs using a common rubric. Both agreed to the use of their labels in this research. The analysis uses their judgments about the pairs.

\paragraph{Rubric and sampling.}
A caption is valid if its factual claims agree with the image. A contradicted detail gives wrong detail, a contradicted main subject or scene gives wrong content, and unresolved claims without a visible contradiction give unsure.
Each annotator received a separately shuffled batch with opaque identifiers. Each item showed one image and one assigned caption with the condition and model verdict hidden. The canonical MPNet sample used 40 pairs per similarity-quintile and model-verdict stratum. We weight the estimates by the population sizes of these strata. Fifty original and fifty random pairs served as checks. The four-label agreement across the 500 pairs was $\kappa=0.854$.
On the ten sampled originals flagged by Qwen, both annotators agreed on three wrong details, five valid captions, and two unsure.
Across all 50 original checks, both independently called 3 wrong, and either called 4 wrong.
\input{tab/human_rates}

\paragraph{Retrieval by human label.}
\Cref{tab:human-retrieval-labels} splits the 400 substitutions by their two independent labels and compares original and re-paired retrieval on the same images in each group.
\input{tab/human_retrieval_labels}

\paragraph{Judge calibration.}
Qwen3.8-27B evaluated each pair twice using paraphrased prompts. We used bfloat16, temperature zero, disabled thinking, and at most 256 output tokens per prompt. Each asked the model to compare one image and caption against the same four labels and return a structured verdict. Both prompts must mark the caption wrong for the binary wrong verdict.
When both return wrong labels but differ on subtype, the subtype is recorded as disputed.
The judge's weighted precision against the adjudicated calibration labels is 86.9\% (stratified 95\% interval 82.0--91.3\%) and recall is 93.2\%. Counting either human vote yields 90.2\% precision and requiring both yields 85.9\%.
The judge marked 57 of 500 originals and 497 of 500 random pairs wrong. Both the precision and original-pair checks miss their limits. OpenAI ChatGPT adjudicated the 52 disagreements between the human annotators for judge calibration. The human-only calculations use the independent votes.

\paragraph{Model-judged re-pairings.}
Each round excludes the selected edges that fail the judge rule and solves the assignment again. This continues until the judge marks every selected caption wrong. The single-caption solver reached a complete pairing after 152 rounds, with selector similarity 0.657 and cost 1.094 times the unconstrained MPNet pairing. The caption-group re-pairing reached a complete pairing with a wrong model verdict for each of 25,014 captions.
The full assignments are model-judged. The existing human labels overlap 129 single-caption edges and 20 group-caption edges. These sampled edges belong to the canonical re-pairing or the earlier checks and do not estimate the final probability of a human wrong verdict.

\paragraph{Human-labeled CKA subset.}
All three bars in \cref{fig:cka-curve}c use the same 195 images. The original- and substituted-caption bars give paired delete-one jackknife intervals with fixed subset bandwidths. The random bar gives the central 95\% of 256 draws from the same substituted caption pool. The smaller sample gives a distinct CKA reference from the full-cohort results. The full-cohort random ranges in \cref{fig:cka-curve}a--b span the available 256 draws per fidelity.

On the 195 consensually wrong canonical substitutions, the paired difference in CKA from the original captions is $-0.040$ (95\% interval $[-0.058,-0.022]$). The gain over random permutations of the same substituted caption pool is $+0.375$ (95\% interval $[+0.344,+0.406]$).
The judge-valid matching analysis increases exact accuracies from 56.6, 64.4, and 61.0\% to 61.0, 68.0, and 66.0\%, respectively.

\begin{figure}[htbp]
\centering\includegraphics[width=\textwidth]{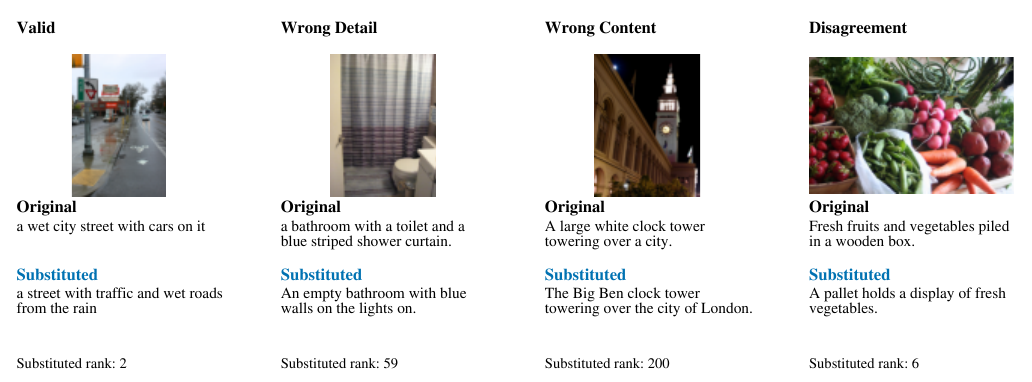}
\caption{\textbf{Human judgments distinguish wrong details from wrong content.}
One labeled substitution per category, chosen by a fixed rule (Supp.~D). Ranks are for image-to-text retrieval with SigLIP B/16 224.}
\label{fig:human-examples}
\end{figure}

\paragraph{Examples and image licenses.}
For \cref{fig:human-examples}, we take the first eligible labeled image in stored order for each category, restricting to images without annotated people. Ranks are for each substituted caption among 5,000 captions.
\begin{itemize}[nosep]
\item Labeled examples: COCO image IDs 391648 (license 3), 23781 (4), 357567 (4), and 93154 (5).
\item Teaser: COCO image IDs 407002 (license 4), 347265 (1), 139872 (4), and 511398 (1).
\end{itemize}

\begin{figure}[htbp]
\centering\includegraphics[width=.82\linewidth]{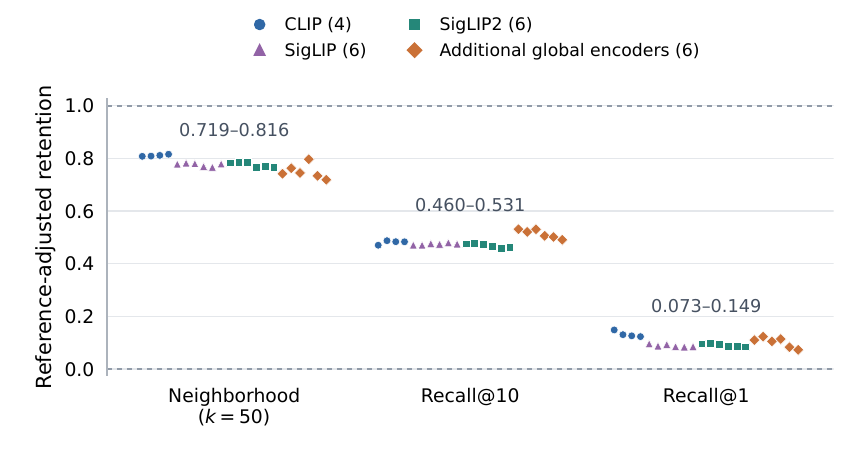}
\caption{\textbf{Neighborhood agreement retains more of its margin than recall on COCO.} Retentions across the 22 models.}
\label{fig:22-ckpt}
\end{figure}
\begin{figure}[htbp]
\centering\includegraphics[width=.8\linewidth]{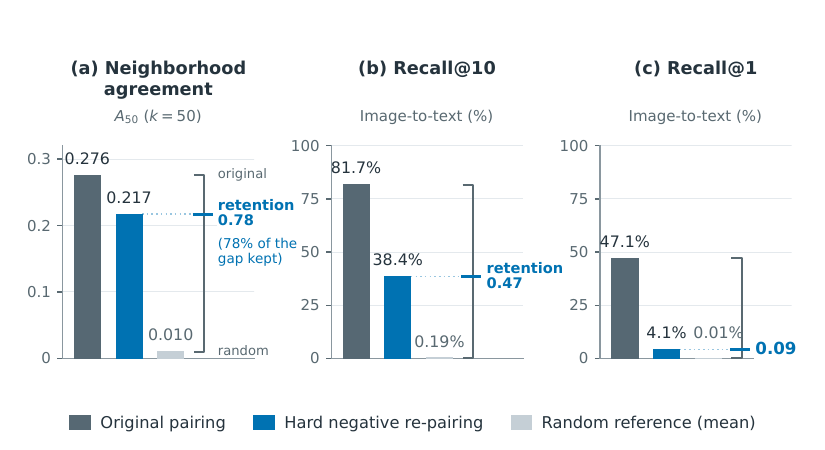}
\caption{\textbf{Retention across diagnostics for SigLIP B/16 224.} The image-to-text comparison uses the original and complete MPNet pairings.}
\label{fig:retention-example}
\end{figure}

%% file: tab/human_rates.tex
\begin{table}[t]
\centering\small
\caption{\textbf{Human wrong-label rates vary with selector similarity.} Population-weighted percentages by quintile, with the two original independent votes.}
\label{tab:human-rates}
\begin{tabular}{@{}lrr@{}}
\toprule
Quintile & Both wrong (\%) & Either wrong (\%) \\
\midrule
All & 61.2 & 65.7 \\
1 & 80.5 & 86.9 \\
2 & 74.3 & 79.4 \\
3 & 63.2 & 65.7 \\
4 & 53.7 & 57.2 \\
5 & 34.2 & 39.1 \\
\bottomrule
\end{tabular}
\end{table}

%% file: tab/human_retrieval_labels.tex
\begin{table}[htbp]
\centering\small
\caption{\textbf{Retrieval and agreement by the two independent human labels.} The first four rows require both annotators to give the same label to the substituted caption. Each metric gives the median of population-weighted means over 22 models. Recall is bidirectional (\%).}
\label{tab:human-retrieval-labels}
\begin{tabular}{@{}lrrrrrrr@{}}
\toprule
Label & $n$ & Orig. R@1 & Orig. R@10 & Orig. $A_{50}$ & Re-paired R@1 & Re-paired R@10 & Re-paired $A_{50}$ \\
\midrule
Valid & 123 & 35.6 & 77.5 & 0.331 & 11.6 & 61.3 & 0.288 \\
Wrong detail & 148 & 52.9 & 87.4 & 0.283 & 3.5 & 35.3 & 0.229 \\
Wrong content & 39 & 64.0 & 85.4 & 0.199 & 0.0 & 10.9 & 0.123 \\
Unsure & 42 & 46.4 & 79.9 & 0.268 & 3.0 & 45.0 & 0.203 \\
Different labels & 48 & 43.1 & 77.6 & 0.278 & 4.9 & 43.7 & 0.208 \\
\bottomrule
\end{tabular}
\end{table}

%% file: sec/supp/e_selection.tex
\section{Model selection}
\label{app:selection}

This appendix supports \autoref{sec:study-selection}. The Flickr30k study, which provides the
held-out evidence, comes first: its protocol, the complete validation and test
panel, the paired bootstrap interval, and the five-caption evaluation. The
retrospective COCO analyses and the post hoc follow-ups come after it.

\subsection{Flickr30k protocol}
\label{app:flickr-protocol}

\paragraph{Protocol fixed before outcomes.}
The study uses Flickr30k \citep{young2014flickr30k} with the standard Karpathy
split: all 1,014 validation images and 1,000 test images, excluding the
training split. The 16-model roster
(\Cref{app:tab:roster}), revisions, extraction settings, caption rule, metrics,
selection rules, and analysis were fixed on 2026-09-09 at 00:09:23 UTC, before
any representation was extracted or any outcome examined. Validation metrics and
selected models were saved at 00:44:03 UTC, before any test outcome was
computed, and the completed result was recorded at 00:46:17 UTC. This is a
prospective protocol within the project, not an external preregistration. The
study evaluates model selection on original pairs; it involves no
re-pairing and no CKA.

\paragraph{Data order and captions.}
Images are ordered by numeric file identity within validation and then within
test, and each image's five captions are ordered by sentence ID. The primary
analysis selects one caption per image with the outcome-independent rule and
seed of the COCO cohort (seed 20260805). The caption graph uses the same
selected captions as retrieval.

\paragraph{Selection rules.}
On validation images only, the agreement rule selects the model with the
highest \(A_{50}\), and the recall rule selects the model with the highest
bidirectional one-caption R@1. Both modality graphs use exact
self-excluding neighbors with \(k=50\). Exact model ties would go to the
first roster entry and score ties to the first gallery index; every reported
maximum is unique. Test images never select a model, and no reverse
orientation is evaluated.

\paragraph{Five-caption evaluation.}
The prespecified secondary metric lets each image query rank all five captions
of every image in its split and counts success when its top caption belongs to
that image; each caption query ranks all images and succeeds when its own image
ranks first. The two directions receive equal weight. A second recall rule maximizes
this validation metric, while the agreement rule keeps the one-caption \(A_{50}\).
The primary selected models are also evaluated on this five-caption test
metric.

\paragraph{Extraction controls.}
Native forward reconstruction and deterministic re-encoding are checked for
every model. The normalized float32 caches share ordered image identities,
captions, and selected caption indices across the roster. Text limits are the
native ones, including 16 tokens for SigLIP So400m/14 224 and 64 for SigLIP 2
So400m/14 224.

\subsection{Complete Flickr30k panel}
\label{app:flickr-panel}

\Cref{app:tab:flickr-panel} extends \autoref{tab:flickr30k-checkpoint-selection} with the test \(A_{50}\)
and both five-caption recalls.

\begin{table}[htbp]
\centering
\caption{Complete Flickr30k validation and test panel. \(A_{50}\) is a fraction;
R@1 columns are bidirectional percentages. ``One'' uses the selected caption,
and ``Five'' counts all five captions of an image as positives. Bold validation
values are the selected maxima; test values do not select models.
Validation has 1,014 images and test 1,000, and each five-caption text gallery
is five times as large. Absolute values are not comparable with the larger COCO
galleries.}
\label{app:tab:flickr-panel}
\footnotesize
\setlength{\tabcolsep}{3.1pt}
\begin{tabular}{@{}lrrrrrr@{}}
\toprule
 & \multicolumn{3}{c}{Validation} & \multicolumn{3}{c}{Test} \\
\cmidrule(lr){2-4}\cmidrule(l){5-7}
Model & \(A_{50}\) & One R@1 & Five R@1 & \(A_{50}\) & One R@1 & Five R@1 \\
\midrule
CLIP B/32 224 & 0.202288 & 62.72 & 69.20 & 0.201140 & 59.50 & 69.12 \\
CLIP B/16 224 & 0.207160 & 66.37 & 72.99 & 0.207600 & 61.40 & 72.32 \\
CLIP L/14 224 & 0.207968 & 69.77 & 77.25 & 0.211140 & 65.40 & 74.99 \\
CLIP L/14 336 & 0.214162 & 71.94 & 79.77 & 0.211780 & 67.95 & 77.27 \\
\midrule
SigLIP B/16 224 & 0.231085 & 75.49 & 82.33 & 0.223760 & 74.35 & 81.88 \\
SigLIP B/16 256 & \textbf{0.232978} & 76.13 & 83.47 & 0.228000 & 75.20 & 82.78 \\
SigLIP B/16 384 & 0.230888 & 78.90 & 85.12 & 0.227240 & 78.25 & 85.57 \\
SigLIP L/16 256 & 0.224260 & 80.87 & 86.79 & 0.223020 & 79.65 & 85.38 \\
SigLIP L/16 384 & 0.222032 & \textbf{82.64} & \textbf{88.66} & 0.215520 & 80.95 & 87.56 \\
SigLIP So400m/14 224 & 0.206114 & 76.38 & 84.48 & 0.207420 & 75.65 & 83.12 \\
\midrule
SigLIP 2 B/16 224 & 0.222150 & 73.62 & 81.94 & 0.219320 & 73.15 & 81.72 \\
SigLIP 2 B/16 256 & 0.221558 & 75.74 & 82.62 & 0.220260 & 74.70 & 82.34 \\
SigLIP 2 B/16 384 & 0.222801 & 77.81 & 84.91 & 0.219200 & 77.15 & 83.94 \\
SigLIP 2 L/16 256 & 0.221065 & 78.30 & 85.89 & 0.218560 & 77.35 & 83.92 \\
SigLIP 2 L/16 384 & 0.221696 & 79.98 & 86.84 & 0.215380 & 78.35 & 84.96 \\
SigLIP 2 So400m/14 224 & 0.225819 & 78.94 & 85.52 & 0.223100 & 77.85 & 85.03 \\
\bottomrule
\end{tabular}
\end{table}

\subsection{Held-out consequences and paired bootstrap interval}
\label{app:flickr-results}

The agreement rule selects SigLIP B/16 256, and both recall rules select SigLIP
L/16 384. Their one-caption test recalls are 75.20\% and 80.95\%, a 5.75-point
gap (recall-selected minus agreement-selected), with 95\% paired bootstrap
interval [3.80, 7.80]. On the five-caption test metric the recalls are 82.78\%
and 87.56\%, a 4.78-point gap (\Cref{app:tab:flickr-results}). The five-caption
result is the same whether the recall rule uses one-caption or five-caption
validation recall, because both choose the same model.

\begin{table}[htbp]
\centering
\caption{Test consequences of validation selection on Flickr30k. The
agreement-selected model is SigLIP B/16 256 and the recall-selected
model is SigLIP L/16 384 in both rows. Recalls are percentages; gaps and
intervals are points. Intervals describe paired image-level
variability conditional on the selected models and the test galleries.}
\label{app:tab:flickr-results}\label{tab:flickr30k-checkpoint-selection}
\small
\setlength{\tabcolsep}{4pt}
\begin{tabular}{@{}lrrrr@{}}
\toprule
Test evaluation & Agreement-selected & Recall-selected & Gap & 95\% interval \\
\midrule
One caption (primary) & 75.20 & 80.95 & \(+5.75\) & [\(+3.80\), \(+7.80\)] \\
Five captions (secondary) & 82.78 & 87.56 & \(+4.78\) & [\(+3.62\), \(+5.97\)] \\
\bottomrule
\end{tabular}
\end{table}

\paragraph{Bootstrap.}
Each interval uses 10,000 paired bootstrap draws over image clusters, generated
with NumPy's PCG64 and seed 20260909, and takes the 2.5th and 97.5th
percentiles. A cluster contains an image query and its caption queries; for the
five-caption metric, caption-query outcomes are averaged within each image
before the two directions are averaged. Checkpoints, galleries, ranks, and
validation choices stay fixed. The intervals therefore exclude model-selection
uncertainty, gallery variation, and variation across datasets, and they are not
population-level inference about future datasets or models.

\paragraph{Subgroups and correlations.}
Restricting either rule to the SigLIP family, or to the combined
SigLIP/SigLIP 2 lineage, selects the same two models as the full panel, with
the same gaps. Within CLIP, both rules select CLIP L/14 336. Within SigLIP 2, the
agreement rule selects SigLIP 2 So400m/14 224 and both recall rules select
SigLIP 2 L/16 384: the one-caption gap is \(+0.50\) points (\([-1.15,+2.15]\))
and the five-caption gap is \(-0.07\) points (\([-1.21,+1.08]\)). These
prespecified subgroup comparisons are descriptive and do not show an advantage
within every family. Across the 16 models, the agreement--recall Kendall
\(\tau_b\) is \(+0.3667\) for one-caption and \(+0.3167\) for five-caption recall
on validation, and \(+0.3167\) and \(+0.3833\) on test. These associations are
positive, yet agreement still selects a worse retriever, which is the point of
\autoref{sec:study-selection}.

\subsection{Retrospective COCO selection}
\label{app:coco-selection}

These analyses use the 16-model COCO panel of \Cref{app:retrieval} with
original pairs. They were carried out after the full-cohort outcomes were known
and are retrospective.

\paragraph{Full cohort.}
On the complete 5,000-pair cohort, agreement selection picks CLIP L/14 336 and
recall selection picks SigLIP L/16 384, whose bidirectional R@1 values differ by
14.57 points (\Cref{app:tab:coco-full}). The panel-wide agreement--recall
Kendall \(\tau_b\) is \(-0.4667\). This description uses the same images for
selection and evaluation.

\begin{table}[htbp]
\centering
\caption{Original-pair COCO metrics for the 16-model panel. Bold entries
are the maxima selected by each rule. Kendall \(\tau_b=-0.4667\).}
\label{app:tab:coco-full}
\small
\setlength{\tabcolsep}{5pt}
\begin{tabular}{@{}lrr@{}}
\toprule
Model & \(A_{50}(p_0)\) & Bidirectional R@1 \\
\midrule
CLIP B/32 224 & 0.274164 & 32.04\% \\
CLIP B/16 224 & 0.277868 & 34.59\% \\
CLIP L/14 224 & 0.283524 & 37.22\% \\
CLIP L/14 336 & \textbf{0.290168} & 38.30\% \\
\midrule
SigLIP B/16 224 & 0.275812 & 47.23\% \\
SigLIP B/16 256 & 0.279436 & 47.30\% \\
SigLIP B/16 384 & 0.276444 & 49.17\% \\
SigLIP L/16 256 & 0.256448 & 51.10\% \\
SigLIP L/16 384 & 0.249548 & \textbf{52.87\%} \\
SigLIP So400m/14 224 & 0.239244 & 50.81\% \\
\midrule
SigLIP 2 B/16 224 & 0.286752 & 46.88\% \\
SigLIP 2 B/16 256 & 0.287848 & 47.85\% \\
SigLIP 2 B/16 384 & 0.287448 & 49.57\% \\
SigLIP 2 L/16 256 & 0.271912 & 50.16\% \\
SigLIP 2 L/16 384 & 0.270448 & 50.68\% \\
SigLIP 2 So400m/14 224 & 0.267132 & 49.63\% \\
\bottomrule
\end{tabular}
\end{table}

Within the combined SigLIP/SigLIP 2 lineage, agreement selects SigLIP 2 B/16 256
(47.85\%) and recall selects SigLIP L/16 384 (52.87\%), a 5.02-point gap. The
prespecified family gaps are 0 for CLIP, 5.57 for SigLIP, and 2.83 for SigLIP 2.

\paragraph{Disjoint halves.}
One outcome-independent permutation of the 5,000 identities, generated with
\texttt{numpy.random.Generator(numpy.random.PCG64(20260905))}, splits the cohort
into halves A (first 2,500) and B; neither half is re-sorted, and both
modalities share the order. The partition was specified after the full-cohort
results. For each half and model, both 50-neighbor graphs are rebuilt
within the half, and scores are ranked against that half's 2,500-item gallery.
One half selects and the other evaluates (\Cref{app:tab:coco-split}). Raw
\(A_{50}\) is higher in a half because \(c_{50}=50/(n-1)\) grows as \(n\)
shrinks; the graph algorithm is unchanged.

\begin{table}[htbp]
\centering
\caption{COCO disjoint-half selection. The gap is the recall-selected minus the
agreement-selected evaluation R@1, in points. The two orientations
are a split-sensitivity analysis of one cohort.}
\label{app:tab:coco-split}
\footnotesize
\setlength{\tabcolsep}{4pt}
\begin{tabular}{@{}llrrrr@{}}
\toprule
Orientation & Model selected by & Selection \(A_{50}\) & Selection R@1 & Evaluation R@1 & Gap \\
\midrule
\multirow{2}{*}{A \(\to\) B}
 & Agreement: CLIP L/14 336 & 0.341536 & 46.90\% & 45.00\% & --- \\
 & Recall: SigLIP L/16 384 & 0.281768 & 62.20\% & 59.84\% & \(+14.84\) \\
\midrule
\multirow{2}{*}{B \(\to\) A}
 & Agreement: CLIP L/14 336 & 0.338864 & 45.00\% & 46.90\% & --- \\
 & Recall: SigLIP L/16 384 & 0.283496 & 59.84\% & 62.20\% & \(+15.30\) \\
\bottomrule
\end{tabular}
\end{table}

Secondary split gaps are \(+5.22\) and \(+4.50\) points within the combined
SigLIP lineage, \(+5.62\) and \(+5.42\) within SigLIP, and \(+2.16\) and \(+2.92\)
within SigLIP 2. Within CLIP they are zero, because both rules select CLIP L/14
336 in both orientations.

These selection results compare original-pair measurements. They are distinct
from the retention-ranking analysis of \Cref{app:retrieval-ranking}, which
concerns responses to re-pairing normalized by each model's own original and
random values.

\subsection{Post hoc follow-ups}
\label{app:selection-followups}

The analyses below were specified after the Flickr30k and COCO outcomes were
known. They add no new held-out evaluation.

\paragraph{Transfer across datasets.}
Selecting on one dataset and reading the other's recall
(\Cref{app:tab:transfer}) reuses the complete COCO panel and the Flickr30k
results, with lexicographic tie-breaking by model key. Both transfer gaps
were known before this follow-up was specified, so these are not prospective
replications.

\begin{table}[htbp]
\centering
\caption{Transfer of the two selection rules across datasets. Bidirectional
one-caption R@1 in percent; the gap is recall-selected minus agreement-selected.
COCO uses 5,000 images and Flickr30k test 1,000. No interval is reported.}
\label{app:tab:transfer}
\small
\setlength{\tabcolsep}{3pt}
\begin{tabularx}{\linewidth}{@{}YYYrrr@{}}
\toprule
Source \(\to\) target & Agreement-selected & Recall-selected & Agreement R@1 & Recall R@1 & Gap \\
\midrule
COCO \(\to\) Flickr30k test & CLIP L/14 336 & SigLIP L/16 384 & 67.95 & 80.95 & \(+13.00\) \\
Flickr30k validation \(\to\) COCO & SigLIP B/16 256 & SigLIP L/16 384 & 47.30 & 52.87 & \(+5.57\) \\
\bottomrule
\end{tabularx}
\end{table}

\paragraph{The sign reversal between COCO and Flickr30k.}
The agreement--recall correlation is negative on COCO (\(\tau_b=-0.467\)) and
positive on Flickr30k validation (\(+0.367\)). This follow-up reuses the 16
COCO and Flickr30k validation caches; the subset construction and the complete
scale grid were specified before any new outcome was computed. Original caption
choices and pair identities stay fixed, and there is no new extraction,
re-pairing, or test evaluation.

Most of the reversal occurs between families. At \(k=50\), CLIP's within-family
\(\tau_b\) is \(+1.000\) on both datasets and SigLIP's is \(-0.467\) on both
(\Cref{app:tab:sign-flip-families}). Across families, agreement and recall order
18 of 84 model pairs concordantly on COCO and 64 of 84 on Flickr30k; within
families, the counts are 14 and 18 of 36, with no ties. Net concordance
(concordant minus discordant pairs) rises from \(-56\) to \(+44\), which
decomposes into \(+92\) between families and \(+8\) within families. This is a
pair-count identity, not a causal attribution. Across the 120 model pairs,
the agreement ordering changes in 59 and the recall ordering in nine: CLIP has
relatively high agreement but low recall on COCO, and relatively low values of
both on Flickr30k.

\begin{table}[H]
\centering
\caption{Original-cohort agreement--recall associations at \(k=50\). COCO uses 5,000 items and Flickr30k validation 1,014. Subgroup coefficients describe fixed groups of four or six models.}
\label{app:tab:sign-flip-families}
\small
\begin{tabular}{@{}lrr@{}}
\toprule
Checkpoints & COCO \(\tau_b\) & Flickr30k validation \(\tau_b\) \\
\midrule
All 16 & \(-0.467\) & \(+0.367\) \\
CLIP (4) & \(+1.000\) & \(+1.000\) \\
SigLIP (6) & \(-0.467\) & \(-0.467\) \\
SigLIP 2 (6) & \(-0.467\) & \(+0.067\) \\
\bottomrule
\end{tabular}
\end{table}

For each dataset, 25 seeded permutations define nested 500- and 1,000-image
subsets, shared across models, metrics, and scales; the two datasets use
separate random streams. Retained images keep their order and selected caption,
and both neighbor sets and retrieval galleries are rebuilt inside each subset.
At matched \(n=1{,}000\) and \(k=50\), the median \(\tau_b\) is \(-0.583\) on COCO
(range \(-0.762\) to \(-0.400\)) and \(+0.400\) on Flickr30k (\(+0.333\) to
\(+0.500\)), and every subset keeps its dataset's sign at all four scales
(\Cref{app:tab:sign-flip-grid}). At \(n=500\) and \(k=50\), the signs again differ in
every subset, with medians \(-0.393\) and \(+0.544\). Gallery-size mismatch and
the single choice \(k=50\) therefore do not explain the reversal at matched size.
The grid also shows scale dependence: original COCO has \(\tau_b=+0.100\) at
\(k=10\), and some 500-image conditions have mixed signs
(\Cref{app:tab:sign-flip-grid}). At \(k=50\), the agreement and recall maxima
differ in all 100 subset conditions. On Flickr30k at \(n=1{,}000\), every draw
selects SigLIP B/16 256 by agreement and SigLIP L/16 384 by recall, matching the
original validation choices; COCO at that size selects CLIP L/14 336 by agreement
in 18 of 25 draws and SigLIP L/16 384 by recall in 25 of 25. These are
subset-selection disagreements, not new held-out costs.

\begin{table}[tbp]
\centering
\caption{Complete size/scale sensitivity grid. Subset rows show the median Kendall \(\tau_b\), full minimum--maximum range, and positive/zero/negative counts across 25 draws. Rows with \(n=5{,}000\) or \(1{,}014\) are single original-cohort references. Ranges are descriptive, not confidence intervals.}
\label{app:tab:sign-flip-grid}
\small
\setlength{\tabcolsep}{4pt}
\begin{tabular}{@{}lrrrcrr@{}}
\toprule
Dataset & \(n\) & \(k\) & \(k/n\) & \(\tau_b\) & Min, max & + / 0 / -- \\
\midrule
COCO & 500 & 5 & 1.0\% & \(-0.312\) & \(-0.644\), \(+0.143\) & 1 / 0 / 24 \\
COCO & 500 & 10 & 2.0\% & \(-0.450\) & \(-0.740\), \(-0.194\) & 0 / 0 / 25 \\
COCO & 500 & 20 & 4.0\% & \(-0.467\) & \(-0.790\), \(-0.167\) & 0 / 0 / 25 \\
COCO & 500 & 50 & 10.0\% & \(-0.393\) & \(-0.756\), \(-0.075\) & 0 / 0 / 25 \\
COCO & 500 & 100 & 20.0\% & \(-0.100\) & \(-0.346\), \(+0.295\) & 7 / 1 / 17 \\
\addlinespace[3pt]
COCO & 1,000 & 10 & 1.0\% & \(-0.383\) & \(-0.633\), \(-0.183\) & 0 / 0 / 25 \\
COCO & 1,000 & 20 & 2.0\% & \(-0.533\) & \(-0.795\), \(-0.350\) & 0 / 0 / 25 \\
COCO & 1,000 & 50 & 5.0\% & \(-0.583\) & \(-0.762\), \(-0.400\) & 0 / 0 / 25 \\
COCO & 1,000 & 100 & 10.0\% & \(-0.510\) & \(-0.711\), \(-0.200\) & 0 / 0 / 25 \\
\addlinespace[3pt]
COCO & 5,000 & 10 & 0.2\% & \(+0.100\) & --- & --- \\
COCO & 5,000 & 20 & 0.4\% & \(-0.117\) & --- & --- \\
COCO & 5,000 & 50 & 1.0\% & \(-0.467\) & --- & --- \\
COCO & 5,000 & 100 & 2.0\% & \(-0.600\) & --- & --- \\
\addlinespace[3pt]
Flickr30k val. & 500 & 5 & 1.0\% & \(+0.159\) & \(-0.110\), \(+0.510\) & 19 / 1 / 5 \\
Flickr30k val. & 500 & 10 & 2.0\% & \(+0.319\) & \(+0.000\), \(+0.567\) & 24 / 1 / 0 \\
Flickr30k val. & 500 & 20 & 4.0\% & \(+0.427\) & \(+0.176\), \(+0.605\) & 25 / 0 / 0 \\
Flickr30k val. & 500 & 50 & 10.0\% & \(+0.544\) & \(+0.303\), \(+0.628\) & 25 / 0 / 0 \\
Flickr30k val. & 500 & 100 & 20.0\% & \(+0.633\) & \(+0.483\), \(+0.733\) & 25 / 0 / 0 \\
\addlinespace[3pt]
Flickr30k val. & 1,000 & 10 & 1.0\% & \(+0.243\) & \(+0.092\), \(+0.317\) & 25 / 0 / 0 \\
Flickr30k val. & 1,000 & 20 & 2.0\% & \(+0.343\) & \(+0.300\), \(+0.417\) & 25 / 0 / 0 \\
Flickr30k val. & 1,000 & 50 & 5.0\% & \(+0.400\) & \(+0.333\), \(+0.500\) & 25 / 0 / 0 \\
Flickr30k val. & 1,000 & 100 & 10.0\% & \(+0.527\) & \(+0.483\), \(+0.550\) & 25 / 0 / 0 \\
\addlinespace[3pt]
Flickr30k val. & 1,014 & 10 & 1.0\% & \(+0.226\) & --- & --- \\
Flickr30k val. & 1,014 & 20 & 2.0\% & \(+0.333\) & --- & --- \\
Flickr30k val. & 1,014 & 50 & 4.9\% & \(+0.367\) & --- & --- \\
Flickr30k val. & 1,014 & 100 & 9.9\% & \(+0.533\) & --- & --- \\
\bottomrule
\end{tabular}
\end{table}

All 32 original model--dataset records reproduce their saved agreement and
recall values, and direct distance and set-intersection reconstructions agree
exactly in 64 preselected subset cases covering every model and both
datasets. The overlapping subsets measure sensitivity within the retained
datasets; their ranges are not population intervals, and the remaining
dataset-dependent ranking difference does not isolate image content, caption
properties, architecture, or training data as its cause.

\paragraph{Oracle regret, rank weighting, and model-pool sensitivity.}
These diagnostics use only the retained scalar tables of the Flickr30k study and
the COCO split analysis. For a model pool \(\mathcal M\), let \(r_m\) be
model \(m\)'s held-out bidirectional R@1 as a fraction and
\(r^*=\max_{m\in\mathcal M}r_m\). For the agreement (G) or recall (R) rule \(g\),
\begin{equation}
 O_g(\mathcal M)=100\,(r^*-r_{g(\mathcal M)}),\qquad
 \mathrm{Rel@1}_g(\mathcal M)=r_{g(\mathcal M)}/r^*,
 \label{app:eq:regret}
\end{equation}
an observed single-pick analogue of model-selection regret
\citep{renggli2022which}, with Rel@1 following \citet{agostinelli2022stable}.
The paired gap \(100(r_R-r_G)=O_G-O_R\) equals the agreement rule's regret only
when the recall rule attains the oracle. Correlations use tie-corrected Kendall
\(\tau_b\) and the additive hyperbolic weighted Kendall coefficient of
\texttt{scipy.stats.weightedtau}: with zero-based importance ranks \(u_i\), each
pair has weight \(1/(u_i+1)+1/(u_j+1)\). Our \(\tau_w\) averages the coefficients computed
with descending lexicographic ranks of (agreement, recall) and (recall,
agreement), matching \texttt{scipy.stats.weightedtau} with \texttt{rank=True};
\(\tau_w^R\) uses recall-first ranks only. No significance test is inferred. In
the leave-one-model-out analysis, each model is omitted in turn, both rules
reselect on the remaining 15, and the oracle is recomputed among them
(\Cref{app:tab:selection-diagnostics,app:tab:lomo-summary,app:tab:lomo-all}).
These overlapping pools measure sensitivity to the finite roster; they are not
independent replications or population intervals.

\begin{table}[htbp]\centering\small
\caption{Full-panel selection diagnostics. Oracle regrets are in points; Rel@1 is a fraction. Correlations use selection-split agreement and one-caption recall. G and R identify the agreement and recall rules. The Flickr30k rows share their selection pairs. $\tau_w^{R}$ uses recall-first importance ranks; $\tau_w$ averages both importance orders.}
\label{app:tab:selection-diagnostics}
\setlength{\tabcolsep}{3pt}
\begin{tabular}{@{}lrrrrrrr@{}}\toprule
Study & Oracle G & Oracle R & Rel@1 G & Rel@1 R & $\tau_b$ & $\tau_w$ & $\tau_w^{R}$ \\ \midrule
Flickr30k, one caption & 5.75 & 0.00 & 0.9290 & 1.0000 & +0.3667 & +0.1914 & +0.2364 \\
Flickr30k, five captions & 4.78 & 0.00 & 0.9454 & 1.0000 & +0.3667 & +0.1914 & +0.2364 \\
COCO A $\to$ B & 14.84 & 0.00 & 0.7520 & 1.0000 & $-$0.5272 & $-$0.5889 & $-$0.6614 \\
COCO B $\to$ A & 15.30 & 0.00 & 0.7540 & 1.0000 & $-$0.5667 & $-$0.6011 & $-$0.7250 \\
\bottomrule\end{tabular}\end{table}
\begin{table}[htbp]\centering\small
\caption{Leave-one-model-out sensitivity. Each row summarizes all 16 omissions, with both choices and the oracle recomputed among the remaining 15 models. Ranges are extrema across omissions, not confidence intervals. The paired gap is R minus G; it equals oracle regret G minus oracle regret R.}
\label{app:tab:lomo-summary}
\setlength{\tabcolsep}{4pt}
\begin{tabular}{@{}lrrrr@{}}\toprule
Study & Oracle regret G & Oracle regret R & Paired gap & R better \\ \midrule
Flickr30k, one caption & [4.45, 6.60] & [0.00, 0.00] & [4.45, 6.60] & 16/16 \\
Flickr30k, five captions & [2.79, 5.68] & [0.00, 0.19] & [2.60, 5.68] & 16/16 \\
COCO A $\to$ B & [13.36, 16.16] & [0.00, 1.58] & [11.78, 16.16] & 16/16 \\
COCO B $\to$ A & [4.50, 15.30] & [0.00, 0.48] & [4.50, 15.30] & 16/16 \\
\bottomrule\end{tabular}\end{table}
\begin{table}[htbp]\centering\footnotesize
\caption{Oracle regret for every model omission (points). Each G/R pair gives the agreement/recall rule's regret against the best held-out model in that reduced pool. No interval is inferred from these overlapping subsets.}
\label{app:tab:lomo-all}
\setlength{\tabcolsep}{4pt}
\begin{tabular}{@{}lrrrrrrrr@{}}\toprule
 & \multicolumn{2}{c}{Flickr30k one} & \multicolumn{2}{c}{Flickr30k five} & \multicolumn{2}{c}{COCO A$\to$B} & \multicolumn{2}{c}{COCO B$\to$A} \\
Omitted model & G & R & G & R & G & R & G & R \\ \midrule
CLIP B/32 224 & 5.75 & 0.00 & 4.78 & 0.00 & 14.84 & 0.00 & 15.30 & 0.00 \\
CLIP B/16 224 & 5.75 & 0.00 & 4.78 & 0.00 & 14.84 & 0.00 & 15.30 & 0.00 \\
CLIP L/14 224 & 5.75 & 0.00 & 4.78 & 0.00 & 14.84 & 0.00 & 15.30 & 0.00 \\
CLIP L/14 336 & 5.75 & 0.00 & 4.78 & 0.00 & 16.16 & 0.00 & 4.50 & 0.00 \\
SigLIP B/16 224 & 5.75 & 0.00 & 4.78 & 0.00 & 14.84 & 0.00 & 15.30 & 0.00 \\
SigLIP B/16 256 & 6.60 & 0.00 & 5.68 & 0.00 & 14.84 & 0.00 & 15.30 & 0.00 \\
SigLIP B/16 384 & 5.75 & 0.00 & 4.78 & 0.00 & 14.84 & 0.00 & 15.30 & 0.00 \\
SigLIP L/16 256 & 5.75 & 0.00 & 4.78 & 0.00 & 14.84 & 0.00 & 15.30 & 0.00 \\
SigLIP L/16 384 & 4.45 & 0.00 & 2.79 & 0.19 & 13.36 & 1.58 & 13.72 & 0.48 \\
SigLIP So400m/14 224 & 5.75 & 0.00 & 4.78 & 0.00 & 14.84 & 0.00 & 15.30 & 0.00 \\
SigLIP 2 B/16 224 & 5.75 & 0.00 & 4.78 & 0.00 & 14.84 & 0.00 & 15.30 & 0.00 \\
SigLIP 2 B/16 256 & 5.75 & 0.00 & 4.78 & 0.00 & 14.84 & 0.00 & 15.30 & 0.00 \\
SigLIP 2 B/16 384 & 5.75 & 0.00 & 4.78 & 0.00 & 14.84 & 0.00 & 15.30 & 0.00 \\
SigLIP 2 L/16 256 & 5.75 & 0.00 & 4.78 & 0.00 & 14.84 & 0.00 & 15.30 & 0.00 \\
SigLIP 2 L/16 384 & 5.75 & 0.00 & 4.78 & 0.00 & 14.84 & 0.00 & 15.30 & 0.00 \\
SigLIP 2 So400m/14 224 & 5.75 & 0.00 & 4.78 & 0.00 & 14.84 & 0.00 & 15.30 & 0.00 \\
\bottomrule\end{tabular}\end{table}
\FloatBarrier

\FloatBarrier

%% file: sec/supp/e_followup.tex
\subsection{Response within families and across scales}
\input{tab/families}
\Cref{tab:families} gives the split-by-family correlations.
Within SigLIP and SigLIP~2, the $A_{50}$ response is positive on every split.
For the secondary 16-model set it has 18 positive Flickr30k subsets out of 25 and falls short of the 23-subset rule.
The 22-model set has 25 and 24 positive subsets on COCO and Flickr30k.
At $k=10$ the response changes 68 of 231 cross-dataset orders, compared with 25 at $k=50$.
At matched $k/n$, the COCO-$k=50$ versus Flickr30k-$k=10$ MPNet response changes 57 orders, and RBF CKA at those scales changes 69.

Selectivity medians are 0.790 on COCO, 0.734 and 0.740 on its halves, and 0.848 and 0.854 on Flickr30k validation and test.
The partial correlations controlling the text tower's survival range from $+0.515$ to $+0.801$.
Across the 16 shared models at $k=50$, text-reference agreement changes 59 of 120 model orders across data sets, image-reference agreement changes one, and recall changes nine.
The sign change survives the person-content split and exact caption-length matching.

\begin{figure}[htbp]
\centering
\includegraphics[width=3.25in]{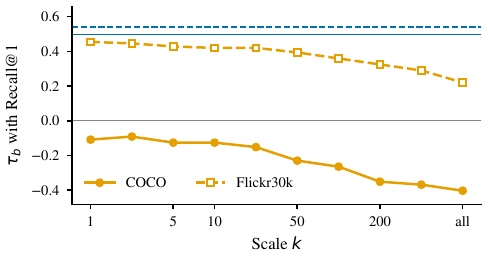}
\caption{\textbf{A finer scale does not give the level a consistent sign.}
Kendall $\tau_b$ between CKA at scale $k$ and Recall@1 over 22 models. Horizontal lines show neighborhood agreement response.}
\label{fig:scale-recall}
\end{figure}

Across all scales and both kernels, none meets the trained-retriever tracking rule. \Cref{fig:scale-recall} shows the RBF results. For linear CKA, all ten correlations on Flickr30k validation are positive and all full COCO correlations are negative except at two neighbors ($\tau_b=+0.0043$). At $k=10$, RBF CKA is negative on both COCO halves and positive on Flickr30k validation. All 18 nearby variants give the same pick on each selection split, and its regret runs from 2.10 to 6.02 points.
The comparison with the $A_{50}$ level on Flickr30k has zero difference because both pick E5-V.

On Flickr30k, the MPNet $A_{50}$ response's selected model exceeds the $A_{50}$ level's pick by 1.55 recall points in the 22-model set (paired 95\% interval $[-0.50,+3.65]$) and by 3.15 points in the 16-model set (paired 95\% interval $[+0.85,+5.40]$).

%% file: tab/families.tex
\begin{table}[t]
\centering\footnotesize\setlength{\tabcolsep}{3pt}
\caption{\textbf{Response correlations vary within families.} Kendall $\tau_b$ with bidirectional Recall@1. The response is the share of $A_{50}$ lost under MPNet re-pairing. With six models, $\tau_b$ changes in steps of about 0.13, and with four models in steps of about 0.33.}
\label{tab:families}
\begin{tabular}{@{}llrrrrr@{}}
\toprule
Family & Reading & COCO & A & B & F30k val. & test \\
\midrule
SigLIP (6) & $A_{50}$ level & $-0.47$ & $-0.60$ & $-0.60$ & $-0.47$ & $-0.33$ \\
SigLIP (6) & CKA level & $-0.73$ & $-0.73$ & $-0.87$ & $+0.47$ & $+0.33$ \\
SigLIP (6) & Response & $+0.73$ & $+0.73$ & $+0.87$ & $+0.47$ & $+0.60$ \\
SigLIP~2 (6) & $A_{50}$ level & $-0.47$ & $-0.55$ & $-0.60$ & $+0.07$ & $-0.33$ \\
SigLIP~2 (6) & CKA level & $-0.60$ & $-0.55$ & $-0.60$ & $+0.47$ & $+0.47$ \\
SigLIP~2 (6) & Response & $+0.33$ & $+0.69$ & $+0.47$ & $+0.87$ & $+0.73$ \\
CLIP (4) & $A_{50}$ level & $+1.00$ & $+1.00$ & $+1.00$ & $+1.00$ & $+1.00$ \\
CLIP (4) & CKA level & $-0.33$ & $-0.67$ & $-0.33$ & $+0.33$ & $+1.00$ \\
CLIP (4) & Response & $-1.00$ & $+0.67$ & $+0.33$ & $-1.00$ & $+0.00$ \\
\bottomrule
\end{tabular}
\end{table}

%% file: sec/supp/f_extensions.tex
\section{Additional operators and modalities}
\label{app:extensions}

This appendix supports the late-interaction and audio results at the end of
\autoref{sec:study-profile} and reports two further extensions: within-video segment
assignment and dense captions. The six additional global image--text encoders
are part of the 22-model panel and are reported in \Cref{app:retrieval}.

\subsection{Protocols and provenance}
\label{app:extension-protocols}

\paragraph{Status.}
Each runner serializes its roster, revisions, assignments, capability gate, and
scoring route with its output, and these were fixed before the final
result-bearing run of each arm. The arms were combined into the present
cross-setting argument after their outcomes were inspected, so we treat the
patterns across rosters as descriptive breadth, not as population-level
inference about models.

\paragraph{Checkpoints.}
Every system has public weights and a public model or paper record
(\Cref{app:tab:extension-roster}). The LanguageBind loader exposes no revision
argument; the listed revision records the intended model, the run stores
the revision the loader resolved, and the two match. All other loaders enforce
the listed revision.

\begin{table}[htbp]
\centering
\caption{Extension systems at their evaluated revisions.}
\label{app:tab:extension-roster}
\small
\setlength{\tabcolsep}{4pt}
\begin{tabularx}{\linewidth}{@{}p{0.26\linewidth}Xl@{}}
\toprule
Role & Public model & Revision \\
\midrule
Late interaction & \texttt{vidore/colqwen2.5-v0.2} \citep{faysse2025colpali,vidore2026colqwen25} & \texttt{dcbe8d9cede5} \\
Late interaction, video, audio & \texttt{vidore/colqwen-omni-v0.1} \citep{vidore2026colqwenomni} & \texttt{c0f50696f483} \\
Late interaction & \texttt{tencent/EVIE-Preview-4.5B} \citep{tencent2026evie} & \texttt{1c43543b000b} \\
Video & \texttt{LanguageBind/LanguageBind\_Video} \citep{zhu2024languagebind} & \texttt{84bed6cfe757} \\
Video, audio & \texttt{Tevatron/OmniEmbed-v0.1} \citep{ma2025tevatron} & \texttt{fa8be315e74d} \\
Video & \texttt{Qwen/Qwen3-VL-Embedding-2B} and \texttt{-8B} \citep{li2026qwen3vlembedding} & \Cref{app:tab:roster} \\
Audio & \texttt{laion/clap-htsat-unfused} \citep{wu2023clap} & \texttt{8fa0f1c6d043} \\
Audio & \texttt{laion/clap-htsat-fused} \citep{wu2023clap} & \texttt{365dea6ef167} \\
Audio & \texttt{laion/larger\_clap\_general} \citep{wu2023clap} & \texttt{ada0c23a36c4} \\
\bottomrule
\end{tabularx}
\end{table}

\paragraph{Preprocessing.}
Late-interaction processors are dynamic: 3,136--602,112 pixels for ColQwen2.5,
up to 802,816 pixels for ColQwen-Omni, and 65,536--786,432 pixels for EVIE.
Captions are passed verbatim to \texttt{encode\_query} with no template, and
images to \texttt{encode\_document}. Audio is mono float32, resampled to 48~kHz
for CLAP and 16~kHz for the Whisper-fronted ColQwen-Omni and OmniEmbed paths.
OmniEmbed is evaluated with its declared query prompt and with bare text; these
are separate configurations and are not averaged.

\paragraph{Assignments, reference, and gate.}
The late-interaction cells use the COCO cohort and reuse the canonical MPNet
assignment of the retrieval study. The audio arm selects one of the five
captions of each AudioCaps test clip with a deterministic hash (seed 20260805),
which observes no embedding or model, and solves the MPNet assignment of
\Cref{app:assignment} over those 975 captions. Both arms use as random reference
the mean of 256 random derangements from the extension runner, seeded from
20260810, each a complete bijection with no fixed point and no identical
normalized caption (\Cref{app:tab:random-sets}). Retention follows
\autoref{eq:retention} without clipping, and bidirectional values average the raw
directional recalls first. Before any retention is reported, a capability gate
requires an original-minus-reference bidirectional R@10 of at least 0.05 and an
original R@10 above five times the uniform-chance R@10. A configuration that
fails is left uninterpreted rather than counted as low retention. In the audio
arm, a negative directional retention remains visible but precludes interpreting
that configuration's bidirectional value.

\subsection{Native late interaction}
\label{app:late-interaction}

The three systems use their deployed MaxSim operator, provided by
\texttt{sentence\_transformers.MultiVectorEncoder} and not reimplemented. For
caption token vectors \(a_1,\ldots,a_m\) and image vectors
\(b_1,\ldots,b_\ell\),
\begin{equation}
 s(a,b)=\sum_{u=1}^{m}\max_{1\le w\le\ell}\langle a_u,b_w\rangle.
 \label{app:eq:maxsim}
\end{equation}
Token vectors and any normalization are those returned by each model. A
single pooled representation, and hence a neighbor graph, is not defined, so
only retrieval is measured. One text-query/image-document matrix is fixed for
every assignment. Image-to-text and text-to-image denote row-wise and
column-wise ranking of that same asymmetric matrix; no reverse image-query
operator is computed.

\begin{table}[htbp]
\centering
\caption{Native late interaction on the COCO cohort. Both directions rank the
same text-query/image-document MaxSim matrix. The bidirectional rows average raw
directional recalls before retention (\Cref{app:notation}). The ordering
\(\rho_{\mathrm{R@10}}>\rho_{\mathrm{R@1}}\) holds in five of six
directions and in all three bidirectional aggregates; ColQwen2.5 image-to-text
is the exception.}
\label{app:tab:late-interaction}
\small
\begin{tabular}{@{}llrrc@{}}
\toprule
Model & Direction & \(\rho_{\mathrm{R@10}}\) & \(\rho_{\mathrm{R@1}}\) & \(\rho_{\mathrm{R@10}}>\rho_{\mathrm{R@1}}\) \\
\midrule
\multirow{3}{*}{ColQwen2.5} & Image to text & 0.3627 & 0.3725 & \textbf{no} \\
 & Text to image & 0.4673 & 0.0994 & yes \\
 & Bidirectional & 0.4468 & 0.1060 & yes \\
\midrule
\multirow{3}{*}{ColQwen-Omni} & Image to text & 0.4276 & 0.4192 & yes \\
 & Text to image & 0.4620 & 0.0889 & yes \\
 & Bidirectional & 0.4578 & 0.0931 & yes \\
\midrule
\multirow{3}{*}{EVIE} & Image to text & 0.3059 & 0.1850 & yes \\
 & Text to image & 0.4502 & 0.0835 & yes \\
 & Bidirectional & 0.4036 & 0.0912 & yes \\
\bottomrule
\end{tabular}
\end{table}

\Cref{app:tab:late-interaction} gives both directions and the bidirectional
aggregate. R@10 retention exceeds R@1 retention in five of six directions and in
all three aggregates. The exception is ColQwen2.5 image-to-text, where each image
ranks candidate captions by text-query MaxSim, which sums over caption tokens
without length normalization. Caption length is a candidate explanation that we
did not test.

\subsection{Audio--text retrieval}
\label{app:audio}

On AudioCaps \citep{kim2019audiocaps}, all six configurations pass the
capability gate. Five interpretable configurations keep
\(\rho_{\mathrm{R@10}}>\rho_{\mathrm{R@1}}\) in both directions and in the
aggregate: three CLAP models and OmniEmbed with and without its query
prompt (\Cref{app:tab:audio-r1,app:tab:audio-r10}). ColQwen-Omni's bidirectional
value is not interpreted, because its audio-to-text R@1 retention is negative;
its directional values remain in the tables. The cohort has 975 clips, so
absolute recalls are not compared with COCO.

\begin{table}[htbp]
\centering
\caption{AudioCaps Recall@1 for all six configurations. Original, hard negative, and reference recalls are percentages; $\rho$ is dimensionless. A and T denote audio and text. Mean first averages the directional recalls. All six configurations pass the capability gate; the final column records retention interpretation. The ColQwen-Omni combined profile is not interpreted because of its negative audio-to-text R@1 retention; its directional results remain visible.}
\label{app:tab:audio-r1}
\small
\setlength{\tabcolsep}{3pt}
\begin{tabularx}{\linewidth}{@{}p{0.23\linewidth}lrrrrY@{}}
\toprule
Configuration & Direction & Original & Hard neg. & Reference & $\rho$ & Status \\
\midrule
CLAP HTSAT unfused & A$\to$T & 31.18 & 5.13 & 0.0701 & 0.1626 & Pass \\
 & T$\to$A & 29.64 & 6.36 & 0.0769 & 0.2125 & Pass \\
 & Mean & 30.41 & 5.74 & 0.0735 & 0.1869 & Pass \\
\addlinespace[3pt]
CLAP HTSAT fused & A$\to$T & 32.00 & 5.54 & 0.0677 & 0.1713 & Pass \\
 & T$\to$A & 27.38 & 6.15 & 0.0721 & 0.2227 & Pass \\
 & Mean & 29.69 & 5.85 & 0.0699 & 0.1950 & Pass \\
\addlinespace[3pt]
Larger CLAP general & A$\to$T & 26.46 & 6.26 & 0.0753 & 0.2343 & Pass \\
 & T$\to$A & 17.85 & 5.54 & 0.0901 & 0.3068 & Pass \\
 & Mean & 22.15 & 5.90 & 0.0827 & 0.2635 & Pass \\
\addlinespace[3pt]
OmniEmbed, instructed & A$\to$T & 11.69 & 1.85 & 0.1006 & 0.1506 & Pass \\
 & T$\to$A & 36.62 & 5.54 & 0.0645 & 0.1498 & Pass \\
 & Mean & 24.15 & 3.69 & 0.0825 & 0.1500 & Pass \\
\addlinespace[3pt]
OmniEmbed, bare & A$\to$T & 21.23 & 3.38 & 0.0777 & 0.1563 & Pass \\
 & T$\to$A & 33.95 & 5.23 & 0.0761 & 0.1522 & Pass \\
 & Mean & 27.59 & 4.31 & 0.0769 & 0.1538 & Pass \\
\addlinespace[3pt]
ColQwen-Omni & A$\to$T & 1.23 & 0.10 & 0.1082 & $-$0.0050 & Negative \\
 & T$\to$A & 20.92 & 5.64 & 0.0889 & 0.2665 & Pass \\
 & Mean & 11.08 & 2.87 & 0.0986 & --- & Not interpreted \\
\bottomrule
\end{tabularx}
\end{table}

\begin{table}[htbp]
\centering
\caption{AudioCaps Recall@10 for all six configurations. Original, hard negative, and reference recalls are percentages; $\rho$ is dimensionless. A and T denote audio and text. Mean first averages the directional recalls. All six configurations pass the capability gate; the final column records retention interpretation. The ColQwen-Omni combined profile is not interpreted because of its negative audio-to-text R@1 retention; its directional results remain visible.}
\label{app:tab:audio-r10}
\small
\setlength{\tabcolsep}{3pt}
\begin{tabularx}{\linewidth}{@{}p{0.23\linewidth}lrrrrY@{}}
\toprule
Configuration & Direction & Original & Hard neg. & Reference & $\rho$ & Status \\
\midrule
CLAP HTSAT unfused & A$\to$T & 81.13 & 44.00 & 0.9207 & 0.5371 & Pass \\
 & T$\to$A & 72.72 & 42.46 & 0.9427 & 0.5785 & Pass \\
 & Mean & 76.92 & 43.23 & 0.9317 & 0.5566 & Pass \\
\addlinespace[3pt]
CLAP HTSAT fused & A$\to$T & 78.56 & 40.51 & 0.9367 & 0.5098 & Pass \\
 & T$\to$A & 68.41 & 39.08 & 0.9411 & 0.5652 & Pass \\
 & Mean & 73.49 & 39.79 & 0.9389 & 0.5356 & Pass \\
\addlinespace[3pt]
Larger CLAP general & A$\to$T & 76.00 & 43.28 & 0.9808 & 0.5639 & Pass \\
 & T$\to$A & 66.26 & 38.97 & 0.9483 & 0.5823 & Pass \\
 & Mean & 71.13 & 41.13 & 0.9645 & 0.5724 & Pass \\
\addlinespace[3pt]
OmniEmbed, instructed & A$\to$T & 31.90 & 14.15 & 0.9948 & 0.4258 & Pass \\
 & T$\to$A & 81.54 & 35.69 & 0.9351 & 0.4312 & Pass \\
 & Mean & 56.72 & 24.92 & 0.9649 & 0.4297 & Pass \\
\addlinespace[3pt]
OmniEmbed, bare & A$\to$T & 58.97 & 26.36 & 0.9559 & 0.4378 & Pass \\
 & T$\to$A & 76.62 & 34.36 & 0.9311 & 0.4417 & Pass \\
 & Mean & 67.79 & 30.36 & 0.9435 & 0.4400 & Pass \\
\addlinespace[3pt]
ColQwen-Omni & A$\to$T & 6.56 & 3.59 & 1.0553 & 0.4601 & Pass \\
 & T$\to$A & 61.54 & 37.74 & 0.9559 & 0.6072 & Pass \\
 & Mean & 34.05 & 20.67 & 1.0056 & --- & Not interpreted \\
\bottomrule
\end{tabularx}
\end{table}

\subsection{Within-video segment assignment}
\label{app:temporal}

This arm changes which annotated segment of a video each caption designates,
with every score fixed; it has no relational diagnostic. ActivityNet Captions
\citep{krishna2017dense} supplies 4,917 decoded validation videos, of which
3,899 have at least three annotated segments. In 39 of these videos, 54 of the
17,505 annotated spans (0.31\%) extend past the decoded file and are trimmed to
the available footage. DiDeMo \citep{hendricks2017localizing} supplies 1,019
decoded test videos. Each description's annotator spans are reduced to the modal
five-second span, with the earliest span breaking ties, and videos with fewer
than three segments or only one distinct span are excluded, leaving 851. Segment
clips use eight uniformly spaced interior frames within the annotated bounds;
LanguageBind applies its own matching eight-frame sampler.

For a video with \(m\) eligible segments, the identity assigns caption \(j\) to
segment \(j\), reversal maps \(j\) to \(m-1-j\), the cyclic shift moves every
target by one, and the seeded derangement has no fixed point and differs from
both reversal and shift. Segment R@1 is computed within the video's fixed
\(m\times m\) score matrix and averaged over videos; chance is the cohort mean of
\(1/m\). LanguageBind's whole-file decoder cannot process the long DiDeMo files
within the shared resource contract, so that cell is absent rather than
evaluated on a shortened cohort. This yields five ActivityNet cells and four
DiDeMo cells (\Cref{app:tab:temporal}).

\begin{table}[htbp]
\centering
\caption{Within-video segment R@1. The identity assignment exceeds chance in
all nine cells, and every scrambled assignment is at or below chance. Reversal
can keep a middle fixed point when \(m\) is odd, so it is not a complete
derangement.}
\label{app:tab:temporal}
\small
\setlength{\tabcolsep}{3.5pt}
\begin{tabular}{@{}llrrrrr@{}}
\toprule
Model & Cohort & Identity & Reverse & Shift & Derange & Chance \\
\midrule
ColQwen-Omni & ActivityNet & \textbf{0.4137} & 0.2575 & 0.2296 & 0.2309 & 0.2783 \\
LanguageBind & ActivityNet & \textbf{0.4138} & 0.2614 & 0.2329 & 0.2251 & 0.2783 \\
OmniEmbed & ActivityNet & \textbf{0.4518} & 0.2499 & 0.2213 & 0.2120 & 0.2783 \\
Qwen3-VL-Embedding-2B & ActivityNet & \textbf{0.4430} & 0.2530 & 0.2204 & 0.2171 & 0.2783 \\
Qwen3-VL-Embedding-8B & ActivityNet & \textbf{0.4560} & 0.2504 & 0.2224 & 0.2087 & 0.2783 \\
\midrule
ColQwen-Omni & DiDeMo & \textbf{0.3331} & 0.2476 & 0.2234 & 0.2228 & 0.2528 \\
OmniEmbed & DiDeMo & \textbf{0.3606} & 0.2415 & 0.2239 & 0.2099 & 0.2528 \\
Qwen3-VL-Embedding-2B & DiDeMo & \textbf{0.3583} & 0.2331 & 0.2273 & 0.2052 & 0.2528 \\
Qwen3-VL-Embedding-8B & DiDeMo & \textbf{0.3765} & 0.2361 & 0.2213 & 0.1985 & 0.2528 \\
\bottomrule
\end{tabular}
\end{table}

These results show that the deployed scorers support the annotated segment
assignments; they do not isolate sequence reasoning.

\subsection{Dense captions}
\label{app:granularity}

This arm asks whether longer, denser captions change R@10 retention. DOCCI
\citep{ono2024docci} uses its official 5,000-image test split, each model's
native image processor, and the model's own text window; cells with more than
5\% truncation are flagged rather than removed. The prespecified outcome is
\texttt{no\_effect}: mean model-level changes in R@10 retention from COCO
to DOCCI are \(-0.0292\) across three long-context models and \(-0.0226\)
across the 16 short-context models, short of the frozen \(-0.05\) threshold
(\Cref{app:tab:granularity}). The long-context group truncates 0--3.76\% of DOCCI
captions and the short-context group 94.40--100\%. The group values are means of
per-model changes, not retentions recomputed from averaged recalls. The COCO
side of each change comes from the extension run's own snapshot of the panel
(\Cref{app:artifacts}). E5-V was also evaluated but belongs to neither group.
Because the cohorts differ in images as well as captions, and many dense
captions are truncated, this result does not identify a causal effect of caption
density.

\begin{table}[htbp]
\centering
\caption{The two primary caption-granularity groups and their members. Each $\Delta\rho_{\mathrm{R@10}}$ is DOCCI minus COCO, with bidirectional recalls averaged before reference adjustment. Group values are arithmetic means of model-level changes. Truncation is the percentage of DOCCI captions truncated by that model's evaluated text pipeline.}
\label{app:tab:granularity}
\small
\setlength{\tabcolsep}{4pt}
\begin{tabularx}{\linewidth}{@{}p{0.18\linewidth}Yrr@{}}
\toprule
Group & Model & $\Delta\rho_{\mathrm{R@10}}$ & Truncated (\%) \\
\midrule
Long context & Jina-CLIP-v2 & $-$0.0357 & 0.00 \\
 & LongCLIP-GmP ViT-L/14 & $-$0.0293 & 3.76 \\
 & LongCLIP-L (Diffusers) & $-$0.0226 & 3.76 \\
\addlinespace[2pt]
Group mean & 3 models & $-$0.0292 & --- \\
\midrule
Short context & CLIP B/16 224 & $-$0.0304 & 94.40 \\
 & CLIP B/32 224 & $-$0.0377 & 94.40 \\
 & CLIP L/14 224 & $-$0.0172 & 94.40 \\
 & CLIP L/14 336 & $-$0.0139 & 94.40 \\
 & SigLIP 2 B/16 224 & $-$0.0396 & 98.76 \\
 & SigLIP 2 B/16 256 & $-$0.0324 & 98.76 \\
 & SigLIP 2 B/16 384 & $-$0.0313 & 98.76 \\
 & SigLIP 2 L/16 256 & $-$0.0211 & 98.76 \\
 & SigLIP 2 L/16 384 & $-$0.0111 & 98.76 \\
 & SigLIP 2 So400m/14 224 & $-$0.0216 & 98.76 \\
 & SigLIP B/16 224 & $-$0.0196 & 98.52 \\
 & SigLIP B/16 256 & $-$0.0173 & 98.52 \\
 & SigLIP B/16 384 & $-$0.0208 & 98.52 \\
 & SigLIP L/16 256 & $-$0.0127 & 98.52 \\
 & SigLIP L/16 384 & $-$0.0208 & 98.52 \\
 & SigLIP So400m/14 224 & $-$0.0142 & 100.00 \\
\addlinespace[2pt]
Group mean & 16 models & $-$0.0226 & --- \\
\bottomrule
\end{tabularx}
\end{table}

\FloatBarrier

%% file: sec/supp/g_reproduction.tex
\section{Reproduction and protocol records}
\label{app:reproduction}

This appendix records when each analysis was specified relative to its
outcomes, the software environments, and where the code and results needed to
check every reported quantity are found.

\subsection{Study status}
\label{app:status}

\Cref{app:tab:status} states, for each analysis, what was fixed before its
outcomes were seen. These are protocol records kept within the project, not
external preregistrations, and later analyses do not make earlier results
prospective.

\begin{table}[htbp]
\centering
\caption{What was fixed before each analysis's outcomes, and how each analysis
relates to earlier observations.}
\label{app:tab:status}
\scriptsize
\setlength{\tabcolsep}{3pt}
\begin{tabularx}{\linewidth}{@{}p{0.25\linewidth}YYl@{}}
\toprule
Analysis & Fixed before its outcomes & Relation to earlier observations & Section \\
\midrule
Invariance checks and semantic assignment & Cohort and caption rules, strengths,
metrics, external objectives, random generator & Original study &
\Cref{app:controls,app:retrieval-representative} \\
CLIP and SigLIP 2 replications & Checkpoints, shared assignments, graph scales,
validation rules & Follow-ups to the SigLIP result &
\Cref{app:control-results,app:retrieval-representative} \\
Rebuilt cohorts; cost-matched alternatives & Reconstruction and allocation rules,
before each run & Sensitivity analyses of existing results &
\Cref{app:retrieval-sensitivity} \\
16-model panel & Roster, ranking and inversion rules & Three rows already
known & \Cref{app:retrieval-panel,app:retrieval-ranking} \\
Six-encoder agreement follow-up & Cohort, saved assignments, graph scales,
reporting rule, before agreement outcomes & Post hoc; retrieval outcomes already
known & \Cref{app:retrieval-panel} \\
CKA reconstruction & Source audit and reconstruction choices & Published
shuffle values known; semantic outcomes unseen & \Cref{app:cka-curve} \\
CKA alternatives, matched preprocessing, conditional ranks & Construction gates,
preprocessing, summary rules & After the original CKA and rank results &
\Cref{app:cka-alternatives,app:cka-variants,app:conditional} \\
Annotation-only CKA and kernel response & MPNet-free tie rule; assignment saved
before evaluation & After the original results & \Cref{app:cka-annotation} \\
Linear and row-normalized fidelity comparison & Reuse of saved assignments; no
construction or selection & Post hoc (2026-09-25) & \Cref{app:cka-variants} \\
QAP audit & Problems, starts, feasibility and evaluation rules & Follow-up of the
published procedure & \Cref{app:qap} \\
Flickr30k selection & Entire protocol before extraction; validation choices saved
before test & Prospective within the project, after the COCO results &
\Cref{app:flickr-protocol,app:flickr-results} \\
COCO selection & Partition before half-specific metrics & Retrospective &
\Cref{app:coco-selection} \\
Transfer, sign reversal, oracle diagnostics & Subsets, scale grid, and
validation cases before new outcomes & Post hoc & \Cref{app:selection-followups} \\
Operator and modality extensions & Rosters, gates, and analysis rules before the
final runs & Cross-setting synthesis is descriptive & \Cref{app:extensions} \\
\bottomrule
\end{tabularx}
\end{table}

\subsection{Environments}
\label{app:environment}

The research profiles target Python 3.11. The repository's
\texttt{requirements.txt} pins the review and CPU-check dependencies, and
\texttt{requirements-experiments.txt} pins the primary extraction versions for
SigLIP, CLIP, SigLIP 2, and the CKA study; \texttt{docs/ENVIRONMENT.md} documents
separate environments for systems that need other library versions. Every result
family also stores the software environment of its run. The CKA study used
Python 3.11.2, NumPy 1.26.4, SciPy 1.16.3, PyTorch 2.6.0, Transformers 4.57.6,
and Sentence Transformers 3.4.1. The six additional encoders used PyTorch 2.6.0,
Sentence Transformers 5.4.0, and Transformers 4.57.6 (Jina-CLIP-v2 and both
LongCLIP variants) or 5.5.0 (E5-V and both Qwen3-VL-Embedding models). The pins
are not complete transitive lockfiles, so reconstruction is validated against
stated tolerances and structural identities rather than byte equality.

\subsection{Code, data, and artifacts}
\label{app:artifacts}

The code-and-results supplement contains the implementation, dependencies,
protocols, the compact records behind every table and figure, and a file
manifest. It omits full embedding caches, model weights, and source datasets,
which are public inputs identified by revision in the protocols.
\Cref{app:tab:artifacts} maps each result family to its compact records and to
the additional inputs that a full recomputation requires.

\small
\setlength{\tabcolsep}{3.5pt}
\begin{longtable}{@{}p{0.06\linewidth}p{0.25\linewidth}p{0.36\linewidth}p{0.25\linewidth}@{}}
\caption{Result families. Compact records reproduce every reported number;
full recomputation additionally needs the inputs in the last column.}
\label{app:tab:artifacts}\\
\toprule
ID & Result family & Compact records included & Additional full inputs \\
\midrule
\endfirsthead
\toprule
ID & Result family & Compact records included & Additional full inputs \\
\midrule
\endhead
A1 & Invariance checks & Protocol, implementation, result report & Embedding caches, control arrays \\
A2 & Semantic assignment (SigLIP) & Assignment and metric code, protocol, report & Assignments, graphs, ranks \\
A3 & CLIP replication & Protocol, code, directional and control report & CLIP caches, full arrays \\
A4 & Rebuilt cohorts & Reconstruction and validation code, all-25 report & Retained-cohort graphs, assignments \\
A5 & Cost-matched alternatives & Frontier and ensemble code, protocols, reports & Costs, 128 assignments, rank arrays \\
A6 & SigLIP 2 replication & Protocol, code, reference and cohort reports & SigLIP 2 caches, full arrays \\
A7 & 16-model panel & Canonical panel record, profile and ranking report & Embeddings, graphs, query-level arrays \\
A8 & Late interaction & Final records, roster and scoring code & Model caches, score matrices \\
A9 & Video and temporal & Final records, cohort and sampling code & Videos, segment representations \\
A10 & Audio & All six configuration records, gates, code & Audio files, representations \\
A11 & Dense captions & Per-model records, group analysis, truncation & DOCCI images, embedding caches \\
A12 & CKA reconstruction & Source audit, protocol, all curve points, validation & Image and caption-group representations \\
A13 & Diagnostic comparison & Diagnostic records and implementation & Saved representations, assignments \\
A14 & COCO full-cohort selection & Original metrics, selections, report & Embeddings \\
A15 & COCO split selection & Split indices, half-specific metrics, choices, gaps & Model embedding caches \\
A16 & QAP audit & Input and feasibility manifests, outcomes, code & Matching inputs, 1,500 permutations \\
A17 & CKA alternatives and matched preprocessing & Construction records, all 128 CKA values, six preprocessing cells, conditional ranks & Assignment, feature, and rank arrays \\
A18 & Flickr30k selection & Protocol, split and caption manifest, validation choices, test metrics, query-level successes, intervals, code & Images, embedding caches \\
A19 & Annotation-only CKA and transfer & Specification, annotation-only assignment and categories, kernel responses, object overlap, transfer & A12 inputs, A1/A2 retrieval inputs \\
A20 & Sign-reversal follow-up & Protocol, subset indices, metric and correlation tables, decomposition, validation & A7 and A18 embedding caches \\
A21 & Six-encoder agreement and 22-model summary & Cohort, saved assignments, per-model records and arrays, extraction records, combined summary & Six embedding caches; A7 arrays \\
A22 & Linear and row-normalized CKA & All variant values, random draws, anchor checks & A12 and A17 arrays \\
\bottomrule
\end{longtable}
\normalsize

\paragraph{Locations and commands.}
Reports for A1--A7 are under \path{experiments/siglip_coco_fingerprint/}; the
compact A7 record is \path{artifacts/checkpoint_panel/}. A8--A11 and A14--A15
are under \path{artifacts/extensions/}, whose index identifies superseded
records. A12, A13, A16, A17, and A22 are under
\path{artifacts/published_cka_usefulness/}, A18 under
\path{artifacts/flickr30k_selection/}, A19 under
\path{artifacts/published_cka_usefulness/mechanism_transfer/}, A20 under
\path{artifacts/selection_sign_flip/}, and A21 under
\path{artifacts/global_graph_retention/}. The oracle, Rel@1, weighted-rank, and
model-omission diagnostics are derived from A15 and A18 and stored in
\path{artifacts/selection_diagnostics/}. From the repository root, the combined
appendix tables and relabeled figures are regenerated from included records by
\begin{quote}\small
\texttt{python paper/iclr2027-appendix/make\_tables.py}\\
\texttt{python paper/iclr2027-appendix/make\_figures.py}
\end{quote}
and the combined 22-model summary and the linear/normalized CKA comparison
are recomputed, given their external arrays, by
\path{experiments/global_graph_retention/panel_summary.py} and
\path{experiments/published_cka_usefulness/fidelity_variants.py}. Each
experiment directory's README gives its commands.

\paragraph{Validation.}
The recorded checks cover ordered cohorts, finite representations, native-score
reconstruction, intervention identities, bijection and exclusion constraints,
tie rules, and agreement between full arrays and reported outcomes; extension
checks cover capability gates, directions, and temporal counts. Where only compact
records are included, they can be checked against the reported numbers, but the
omitted embeddings cannot be regenerated from them. Correctness is assessed
through structural identities and scientific outcomes rather than file digests.

\FloatBarrier

%% file: sec/supp/g_followup.tex
\subsection{Decisions for the scale experiments}
The protocols specify what each analysis reads, the primary splits, and the decision rules. The scale experiment's committed record predates its first outcomes. The projector recipe and proxy table were saved before the full projector fits. Some earlier follow-ups have run records without uniformly committed pre-run versions.
\begin{center}
\small
\begin{tabularx}{\linewidth}{@{}p{.20\linewidth}Y@{}}
\toprule
Analysis & Outcome \\
\midrule
A: matching & Independent CPU replay exceeds the $10^{-9}$ tolerance. The objective tests use the archived kernels. \\
B: response & The 22-model set passes. The 16-model set misses the Flickr30k subset criterion. \\
C: resolution & The 22-model scale and similarity curves retain the original ordering and report crossings per dataset. \\
D: human and model labels & Calibration misses the precision and original-pair limits. The human-only result separates the independent votes from adjudication. \\
E: before alignment & All 360 fits have finite losses and improve from initial to final epoch. \\
F: source of the sign change & The reference comparison locates most order changes in the text tower. Content and caption-length checks miss their criteria. \\
G: scale & Seven replay checks pass. Both swap predictions hold. MPNet misses the crossover criterion. Fine-scale CKA misses the trained-retriever tracking rule and the combined encoder-grid rule. \\
\bottomrule
\end{tabularx}
\end{center}
For each figure, the compact record contains the plotted values and the numerical ledger gives their source. The full representations and public images are inputs to recomputation. Evaluation only uses held-out queries from the designated split. Paired bootstrap intervals resample held-out images.

%% file: sec/supp/h_grid.tex
\section{Before alignment}
\label{supp:grid}
The vision encoders are DINOv2 small, base, and large, ViT base and large, and ConvNeXt large. The text encoders are E5 small, base, and large and BGE small, base, and large.
The symmetric contrastive loss trains one projector per modality with output width 512. The MLP has one GELU hidden layer of width 512.
We use batches of 4,096, 30 epochs, AdamW with weight decay 0.01 and cosine decay, and initial temperature 0.07.
We select a learning rate from 0.0003, 0.001, and 0.003 on the fixed learning-rate validation set, then report seeds 20274100--20274102.
The separately trained one-epoch proxy uses 10,000 examples with seed 20274200 and learning rate 0.001.
The median range of the final five losses as a fraction of the last loss is 0.51\%, with maximum 1.11\%.

\input{tab/proxies_full}
\input{tab/grid_detail}
At ten neighbors, CKA improves rank correlation relative to all-pairs CKA in all eight evaluations. Its worst single-encoder omission regret of 6.98 points exceeds the 4.52-point limit set by all-pairs CKA, so it falls short of the combined criterion.
\begin{figure}[htbp]
\centering\includegraphics[width=3.25in]{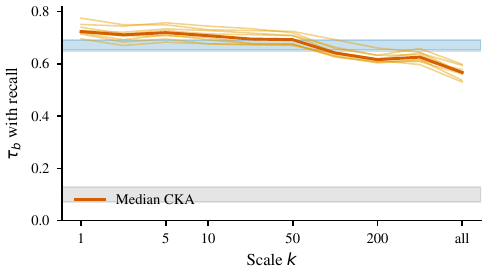}
\caption{\textbf{Finer scales improve the ranking of encoder pairs before alignment.}
Kendall $\tau_b$ between CKA at scale $k$ and retrieval for 36 encoder pairs. Thin lines show eight encoder-grid evaluations, and the bold line shows their median. The bands give the ranges of the MPNet $A_{50}$ response (blue) and the $A_{50}$ level (gray).}
\label{fig:scale-encoders}
\end{figure}

The rank sum of unimodal quality gives mean correlations over fixed vision encoders of 0.84--0.93 with the relative performance of text encoders. Fixed text encoders give much weaker vision rankings ($-0.04$ to $+0.13$).
The removed part of neighborhood agreement correlates with text quality at 0.89 on the MPNet comparison. The surviving part correlates with vision quality at 0.87.
The full record includes residual correlations, predictions with a tower held fixed, paired top-one and top-three regret intervals, and each single-encoder deletion.

%% file: tab/proxies_full.tex
\begin{table}[t]
\centering\small
\caption{\textbf{Rank correlation and top-pick regret differ across proxies.} The range of full-grid regret and the maximum single-encoder omission regret are in recall points.}
\label{tab:proxies-full}
\begin{tabular}{@{}lrrr@{}}
\toprule
Proxy & $\tau_b$ & Regret & Worst \\
\midrule
MPNet $A_{50}$ response & 0.65--0.69 & 1.2--2.7 & 4.0 \\
$\mathrm{CKA}_{10}$ & 0.68--0.74 & 0.0--1.1 & 7.0 \\
All-pairs CKA & 0.53--0.60 & 0.0--1.1 & 4.5 \\
Linear CKA & 0.53--0.60 & 0.0--1.1 & 4.5 \\
Annotation $A_{50}$ response & 0.50--0.58 & 0.7--4.5 & 4.5 \\
MPNet CKA response & 0.46--0.55 & 0.0--0.9 & 8.9 \\
One-epoch probe & 0.41--0.45 & 0.0--3.3 & 4.8 \\
Parameter count & 0.30--0.36 & 0.1--1.6 & 12.6 \\
Unimodal quality & 0.21--0.26 & 0.0--0.9 & 12.7 \\
$A_{10}$ level & 0.17--0.23 & 0.0--0.9 & 12.7 \\
$A_{50}$ level & 0.07--0.13 & 0.0--0.9 & 12.7 \\
\bottomrule
\end{tabular}
\end{table}

%% file: tab/grid_detail.tex
\begin{table}[t]
\centering\small
\caption{\textbf{The finer CKA scale ranks better in every encoder-grid evaluation.} The last two columns show the fine-scale top-pick regret on the full grid and its maximum over single-encoder omissions.}
\label{tab:grid-detail}
\begin{tabular}{@{}llrrrrr@{}}
\toprule
Projector & Evaluation & CKA$_{10}$ $	au_b$ & CKA $	au_b$ & $A_{50}$ response $	au_b$ & Regret & Max. \\
\midrule
Linear & COCO R@1 & 0.727 & 0.594 & 0.667 & 0.00 & 3.52 \\
Linear & COCO R@10 & 0.714 & 0.568 & 0.673 & 0.00 & 2.21 \\
Linear & Flickr R@1 & 0.676 & 0.530 & 0.667 & 0.00 & 6.35 \\
Linear & Flickr R@10 & 0.692 & 0.578 & 0.657 & 0.00 & 2.28 \\
MLP & COCO R@1 & 0.744 & 0.598 & 0.661 & 0.00 & 3.64 \\
MLP & COCO R@10 & 0.730 & 0.565 & 0.651 & 0.33 & 2.27 \\
MLP & Flickr R@1 & 0.676 & 0.537 & 0.692 & 0.50 & 6.98 \\
MLP & Flickr R@10 & 0.701 & 0.561 & 0.653 & 1.15 & 3.45 \\
\bottomrule
\end{tabular}
\end{table}